\documentclass[conference]{IEEEtran}
\IEEEoverridecommandlockouts
\usepackage{cite}
\usepackage{arydshln}
\usepackage{amsmath,amssymb,amsfonts}
\usepackage{graphicx}
\usepackage{mathrsfs}
\usepackage{amsmath}
\usepackage{textcomp}
\usepackage{xcolor}
\usepackage[export]{adjustbox}
\usepackage{algorithm}
\usepackage{algpseudocode} 
\usepackage{subcaption}
\usepackage{multirow}
\usepackage{booktabs}

\makeatletter
\renewcommand\subsubsection{\@startsection{subsubsection}{3}{\z@}%
	{1.5ex plus 1.5ex minus .2ex}%
	{0.5ex}%
	{\normalfont\normalsize\itshape}}
\makeatother

\newcounter{subsubsubsection}[subsubsection]
\renewcommand\thesubsubsubsection{\alph{subsubsubsection}.}

\newcommand{\subsubsubsection}[1]{%
	\refstepcounter{subsubsubsection}%
	\vspace{1.5ex} 
	\noindent\textit{\thesubsubsubsection\ #1}
	\par\nobreak\vspace{0.5ex} 
}

\def\BibTeX{{\rm B\kern-.05em{\sc i\kern-.025em b}\kern-.08em
		T\kern-.1667em\lower.7ex\hbox{E}\kern-.125emX}}

\begin{document}
	
	\title{ Real-Time Obstacle Avoidance Algorithms for Unmanned Aerial and Ground Vehicles \\
	}
	
	\author{\IEEEauthorblockN{Jingwen Wei}
	}
	
	\maketitle
	
	\begin{abstract}
		The growing prevalence of mobile robots in sectors such as automotive, agriculture, and search and rescue highlights the continuous progress in robotics and autonomous technology. In unmanned aerial vehicles (UAVs), often referred to as drones, reports predominantly focus on visual Simultaneous Localization and Mapping, multi-sensor fusion, and the intricate realm of path planning. However, the application of UAVs in disaster-stricken regions for search and rescue tasks remains underexplored, presenting notable challenges in autonomous navigation.
		
		This report aims to develop advanced methodologies for real-time and secure maneuvering of UAVs in complex three-dimensional environments, a crucial aspect during forest fires. This report builds upon previous studies and addresses specific challenges in drone control systems. It specifically aims to create navigation algorithms for rescue routes in unfamiliar and hazardous environments. The report enhances the efficiency and safety of search and rescue operations in natural disasters through the early warning and timely response capabilities of UAVs.
		
		The report progresses through several stages, starting with the exploration of a 2D fusion navigation method initially intended for mobile robots to ensure safe travel in dynamic and unpredictable conditions. This stage sets the foundation for subsequent enhancements, which involve adapting obstacle configurations and refining the algorithm by incorporating extra decision-making processes and re-planning steps to boost adaptability. In its third stage, the report introduces a novel 3D reactive navigation strategy meticulously designed for collision-free avoidance in realistically simulated forest fire scenarios, essential for piloting UAVs in demanding conditions.
		
		Ultimately, the report proposes a groundbreaking multi-system control approach that integrates unmanned aerial vehicles (UAVs) and unmanned ground vehicles (UGVs) to support comprehensive rescue operations in forest fire settings. Each section of the report provides an exhaustive examination of the identified challenges, offers suitable control models, and substantiates them with thorough mathematical analysis and empirical evidence from simulation results. This structured and comprehensive approach yields a report that is both academically significant and practically vital for enhancing the effectiveness of rescue operations during natural disasters, thus safeguarding human and animal lives. 
	\end{abstract}
	

\section{Introduction}
\subsection{Background}
In the early stages of mobile robotics, the field primarily focused on replacing repetitive manual labor tasks, such as those performed by robotic arms. As unmanned technology has evolved, the scope of applications for unmanned vehicles and drones has expanded significantly. At the outset of mobile robotics development, the principal aim was to emulate human cognitive mechanisms, a goal most evident in the domain of motion control. For example, the design of robotic control systems draws inspiration from the human nervous system, particularly the roles of the spinal cord, cerebellum, and brain in orchestrating movement. This analogy is prominently reflected in three primary aspects of robotic motion control: the coordination of fundamental movements, the development of sophisticated path-planning algorithms, and the implementation of environmental perception and interactive capabilities.

Further, the technology applied to Unmanned Aerial Vehicles (UAVs) spans several pivotal areas: foundational control technology, flight control system development, motion control, as well as mapping and positioning technologies. Each area of research necessitates a progression from software-in-the-loop simulation to more complex hardware-in-the-loop simulations, incorporating a variety of algorithms such as Gazebo, ROS (Robot Operating System), embedded systems, and Visual Simultaneous Localization and Mapping (SLAM).

In terms of motion control, UAVs have experienced significant advancements over the past decades. These advancements have found widespread application in both military and commercial sectors. For instance, UAVs are deployed for aerial irrigation, as demonstrated in \cite{xue2016develop}, to enhance operational efficiency and address the unevenness inherent in manual irrigation. Similarly, the use of sensor technology in the cartographic assessment of agricultural lands, as outlined in \cite{elloumi2018monitoring}, represents another significant application of UAVs in agriculture. Moreover, the deployment of drone technology in hazardous environments substantially improves the safety and feasibility of various operations. In this context, drones are primarily utilized for surveillance purposes, such as monitoring road traffic, mining areas, and disaster zones, thereby enhancing operational precision and significantly reducing associated risks \cite{outay2020applications,ren2019review,erdelj2016uav}. Additionally, the domain of safety and rescue is progressively integrating drone technology to provide more accurate and efficient solutions for emergency responses, as highlighted in \cite{tomic2012toward}.

Building upon this research, this report investigates specific challenges in drone control systems, with a focus on the development of navigation algorithms designed to create safe routes in unexplored and hazardous regions. A critical aspect of UAV safety navigation is the ability to adeptly handle diverse scenarios, avoiding collisions and accurately reaching the intended destination, especially in disaster zones and during safety and rescue missions, where the need for rapid response and enhanced safety standards is paramount. 

\subsection{Research Problems and Contributions}	
The primary aim of this report is to explore methodologies by which Unmanned Aerial Vehicles (UAVs) can be maneuvered safely and efficiently in real-time through complex three-dimensional environments. Central to this objective is the development of a robust collision-free navigation control algorithm that operates effectively in three-dimensional space, crucial for achieving predetermined goals. A key application of this research is the deployment of rescue pathways during forest fires, such as the significant 2019 Australian bushfires which markedly affected meteorological patterns, biodiversity, ecological systems, and resulted in substantial human and material losses. The use of UAVs in monitoring natural disasters plays a vital role in providing timely alerts prior to the onset of calamities and enhancing the effectiveness of early-stage rescue operations for both human and animal populations.

Current theoretical research in this domain, however, requires further enhancement. The study \cite{harikumar2018multi}, for instance, introduces a search and control methodology inspired by the Oxyrris Marina, utilizing sensor data for fire detection and firefighting operations. Despite its efficiency in performing target searches and subsequent actions without the need for communication, this framework's application is somewhat restricted to localized fire incidents, making it less suitable for rapidly spreading fires. The forest firefighting system suggested in \cite{wang2018adaptive}, which employs the Vortex Search (VS) algorithm, must more effectively incorporate fire propagation and modeling to fully meet the critical needs of rescue operations. Additionally, \cite{casbeer2005forest} discusses the use of the EMBYR model for simulating forest fire propagation, but primarily focuses on drones' role in information updating, rather than the execution of rescue routes.

Advancements in route planning technology have been significant, particularly in the realms of global and local path planning. According to \cite{elmokadem2021towards}, navigation methods are classified into map-based and map-less approaches, based on their deployment strategies. With an emphasis on enhancing algorithmic processing speed and optimizing travel strategies, research has predominantly focused on global path planning, especially in contexts where comprehensive map knowledge is available. In contrast, reactive path planning is more suited to navigation in unknown terrains, particularly in situations involving moving obstacles and environments that require no prior mapping. This report aims to further advance path-planning strategies, integrating the advantages of multiple route-planning approaches, especially in scenarios involving deformable challenges and uneven terrain.

From these various perspective, the considered navigation problems focus on the following general research problems:

\begin{enumerate}
	\item How to navigate a single UAV to safely reach the desired objective with collision-free movement in a dynamic environment or the presence of deformable obstacles?
	\item How to securely navigate a single UAV over uneven terrain with dynamic or deformable obstacle avoidance to arrive at the predetermined destinations?
	\item How to deploy multiple UAVs coordinated with UGVs to perform safe and efficient search and rescue missions in forest fire scenarios?
\end{enumerate}

The development of an efficient control strategy with low computational demands and fast responsiveness is a central objective of this report. This strategy integrates a sensor-based path planning method with a rapid global planning algorithm to address the challenges of real-time navigation in complex environments. The theoretical foundation for UAV navigation is based on the perception-decision-control framework, which includes both high-level and low-level control. The Perception-Decision-Control paradigm, a fundamental theory in autonomous systems, facilitates the execution of high-level control commands while providing necessary low-level inputs to the controller. This framework addresses complex real-time navigation problems by using methods that reduce computational load while maintaining operational effectiveness.

The key contributions of this report are summarized as follows:
\begin{enumerate}
	\item Developed a solution for collision-free navigation of non-holonomic ground mobile robots in environments with stationary and dynamic obstacles.
	\item Enhanced the adaptability of non-holonomic ground mobile robots for navigating through environments with deformable obstacles.
	\item Introduced a reactive navigation method ensuring safe operation of UAVs in uneven terrains with unknown stationary and moving obstacles.
	\item  Solved the problem of UAV navigation through dynamic and deformable obstacles in forest fire scenarios with a hierarchical rescue algorithm.
	\item Created a collaborative navigation framework for coordinating UAV and UGV teams, with applications in emergency scenarios like bushfire alerts and disaster evacuation.
\end{enumerate}

\subsection{Report Organisation}

This report is organized into multiple phases to employ UAVs for search and rescue operations throughout forest fire incidents.  The initial phase explores the application of a fusion navigation method for a mobile robot in two-dimensional environments, aiming for safe navigation in dynamic and unknown environments. The research then builds upon this, modifying obstacle configurations and enhancing the original algorithm by incorporating decision-making processes and additional replanning steps. In the third phase, a novel three-dimensional reactive navigation strategy is introduced, specifically designed for safe and collision-free avoidance in realistic simulated scenarios. The final phase proposes a multi-system control approach, integrating drones with unmanned ground vehicles for comprehensive rescue operations in forest fire scenarios. Each phase of the report delineates the problems, proposes relevant control models, conducts rigorous mathematical analyses, and validates these through simulation results. The entire work culminates in a comprehensive research article, providing a coherent structure and extensive subject matter coverage. The following sections offer detailed descriptions of the content above:

\begin{itemize}
\item \textbf{Section \ref{mypaper0}} provides a summary of the advancements in autonomous navigation algorithms and the various applications with UAVs. It offers a survey of the literature on current research on UAV-related perception, motion control, and collision avoidance challenges.
	
\item \textbf{Section \ref{mypaper1}}  introduces a hybrid navigation solution for partially known environments with various obstacles, employing the RRT-connect algorithm for initial waypoint generation and an enhanced switcher control for improved stability and efficiency. This research focuses on 2D spaces with regular-shaped obstacles, with further details in subsequent sections.

\item \textbf{Section \ref{mypaper2}} introduces a real-time obstacle avoidance algorithm for mobile robots in 2D dynamic environments, focusing on deformable obstacles. It employs a multi-layered control system and a replanning mechanism for improved effectiveness and energy efficiency. The algorithm outperforms conventional reactive methods in safety, generating safer paths around dynamic obstacles, demonstrated through simplified 2D fire propagation simulations.

\item \textbf{Section \ref{mypaper3}} presents a novel 3D reactive navigation algorithm for UAVs, enabling collision-free travel in uneven terrains. Using a coordinate conversion matrix, it can quickly compute avoidance plans in complex environments. Simulations confirm its efficacy in diverse terrains, including detailed forest landscapes.

\item \textbf{Section \ref{mypaper4}} simulates drone-based rescue operations in forest fire scenarios by utilizing path planning methods. A model is developed to represent forest fire spread across uneven terrains, incorporating both static and dynamic obstacles. A hybrid navigation technique is introduced specifically for rescue missions, optimizing drone rescue paths using a cost function that integrates a disaster coefficient. This coefficient mitigates risks posed by uneven terrain and fire proximity. The algorithm's effectiveness is assessed by analyzing variables such as environmental complexity, path deployment time, total path length, and computational complexity.

\item \textbf{Section \ref{mypaper5}} developed a multi-system framework for fire rescue operations, integrating both UAVs and UGVs for early fire detection and efficient rescue missions. UAVs employ coverage path planning to map the affected area and prioritize rescue efforts after detecting the fire. The algorithm was tested in both simple and complex environments, with results confirming its effectiveness.

\item \textbf{Section \ref{mypaperend}} presents a comprehensive conclusion of this report, encapsulating a summary of the algorithms and contributions delineated in each section. Additionally, it offers a forward-looking perspective, outlining the potential future developments and prospects in this field of study.

\end{itemize}

Preliminary versions of some results of this reports were presented in \cite{GG1,GG2}.
	
\section{Literature Review}\label{mypaper0}

This section presents a comprehensive review of the recent advancements in the field of autonomous navigation technologies and focuses specifically on the application of mobile robots. The discussion is divided into two main parts. The first section explores the evolution of the collision avoidance algorithms. It is crucial to emphasize these fundamental theories to support the applications of UAVs. The subsequent section evaluates the practical application of UAVs in essential operations, particularly in search and rescue missions and surveillance endeavours.
	
\subsection{Typical obstacle avoidance techniques}
Mobile robotic navigation strategies are typically categorized into three distinct methods: global planning, heuristic planning, and a fusion approach. The choice of method varies based on the specific details of the environment in which they are applied. Generally speaking, the global path planning approach is a more straightforward and widely applicable form of obstacle avoidance. By leveraging extensive worldwide data, this navigation technique is more effective in coordinating activities connected to determining routes. On the other hand, local path planning,  referred to as reactive path planning,  is based on the proposed general kinematic model to find a feasible collision avoidance route.  This approach primarily involves high-level planning to guide low-level control inputs, allowing mobile robots to manoeuvre securely and avoid collisions while adhering to flexible constraints in dynamic surroundings. In contrast to approaches that depend on all available information from the surroundings, local path planning is vital in constantly changing situations.	

\subsubsection{General  analysis of global pathfinding algorithms}	
Numerous scholars have worked on substantial studies on path-planning algorithms for several decades. General path planning can be categorized into two types: map-based and mapless-based. Whether the environment is dynamic and if the planning is done online or offline, motion planning algorithms can also be categorized in different research. Global Planner is often defined as procedures that rely heavily on map environments. These planning techniques are inappropriate for dynamic situations or maps with unknown dangers because they are usually offline and unable to adjust to the most recently implemented modifications in the map at any time. The variability of map changes can directly impact the effectiveness of each global Planner algorithm. Geometric, potential field, search, and sample-based methods appropriately describe global planning approaches. These four path planning methodologies are considered classic, each with unique initial planning characteristics.

The Cell Decomposition Approach and Visibility Graphs are two representative methodologies regarding geometric methods. The Cell Decomposition Approach separates space into precise geometric shapes, generating regions suitable for route planning. This method has been thoroughly discussed in the literature \cite{debnath2021different,kloetzer2014petri,gonzalez2017comparative}. A particular self-referencing technique was suggested in the literature \cite{janet1995essential}, which improves the efficiency of route planning while reducing the complexity of path planning, data storage, and processing time. 

The artificial potential field method simplifies avoiding obstacles by converting map obstructions into repulsive potential fields, guiding vehicles in avoiding manoeuvres. This method is a simple and efficient planning strategy, even though it reduces the difficulty of avoiding obstacles. As mentioned in the literature  \cite{warren1989global}, this approach does have a limitation in that it might not be able to get the vehicle to where it is going since it is to guide it away from blockages and toward the minor possible field. As research  \cite{park2001obstacle} illustrates, several researchers have attempted to tackle this problem in their subsequent research by adding further restrictions or combining it with other techniques.

Some search-based approaches, such as the Dijkstra, A*, and D* algorithms  \cite{wang2011application,tang2021geometric,raheem2018path}, are widely used in path planning \cite{GG3}. The approaches are Search-based because they generate paths by extending from the starting point and traversing the entire map while applying different constraints to find the best route. The A* algorithm improves beyond the standard Dijkstra technique by incorporating a heuristic that assesses the cost from the beginning to the current position and from the current point to the destination at each search step. This guarantees the discovery of the shortest path. Although search-based approaches can ensure optimality, a significant drawback is their high computational cost, which limits their extensive practical applicability. Many researchers have suggested new optimization methodologies within this paradigm to enhance algorithmic performance. One approach to lessen the computing overhead of scanning the entire map is implementing incentive mechanisms or extra constraints  \cite{tang2021geometric,erke2020improved}. \cite{erke2020improved} has improved upon the original A* algorithm by introducing performance evaluation criteria and optimizing the heuristic function. This enhancement has improved obstacle avoidance capabilities and increased the robustness and stability of the method.Despite the widespread use of the A* algorithm in various applications, its high computational cost and inability to adapt to real-time changes in dynamic environments limit its practical applicability in fast-paced rescue operations.

Apart from global algorithms, sampling-based techniques are also a widely used option for path planning.  The Rapidly-exploring Random Tree (RRT) technique, initially introduced by Lavalle \cite{lavalle1998rapidly}, can solve various path planning problems involving holonomic, nonholonomic, and kinodynamic restrictions.   This approach efficiently identifies viable routes within a brief timeframe by constructing stochastic expansion trees.   Sampling-based techniques may need to be more effective in finding suitable pathways due to insufficient coverage or result in energy inefficiency due to excessive coverage.  Improved algorithms like RRT-connect \cite{kuffner2000rrt}, RRT* \cite{wang2020improved}, and Bidirectional RRT \cite{liu2019goal,fan2023uav}, which optimize the original method by restricting the expansion space, have been presented in later studies to overcome these problems.  The Probabilistic Roadmap (PRM) is a well-known random method that constructs a graph by randomly sampling nodes in the free configuration space and connecting them with edges \cite{kannan2016robot}. This approach aims to identify an optimal path from the starting node to the goal node.   However, the PRM approach may need more effectiveness in intricate settings, and the paths it generates could be more consistently ideal.   Current research endeavours are concentrated on algorithm fusion to alleviate these limitations \cite{farooq2017quadrotor,xu2020fast}.
	
\subsubsection{Local avoidance approaches and other strategies} 
Traditional global algorithms often become entrapped in local optimal problems.  Consequently, the concept of local, or namely reactive, planner was introduced by subsequent scholars.  This method incorporating real-time sensor information indicates more success in uncertain or dynamic environments.  Early studies on local planning approaches predominantly utilized the concept of configuration space to restrict the optimization problem within velocity space.  The Dynamic Window Approach (DWA) is a research of this, which entails evaluating several sets of velocities and modelling the robot's velocity trajectory.  According to reference, \cite{seder2007dynamic}, the robot can reach required velocities and obstacle-avoidance trajectories because of limitations related to kinematics, safety constraints, and local navigability inside the velocity space.  The author proposes that the dynamic obstacles should be regarded as the movements of moving units on the grid map to achieve trajectory prediction using collision detection.  This kind of method efficiently devises routes using map data to update continuously.  Furthermore, the research \cite{kobayashi2022local} combined the DWA approach with the robotic arm application to achieve the path design assessment.  The potential pathways are generated by applying DWA and considering the anticipated positions of moving impediments. 

Moreover, the Curvature Velocity method \cite{simmons1996curvature}, Collision Cone approach \cite{chakravarthy1998obstacle}, and Velocity Obstacles \cite{van2011reciprocal} are techniques that modify the robot's immediate velocity in order to avoid obstacles by adjusting its velocity space.  Simmons' early investigation in \cite{simmons1996curvature} on the effects of velocity restrictions on collisions with nearby obstacles suggested that optimal velocity trade-offs may be achieved for efficient obstacle avoidance by integrating the robot's velocity limitations with environmental constraints.  In \cite{chakravarthy1998obstacle}, the concept of collision cones enables the prediction of collisions between two moving entities.  In  \cite{van2011reciprocal}, building on the basic Velocity Obstacle (VO) approach, the Acceleration-Velocity Obstacle (AVO) method was proposed, supplementing new acceleration constraints that overcome the prior requirement for immediate deceleration in robots as identified in previous studies.

The Boundary Following (BF) method is a local planner technique that integrates sensor-based strategies for exploring unknown terrains. Only the visible parts of obstacles are detected by the sensors. This  activates a local obstacle avoidance system to prevent collisions only when the robot encounters an obstacle. Then, the robot records the coordinates of the impact point, triggering the avoidance mechanism only when the distance to an obstacle falls below a specific threshold \cite{matveev2012real}. When employing this method, the detection range of the sensors must be considered. Due to its low computational cost and fast deployment, the BF Method has been widely applied in local planner areas. For instance, in study \cite{matveev2020method}, the objective of reaching a predetermined point while maintaining a safe margin is achieved through limited sensory data and reactive control. Another approach in \cite{saeed2020boundary} is utilising the BF method for finding the optimal collision-free path in complex environments. Inspired by biology, a restricted control input navigation algorithm enables autonomous collision-free flight in dynamic and unknown environments for unmanned aerial vehicles \cite{savkin2013simple,matveev2012real,wang2018strategy}.  To enhance the optimality of real-time computation in path planning, study \cite{xiao2016dynamic} explored a real-time path planning method based on boundary value problems capable of real-time tracking of moving targets within a terrain grid model.

In addition to the significant path-planning strategies previously summarized, other techniques have gained widespread application. For example, the Coverage Path Planning (CPP) method requires mobile robots to generate an optimal path that thoroughly covers the target area without repetition and avoids all obstacles. Owing to this characteristic, this algorithm is extensively used in applications such as cleaning robots, underwater structural inspection, lawn mowing, and agricultural and forestry irrigation \cite{galceran2013survey,elmokadem2019distributed}. Model Predictive Control (MPC) has also seen significant research progress in trajectory planning in recent years through the development of kinematic models. However, due to limitations in obstacle avoidance optimization in MPC, researchers often combine it with other planning methods for more comprehensive path planning studies \cite{hu2019dynamic}. Additionally, the combination of sliding mode controllers with other planners achieves obstacle avoidance and can be used for trajectory optimization \cite{elmokadem2019control,choi2000optimal}.

\subsubsection{Review of Collision avoidance in 2D/3D environments}
In discussing navigation algorithms for unmanned vehicles, it is crucial to differentiate between Unmanned Aerial Vehicles (UAVs) and Unmanned Ground Vehicles (UGVs), primarily due to their distinct operational environments and physical capabilities. One key distinction lies in their respective navigational dimensions. UAVs operate in a three-dimensional space, which necessitates accounting for vertical movement and hovering. This complexity demands advanced control systems for stabilization, altitude management, and navigation in varied air currents. Conversely, UGVs navigate within the constraints of ground-level obstacles, focusing on surface characteristics and gradients. In \cite{jones2023path}, the authors investigated the development level of autonomous unmanned aerial vehicle (UAV) path planning technologies based on environmental complexity classification. The capability of UAVs to perform a variety of aerial manoeuvres in three-dimensional space leads to an exponential increase in the number of potential paths that can be explored. This indicates broader application prospects for UAV systems in more complex planning scenarios in the future and presents unprecedented challenges. UAVs exhibit high manoeuvrability and autonomous data collection capabilities, enabling them to fly at various altitudes and plan optimal trajectories. The 3D reconstruction method presents a potential solution to overcome the challenges of UAVs' limited endurance and other practical flight duration constraints \cite{maboudi2023review}.

The development of path planning technologies for autonomous vehicles has primarily focused on optimizing algorithms to achieve collision-free navigation and enhance real-time adaptability, which are critical for safe operation in dynamic environments. While Model Predictive Control (MPC) is often associated with dynamic vehicle models due to its ability to predict and optimize future states, it is also applicable to kinematic models. For instance, studies have demonstrated that MPC can be effectively employed to ensure accurate path tracking and optimize short-range trajectories even in simplified kinematic contexts. \cite{ju2022mpc} presents an MPC algorithm for nonholonomic mobile robots, which achieves stable path planning under input constraints and unknown disturbances using local sensing information. Similarly, \cite{hoy2012collision} applies MPC to plan short-range trajectories around detected objects, while guaranteeing that the vehicle remains within the sensor’s visible range. Additionally, \cite{barreno2023efficient} presents a computationally efficient control method for underwater vehicles, demonstrating that even with limited computational capacity, it is possible to estimate both kinematics and disturbances, thereby bridging the gap between dynamic and kinematic models in MPC applications.

The combined development of Unmanned Aerial Vehicles (UAVs) and Unmanned Ground Vehicles (UGVs) has become an essential trend in intelligent systems and robotic technologies in recent years. This integration capitalises on both strengths, overcoming their respective limitations, enhancing environmental adaptability, and enabling multi-system operations to execute complex tasks flexibly. For example, the collaborative unmanned systems approach described in \cite{christie2017radiation} can autonomously search for sources of hazardous radiation in a given scenario. The UAV provides essential background information to the unmanned vehicle by scanning a large area, ensuring that the unmanned vehicle can efficiently search the area and verify the presence of radiation sources at the estimated location. In \cite{arbanas2018decentralized}, the authors propose a symbiotic vehicle-ground vehicle control system that provides additional degrees of freedom to unmanned ground vehicles (UGVs) to enable them to cross obstacles and to achieve highly accurate motion planning as well as ad hoc decentralised mission planning for complex tasks.

However, despite these advancements, existing local and global planning algorithms often struggle with balancing computational efficiency and real-time adaptability, particularly in highly dynamic and unpredictable environments. This presents a significant research gap that this report aims to address. In summary, while both global and local path planning techniques have made significant strides, their respective limitations in dynamic settings underscore the need for more robust and adaptive algorithms. The following section will explore specific applications of UAVs in search and rescue missions, where these challenges are especially pronounced.

\subsection{Applications of UAVs}

\subsubsection{Approaches in Monitoring and Surveillance Systems}

In recent decades, the progress in autonomous driving technology has opened up new possibilities for the application of drones across various domains, including surveillance, wireless communication support, search and rescue, and parcel delivery \cite{huang2021navigating,savkin2019range,andujar2018intelligent,GG4,chiang2019impact,GG7,GG6,GG5}. Surveillance tasks Among these applications stand out as particularly significant, where coverage performance and observability play pivotal roles. Given the inherent constraints of drones' Field of View (FoV) and the variability in target detection density, the prompt and practical assessment of surveillance quality is paramount in drone-based surveillance applications.

To address these challenges, the research presented in \cite{huang2019algorithm} delves into exploring how teams of drones can effectively monitor diverse targets, such as pedestrians and vehicles, within limited observable ranges. Their proposed approach focuses on minimizing the average distance between drones and targets, consequently enhancing surveillance quality through the establishment of improved communication links. Additionally, \cite{huang2021navigating} introduces an innovative drone target coverage model, enabling drones to dynamically evaluate the coverage status of surveillance targets and the extent of their observation. Furthermore, \cite{trotta2018joint} puts forth a coverage strategy that leverages consumer drones to maximize the coverage of predefined target areas through centralized optimal solutions. Simultaneously, it aims to enhance cost-effectiveness in distributed deployment scenarios. Of particular interest, \cite{das2018tracking} addresses the intriguing problem of determining the minimum number of drones required to track a set of specified targets effectively. This research investigates target tracking strategies in two distinct scenarios, with the primary objective being the utilization of the fewest mobile trackers to achieve the desired tracking outcomes. These developments underscore the significant strides in integrating drone technology into surveillance applications, offering promising solutions for enhanced performance and efficiency.

\subsubsection{Applications in Search and Rescue}
Natural disasters significantly affect society every year, extending to the economic, ecological, and safety facets of life and property. The unique and unforeseen hazards brought about by disasters such as fires, earthquakes, and maritime mishaps endanger victims and rescue workers. In this context, the application of drone technology has emerged as a significant and prudent development in disaster response. Drones can efficiently perform personnel searches during the early stages of a crisis, when risks are comparatively more minor, by incorporating modern technologies into search and rescue efforts.   This approach effectively prohibits rescue personnel from accessing perilous zones without specific data regarding the presence of survivors. Innovative developments in drone technology, particularly in areas such as swarm formation, mapping, position determination, and the utilisation of advanced sensors, have been effectively incorporated into search and rescue operations.
Furthermore, path planning technology is essential in disaster response plans.   The authors in the literature \cite{andujar2018intelligent} devised a technique for evaluating potential risks using discrete path planning and grid mapping.   They examined different search and rescue tactics inside a lifelike simulated setting and showcased the efficiency of their method.

Micro unmanned aerial vehicles (UAVs), renowned for their affordability and portability, have garnered considerable attention in the wilderness search and rescue domain. They have demonstrated their effectiveness in augmenting the success rate of locating missing individuals \cite{lin2009uav}. These UAVs play a pivotal role in gathering crucial data from disaster-stricken areas and transmitting it to rescue centres, enabling rescue personnel to execute missions without direct proximity to the incident site. This approach facilitates more efficient rescue strategies and elevates the overall quality of rescue operations.

Research within the realm of emergency search and rescue employing UAVs has made significant advancements in several key areas, including visual image recognition, exploration of application scenarios, and flight path planning. Computer vision technologies are employed to process real-time images captured by UAV cameras, utilizing image processing units and GPS data, along with dynamic recognition techniques for target detection \cite{liu2021mode}. Moreover, an effective classification method rooted in K-means clustering has been developed to identify missing individuals in wilderness settings by analyzing aerial imagery and human body proportions. This method entails categorizing and recognizing color clusters in three-dimensional space using automatically selected color categories. It empowers UAVs to navigate disaster areas as required, resulting in a substantial reduction in search and rescue durations \cite{niedzielski2017nested}.

Incorporating drone communication into path planning represents an effective strategy for tackling the challenges encountered in Search and Rescue (SAR) missions \cite{zeng2016wireless}. Throughout the rescue operation, drones seamlessly integrate into the existing communication infrastructure. The research in \cite{coutinho2018unmanned} extends network coverage and serves as essential components, acting as base stations or relay stations for tasks such as area coverage and target detection tracking. This direction focuses on optimizing the path planning of drones while considering communication requirements and constraints, leading to efficiency improvements across various dimensions, including energy consumption, detection speed, and risk assessment.

Furthermore, a jointly optimized multi-drone path planner equips drones with sensors and communication interfaces, enabling them to search for specific fixed targets within unknown territories systematically. These targets can manifest anywhere within the designated search area. Once located, they facilitate the establishment of multiple communication links within the multi-drone system, ensuring continuous surveillance of the target. This path planner streamlines task execution and connectivity performance, achieving a harmonious optimization that simultaneously addresses the search and connectivity objectives \cite{yanmaz2023joint}.

Using drone coverage path planning has become increasingly prevalent in addressing search and rescue (SAR) solutions challenges. In \cite{cho2021coverage}, a two-stage approach was introduced to tackle maritime SAR challenges effectively. The initial stage of this approach employs a region decomposition method, which minimises the search area and transforms it into a graph comprising vertices and edges. Subsequently, a model for generating rapid coverage paths is developed in the second stage. The proposed RSH algorithm is employed within this framework to determine the optimal solution with the shortest time requirements, even in large-scale scenarios. Similarly, in reference \cite{kyaw2020coverage}, researchers applied the Coverage Path Planning (CPP) methodology and introduced an algorithm grounded in Deep Reinforcement Learning (DRL). This algorithm is specifically designed for drones with SAR sensors, enabling efficient area surveillance and optimization of search and rescue operations. Notably, it contributes to a reduction in problem-solving complexity, conservation of energy resources, and the avoidance of inefficient tasks. These advancements demonstrate the efficacy of drone coverage path planning in enhancing SAR mission effectiveness.

While UAVs have demonstrated significant potential in search and rescue operations, current approaches often lack the ability to dynamically adapt to changing environmental conditions in real-time, a critical capability in disaster response scenarios. To address these limitations, this research focuses on developing novel algorithms that integrate advanced path planning with real-time environmental sensing, thereby enhancing the adaptability and efficiency of UAVs in rapidly changing and critical mission environments.

\section{A hybrid algorithm for collision-free navigation of a non-holonomic ground robot in dynamic environments with stationary and moving obstacles}\label{mypaper1}
In this section,  a novel hybrid navigation for collision-free movement in a dynamic environment was proposed. It considers the vehicle operating in obstacles that are both stationary and moving. The proposed method combines global path planning with reactive methods to guarantee safe navigation. This hybrid method provides an optimal solution for safe navigation in a partially known, dynamic environment by combining the benefits of global and reactive navigation to overcome the limitations of relying primarily on one of them. Computer simulations are conducted to validate the performance of the proposed method.

\textbf{Some of the work is part of the paper:} \textbf{J Wei}, A Hybrid Algorithm for Collision-Free Navigation of a Non-Holonomic Ground Robot in Dynamic Environments with Steady and Moving Obstacles, In 2023 42nd Chinese Control Conference (CCC), pp. 2928-2933, Tianjin, China, July 2023.

\subsection{Introduction}

The fundamental requirement of mobile robot navigation is safe operation in an unfamiliar or dynamic environment. The technique to planning is typically separated into global and local path planning, both of which have limitations such as expensive computation and inadequately safe on unknown terrain. To ensure safe navigation, a combination of the two planning approaches is more appropriate for development.

Based on a priori knowledge of the overall environment, global route planning algorithms can execute efficient and optimum motions. These approaches depend on updating the planned path while determining a collision-free path between known obstacles in the environment. In general, the global technique could be categorized as sample node-based or global search-based. Traditional theories A-star \cite{hart1968formal}  path planning, the Dijkstra algorithm  \cite{xu2007improved}, and the D-star technique \cite{koenig2005fast} are all belong to the global search category. These navigation algorithms generally involve heuristic functions to locate the shortest route. Some of algorithms based on sampled points like Rapidly-exploring Random Tree (RRT) \cite{lavalle1998rapidly,kuffner2000rrt}  and Probabilistic Roadmaps (PRM) \cite{kavraki1996probabilistic}  are more popular due to faster speed. Compared with graph search method, this type of navigation can generate faster planned path through generating sampled points in the environment. However, the graph search method is not superior in terms of speed but can find the shortest path. In addition, some global path planning algorithms have failed to handle the dynamic environment navigation problem since these global approaches rely on priori environmental knowledge. Some previously invented approaches tackled this issue by continuously refining the map to modify the initial planned path, which is inefficient and sophisticated computation. Therefore, the dynamic obstacle avoidance problem solved by proposing a local planner approach is more applicable to local obstacle avoidance than global path planning.

Local planner is an alternative technique to plan a route to reach the destination in dynamic or unknown environment. The reactive technique disregards the entire map and focuses on the restricted environment observed by sensors or other methods. In other words, some the reactive or local navigation algorithm still faces the issue of safe navigation with this fragment planning approach, since it needed more effort to compute when the environment is kind of complex or the obstacles is too huge to detect. There several classical types of local navigation algorithms such like dynamic window \cite{fox1997dynamic}  and velocity obstacles  \cite{fiorini1998motion}. In  \cite{missura2019predictive}, the proposed Dynamic Window Approach (DWA) control provides the physical modelling of moving objects and the prediction of future collisions in the environment. The DWA process is expected a static route analysis of moving obstacles. Besides, the concept behind velocity collision detection is to project the relative connection between objects into the velocity domain in order to evaluate if there is a danger of collision which is based on to determine the velocity falls within the obstacles velocity zone.  \cite{chen2018ship} presents a discrete non-linear velocity obstacle collision detection approach to address ship collisions. However, numerous local navigation algorithms are required to be implemented with sensor data. In \cite{benjamin2019obstacle} , one method is to create a manoeuvring objective function for obstacle avoidance by identifying around obstacles, modifying local mistakes at any course point enables the partial avoidance of obstacles. The alternate solution similarly employs live cameras and MPC to produce route trajectories with non-linear restrictions  \cite{nageli2017real}. In comparison to the dynamic window approach, velocity avoidance reduces the amount of computation, similarly to the boundary tracking method (see \cite{savkin2013reactive,savkin2013simple,matveev2011method,matveev2012real} ). The boundary tracking approach enables a rapid response to avoid predicted obstacles with low computational cost. However, this strategy may face trap situation which may cause path finding failed.

To overcome for above shortcomings of applying global and reactive path planning purely, hybrid approaches have been developed. The global planner can generate fast and reliable path based on priori knowledge of initial environment and reactive planner can ensure vehicle to avoid unknown or dynamic potential danger in the local environment. A probabilistic local planner (ProbLP) based approach combined with Dynamic Rapidly-exploring Random Tree (DRRT) search paths enables safe navigation in dynamic unknown complex environments  \cite{d2017safe}. This approach separates the challenge of avoiding obstacles in complicated environment into two portions, with the DRRT algorithm solely taking into account static obstacles and the avoidance of moving obstacles is handled only by ProbLP. In \cite{wang2020dynamics}, the authors use a hybrid algorithm combining fuzzy decision-making (FDM) and Fine Dynamic Window (FDW) to establish a trackable collision avoidance local path by adjusting yaw velocity and acceleration. Besides, A-star method combine with dynamic window reactive method can meet the needs of mobile robots in complex and dynamic environments (see \cite{chen2019hybrid,zhong2020hybrid}). In general, applying hybrid method should consider the adaptability of algorithms. A possible hybrid algorithm is the combination of RRT and sliding mode control methods for obstacle avoidance control in dynamic or unknown environments that exploits the benefits of RRT and integrates it effectively with local obstacle avoidance \cite{elmokadem2018hybrid,elmokadem2021hybrid}. Some approaches make the navigation task to layered mission and the navigation planning controller can implement global to local trajectory optimization in a complete mission \cite{elmokadem2018hybrid,elmokadem2021hybrid,nakhaeinia2015hybrid,niewenhuisen2016layered,zhu2012new}.

Inspiring by previous observation, Although Reactive method is popular applied in dynamic or unknown environment due to the computationally efficient \cite{wang2020dynamics}, using global path planning algorithm in dynamic environment can require more safely constraints as it is not safe enough whenever add new obstacles detected by sensors. Considers the robot's nonholonomic and dynamic constraints, it is extremely challenging to discover the shortest global route with trackable waypoints, requiring the development of hybrid approaches to deal with unforeseeable circumstances.
In this section, a hybrid path planning method is proposed that combines the RRT method (global path planning) with a biologically inspired reactive algorithm for safe navigation in a partially known environment. Assume that the robot can be described by a standard non-holonomic model with a strict angular velocity constraint. For this purpose, RRT is a suitable global path planning choose due to time-efficient. In order to be able to obtain an optimized route, inspired by \cite{elmokadem2018hybrid,yang2010efficient}, the original RRT route can be improved in order to reduce redundant path consumption. In addition, the reactive method proposed by  \cite{savkin2013simple} is inspired by biology and imitates the dynamic obstacle avoidance of animals employing simple motor control rules. This localized approach can reduce the computational complexity and thus achieve the goal. Besides, this navigation strategy utilizes a switch condition to handle the transition between path following and obstacle avoidance. Overall, the main goal of the current study was to overcome the limitations of using a purely global or reactive method while retaining the benefits of both, which can provide rapid responses and lower computation costs.

This section is organized as follows. Section \ref{my1s2} present the system description and the navigation problem. In section \ref{my1s3}, navigation algorithm will be introduced. Then, the computer simulations results are provided by performing the presented method. Section \ref{my1s5} presents brief conclusions.

\subsection{Problem Statement}\label{my1s2}
Consider a partially known closed environment \( Z \) filled with unknown moving obstacles \(\eta=\left\{\eta_1, \eta_2, \cdots, \eta_i\right\} \subset \mathcal{Z}\). In this environment, the boundary and the already present stationary obstacles are known as priori conditions of the environment. A nonholonomic mobile robot \( R \) is required to navigate safely through the environment with collision-free to reach a goal location \( \subset \mathcal{Z} \). To be more specific, the priori known stationary obstacles are utilized to generate a global planned path and the environment with moving obstacles is used to do reactive path planning. In addition, the position of robot \( O_r \) \( \subset \mathcal{Z} \) can be described in x-axis and y-axis \( O_r = [x,y]^T \). We assume that each obstacle \(i\) constitutes a dynamic, closed, bounded planar set. The robot \( R \) can measure the current distance \(d_i(t)\) from its border, defined by
\begin{align}
	{d_i}({t}) =\min _{\boldsymbol{O_r}^{\prime} \in \eta}\left\|\boldsymbol{O_r}-\boldsymbol{O_r}^{\prime}\right\|  \quad \forall t	
\end{align}
\( d_i (t) \) can denote the current distance to the nearest moving obstacle \( \eta_i \). The safe navigation requirement is maintaining a safe distance \(d_{safe}\) between \( R \) and moving obstacles \( \eta_i \) (i.e., \(d_i (t) \geq d_{safe}\), \(\forall t \)).

In this expression, \(\|\cdot\|\) represents the standard Euclidean vector norm. Since \( \eta_i \) is closed, the minimum distance is achieved.

The planner mobile robot $ R $ travels in the given plane with forward speed $ v $ and controlled by the angular velocity $ u $. Both the forward speed and angular velocity are limited with maximum values in $ v_{max} $ ($ v \in [0,v_{max}]  $) and $ u_{max} $ ($ u \in [-u_{max},u_{max}] $). The orientation can be described with $ \theta $ which is measured from $ x-axis $ in counterclockwise direction.
\begin{align}
	& \dot{x} = v \cos \theta \\
	& \dot{y} = v \sin \theta \\
	& \dot{\theta} = u
\end{align}
Moreover, the vehicle can determine the rate of change of this distance, denoted by \(\dot{d}_i(t)\). The centre of mass of obstacle \(i\) moves with a velocity \(v_i(t)\), which the robot knows. Moreover, these velocities \(v_i\) satisfy the following constraint:
\begin{align}
	\left\|v_i(t)\right\| \leq v < v_{max}
\end{align}
Take into account physical limitations, the minimum turning radius of the robot is given by
\begin{align}
	R_m = \dfrac{v_{max}}{u_{max}}
\end{align}
This is a very common mathematical model for non-holonomic robots that is widely used to describe the motion of wheeled ground robots as well as the planar motion of aerial drones, missiles and autonomous marine surface and underwater vehicles, see e.g.\cite{low2007biologically,manchester2004circular,goswami2018sliding,rosenfelder2021cooperative,li2021neural,kanayama1991stable,bui1994shortest}.

In addition, we assumed that the robot could detect obstacles within a certain radius. This robot model can also be applied to a variety of wheeled robots and unmanned aerial vehicles. In this paper, it should be mentioned that by abolishing negative values for $ v $, this model only considers forwards motions.

\subsection{Hybrid Navigation Method}\label{my1s3}
\begin{figure}[htbp]
	\centering
	\resizebox{0.49\textwidth}{!}{\includegraphics{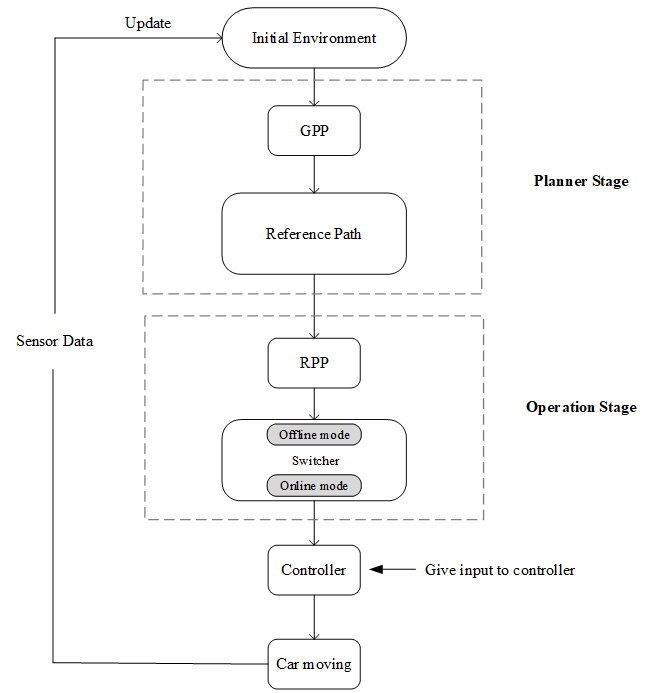}}
	\caption{Flow chart of proposed navigation method}
	\label{my1F1}
\end{figure}

The proposed safety navigation approach has two stages: planner stage and operation stage.
Planner stage utilize Global Path Planning (GPP) algorithm and an initial map based on priori knowledge of the environment. Planner stage mission aimed to find an available path based on initial environment. This finding path can be a reference path for vehicle offline moving in the beginning of operation stage. In operation stage, the car controller is consisting with offline mode and online mode. Reactive path planning (RPP) algorithm involves a switcher to decide the operation mode in each time and provides the needed inputs for vehicle’s controller to achieve collision-free in unknown dangerous or dynamic environment. Operation stage offline mode would be implemented at the first to track the reference path which obtained in planner stage and finish the avoiding unknown obstacles in online mode. Online mode is only triggered by the distance from vehicle to unknown obstacles which can be detected by sensor. Besides, operation stage can distinguish the reference path problem due to the possible situation and can trigger replan command. In the end, an assistant controller is used to generate the required input to achieve the required the linear speed and angular velocity (see Fig.~\ref{my1F1})

\subsubsection{Planner Stage}

The Rapidly exploring Random Tree (RRT) is a sampling-based search algorithm that can quickly find the collision-free path in certainty environment. The reference path generated by RRT is represented by a set of waypoints in Tree $ \varpi =  \{\varpi_1, \varpi_2, \cdots, \varpi_n$ \}.

In general, RRT algorithm generating paths are too long due to some redundant waypoints. In order to obtain the shortest path to connect the start point and end point, pruning process need to be implemented after taking the original path from RRT. The adopt pruning process method are inspirited from \cite{yang2010efficient,elmokadem2018hybrid}, which can remove the redundant waypoints effectively.

\begin{algorithm}
	\caption{Pruning Process}
	\begin{algorithmic}[1]
		\State \textbf{Input:} $\varpi=\varpi_1, \varpi_2, \ldots, \varpi_n$
		\State \textbf{Output:} $P_w=\{p_1, p_2, \ldots, p_j\}$
		\State $P_w=[]$ \Comment{Let the pruned path as an empty array}
		\State $j=n$ \Comment{Let the original path end number as the $j$ initial value}
		\State insert $\varpi_j$ into $P_w$
		\For{$i \in [1, j-1]$}
		\If{the line between point $\varpi_i$ and point $\varpi_j$ is collision-free}
		\State $j=i$
		\State insert $\varpi_j$ into $P_w$
		\EndIf
		\EndFor
		\Statex \Comment{Repeat steps 4-7 until $i=j-1$; Stop when all waypoints are checked for collision-free}
	\end{algorithmic}
\end{algorithm}

\subsubsection{Operation Stage}
\subsubsubsection{Reactive method}\label{my1_reactive}
Applying reactive method in operation stage can generates fast response to detected new obstacles until it is so safe to continue the planned path tracking. The proposed reactive method inspired by  \cite{savkin2013simple} uses the distance between the dynamic obstacles to choose a smallest angle for avoidance. This method is inspired by biology, specifically by mimicking how organisms avoid obstacles in dynamic environments. In particular, the approach draws from the ability of biological entities to achieve complex behaviours using simple local motion control rules. The theoretical foundation of this method is based on studies of animal motion control in biology, particularly the strategy of "navigating around obstacles with a constant curvature" \cite{lee1998guiding}. This strategy has been observed in various biological behaviours, such as the way squirrels run around trees. Similar obstacle avoidance strategies have been proposed and studied in the literature (see \cite{teimoori2010biologically}), primarily for dealing with static obstacles. In \cite{savkin2013simple}, this strategy was extended to the reactive navigation algorithm for dynamic environments. By leveraging simple local motion control rules inspired by biology, this algorithm can achieve effective obstacle avoidance in complex dynamic environments. The real-time navigation performance of this method was validated by Andrey and Chao through extensive computer simulations and experiments using the Pioneer P3-DX wheeled mobile robot, comparing it with the velocity obstacle method \cite{fiorini1998motion,large2005navigation}.

\begin{figure}[htbp]
	\centering
	\resizebox{0.45\textwidth}{!}{\includegraphics{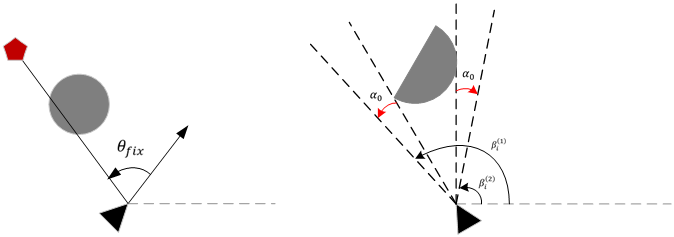}}
	\caption{Description of fixed angle and observation angle of ground robot}
	\label{my1F2}
\end{figure}

This method was separated by offline operation mode and online operation mode. The offline operation mode will operate at maximum forward speed ($ v = v_{max} $) based on the waypoints that have been programmed. It is beneficial to adopt the approach from \cite{elmokadem2018hybrid} in the path tracking section that altered the original controller design ($ u(t)=0 $) \cite{savkin2013simple} so that the vehicle can always move towards the target. Thus, offline operation mode can be described as follows:
\begin{align}
	& u(t) = u_{max}sgn \theta_{fix} \\
	& v(t) = v_{max}
\end{align}

where $ \theta_{fix} $ is the fixed angle of robot which is equivalent to heading to the target minus the current heading, see Fig.~\ref{my1F2}. This equation guarantees the heading of vehicle can always toward target during the routine operation mode.

In addition to the offline mode, the online mode, also can recognize as the obstacle avoidance mode, is activated when the mobile robot detects an obstacle within a safe distance. In our implementation, we adopt a biomimetic inspiration reactive control method which utilize sample local motion control rules to realize complex tasks \cite{savkin2013simple}. Distance detected by sensor is the only required information to implement this method which can achieve the local collision-free in dynamic situation.

When the robot is close enough to an obstacle, the reactive obstacle avoidance mode is triggered. First, the robot calculates the observation angle $ \beta_{i}^{(j)} $ based on the nearest obstacle (as shown in Fig.~~\ref{my1F2}). Observation angle $ \beta_{i}^{(1)}(t) $ and $ \beta_{i}^{(2)}(t) $  are represent the boundaries of the vision cone from the vehicle to obstacle, plus an escape angle $ \alpha_0 $.which is a given value limited in $ 0<\alpha_0<\dfrac{\pi}{2} $.

Next, the vector $ \tau_{i}^{(j)}(t) $ is introduced, which defines the two occlusion lines of an enlarged vision cone of the obstacle:

\begin{align}
	\tau_{i}^{(j)}(t) = (v_{max} - v)[\cos(\beta_{i}^{j}(t)),\sin(\beta_{i}^{j}(t))] \quad \forall j = 1,2
\end{align}

The controller can summary in followed:
\begin{align}
	& u(t) = -u_{max} f(\textcolor{red}{v}+\tau_{r}^{(j)},v_{i}) \\
	& v(t) = \| \textcolor{red}{v} + \tau_{i}^{(j)} \|
	\label{my1eq4}
\end{align}

We need to choose one of direction to achieve the smallest angle $\min _{j=1,2}\left|\varphi\left(v+\tau_i^{(j)}, v_i\right)\right|$ between itself and the velocity of the robot. $ \varphi\left(l_1, l_2\right) $ denote the angle between the vectors  $ l_1 $ and $ l_2 $ measured from $ l_1 $ in the counterclockwise direction $ \varphi\left(l_1, l_2\right) \in(-\pi, \pi] $.  Hence, we can introduce the function $ f\left(l_1,l_2\right) $.
\begin{align}
	f\left(l_1, l_2\right):=\left\{\begin{aligned}
		0, & \quad \varphi\left(l_1, l_2\right)=0 \\
		1, & \quad 0<\varphi\left(l_1, l_2\right) \leq \pi \\
		-1, & \quad -\pi<\varphi\left(l_1, l_2\right)<0
	\end{aligned}\right.
\end{align}

The observation angle $ \beta_{i}^{(j)}(t) $ and the occlusion lines $ \tau_{i}^{(j)}(t) $ can be obtained with the information of the vehicle’s position and distance between vehicle and obstacle. 
The reactive navigation algorithm utilizes the simple control method which inspirited from biology to achieve the local avoiding in dynamic environment. This approach can determine the distance between the unknow dangerous and itself to implement action. Obstacle avoidance mode will be triggered when the detected distance is too small, and vehicle will be out of the current control state until it is safe. With these conditions, the reactive navigation method can be realized to control vehicle achieve the dynamic avoiding.

\subsubsubsection{Switcher}
Totally, vehicle is maintained two modes during operating, one is for normal operation (offline mode) to ensure the robot can follow the reference path to reach the target. Second mode (online mode) would apply the reactive method to avoid the obstacle or unknow danger which detected by sensor. 
Therefore, a switcher is needed to decide the car operation mode. For subsequent description, we renamed the two modes are: routine operation mode (offline mode) $ MO_1 $ and local/reactive escape operation mode (online mode) $ MO_2 $. $ MO_1 $ aims to apply reactive method to achieve the avoid the dynamic or unknown obstacle. The control method come from below equation (\ref{my1eq4}).

Several researchers have applied mode switching to control local obstacle avoidance in previous studies. For instance, \cite{savkin2013simple} utilities distance to identify the vehicle’s current mode and \cite{elmokadem2018hybrid} considered using orientation and distance as the condition to switch controller. However, in this research, we discovered it was easy to enter a dead zone if simply the distance judgement, which introduced from \cite{savkin2013simple}. When using distance to switch from $ MO_2 $ back to $ MO_1 $, it immediately jumped back to $ MO_2 $, resulting in a dead zone. To avoid this situation, inspired by \cite{elmokadem2018hybrid}, we added a condition of the $ \theta_{fix} $, the meaning of $ \theta_{fix} $ is the angle error which has introduced in section \ref{my1_reactive}. That is, when switching from $ MO_2 $ to $ MO_1 $, using $ \theta_{fix} $ instead of the distance judgment, which represents if $ \theta_{fix} = 0 $, the switcher will be activated to $ MO_1 $. In addition, during the experiments, unexpected situations may occur in some specific cases, such as when just switching from $ MO_1 $ to $ MO_2 $, the selection of left and right angle avoidance in $ MO_2 $ may accidentally trigger $ \theta_{fix} = 0  $ condition, which may lead to avoidance fail. So as to solve this problem, we add a parameter time-switch $ \Gamma_s $ in the switcher. This parameter is always a constant value $ \Gamma_s = -1 $ in the $ MO_1 $, and when $ MO_2 $ is triggered, time-switch $ \Gamma_s $ will be initialized to zero and increased at sampling time intervals. The purpose of this is to record the avoidance time in $ MO_2 $. In the switcher, it is necessary to judge that the time-switch $ \Gamma_s $ in $ MO_2 $ cannot be less than a constant value $ K $. If time-switch $ \Gamma_s $ value is too small, escaping from $ MO_2 $ will not be activated. To summarize, the changes provided in this paper greatly improve the success rate of the algorithm, and the closed-loop mode in the switcher will also ensure the stability of the algorithm. The conditions of this Switcher mentioned above can be summarized in the following two points:\\
\textbf{C1:} Switching from mode $M O_1$ to reactive escape operation mode $\mathrm{MO}_2$ when distance to the closest obstacle reduces to the value $F$, i.e., $d_i(t)=d_{\text {safe }}$ and $\dot{d}_i(t)<0$.\\
\textbf{C2:} Switching back to routine operation mode $M O_1$ at any time $t_*$ when vehicle is oriented toward to the target, i.e., $\theta_{\text {fix }}=0$, and $\Gamma_s$ is large than a period time $K$ (i.e., $1.5 \mathrm{sec}$ ).

\subsection{Computer Simulations}
In this section, the performance of the proposed approach through combine RRT algorithm and reactive method is evaluated using computer simulations. Three simulation scenarios are performed to demonstrate the efficiency of the hybrid method. In the end, one simple case and one complex case are used to compared with proposed hybrid navigation approach and reactive strategy. All the cases are filled with several unknown moving obstacles. Black objectives represent priori stationary obstacles and grey or red objectives represent moving obstacles. In all simulations, the constant term of robot is $v_{\max }=3 \mathrm{~m} / \mathrm{s}$ and the maximum angular velocity is $u_{\max }=3 \mathrm{rad} / \mathrm{s}$. we use green line to denote the actual UAV operation path and the red dashed line is the reference path generated from planner stage. All the simulations are performed in Matlab.
\textit{Case 1,2: Dynamic environment with several moving obstacles}

\begin{figure}[htbp]
	\centering
	\begin{subfigure}[b]{0.24\textwidth} 
		\centering
		\includegraphics[width=\textwidth]{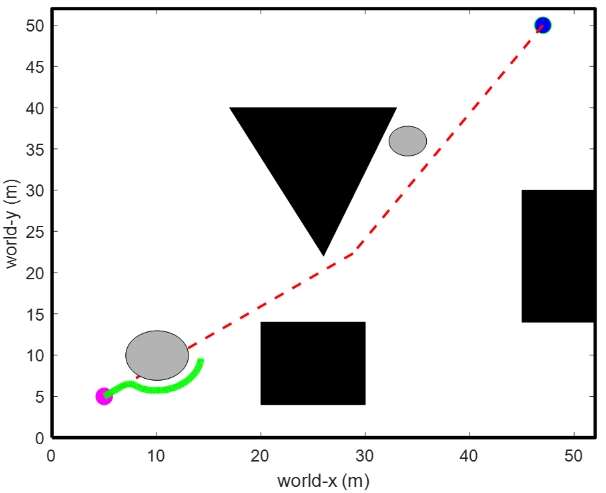}
		\caption{Initial map of case 1}
		\label{my1case1_1}
	\end{subfigure}
	\hfill 
	\begin{subfigure}[b]{0.24\textwidth} 
		\centering
		\includegraphics[width=\textwidth]{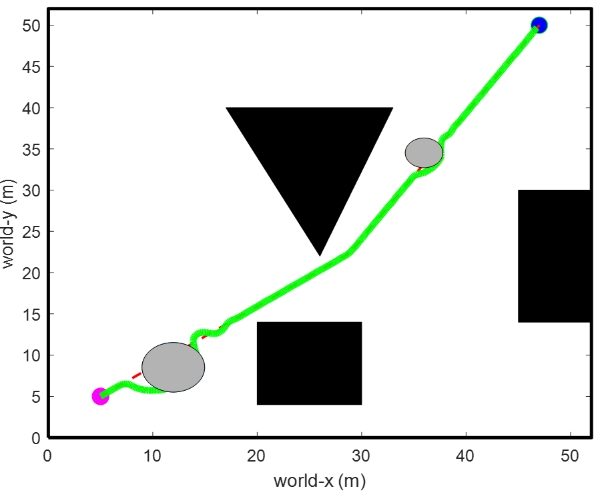}
		\caption{Result of case 1}
		\label{my1case1_2}
	\end{subfigure}
	
	\vspace{0.5cm} 
	
	\begin{subfigure}[b]{0.24\textwidth} 
		\centering
		\includegraphics[width=\textwidth]{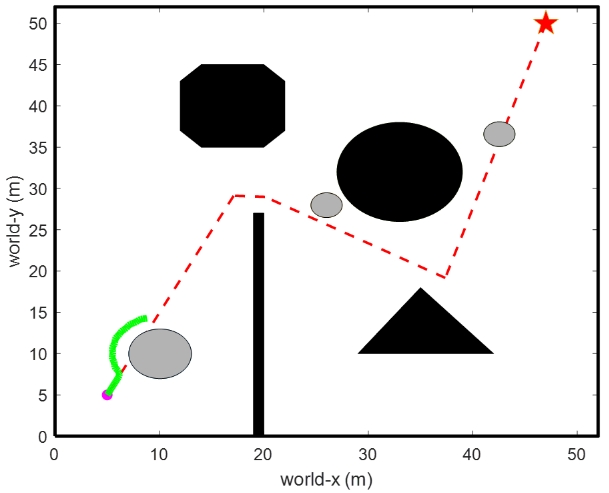}
		\caption{Initial map of case 2}
		\label{my1case1_3}
	\end{subfigure}
	\hfill
	\begin{subfigure}[b]{0.24\textwidth} 
		\centering
		\includegraphics[width=\textwidth]{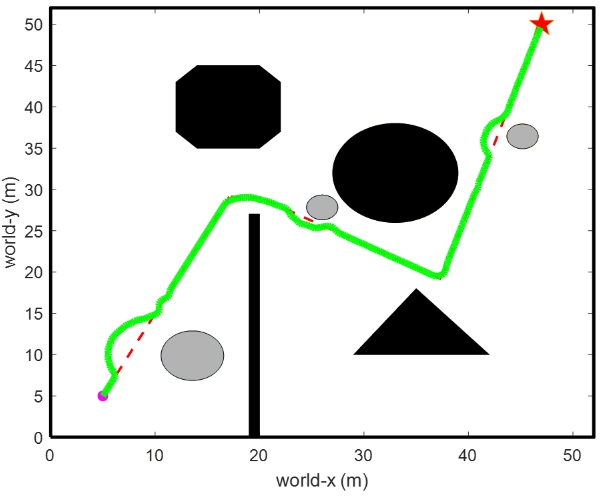}
		\caption{Result of case 2}
		\label{my1case1_4}
	\end{subfigure}
	
	\captionsetup{justification=centering}
	\caption{Simulation results of the Hybrid algorithm in dynamic environments with moving obstacles.}
	\label{my1result1}
\end{figure}

These two scenarios are navigating method that operate in the obstacles which are close to each other. Partly known environment is shown in Fig.~\ref{my1result1}. The moving obstacles $\eta_i$ velocity in case 1 are $v_1=[0.4,-0.2], v_2=[0.3,-0.3]$ respectively, and in case 2 are $v_1=[0.4,-0.3], v_2=$ $[0,-0.1], v_3=[0.4,-0.1]$ respectively. The difference between case 1 and case 2 is the location of moving obstacles, some obstacles are close to stationary obstacles or the walls (black lines) which may result in partial obstacle avoidance failure. The algorithm proposed in this section can also accomplish dynamic obstacle avoidance in the following path.

\textit{Case 3: Dynamic environment with random moving obstacles}

In this environment, the system addresses a dynamic scenario where multiple obstacles move randomly through the space at speeds less than the vehicle's maximum velocity ($ |v_i |<V_{max} $). The global planner in the hybrid algorithm guarantees the planning of effective paths in stationary environments, while the local obstacle avoidance mechanism ensures a rapid response to unknown obstacles.
\begin{figure}[htbp]
	\centering
	\begin{subfigure}[b]{0.24\textwidth}
		\includegraphics[width=\textwidth]{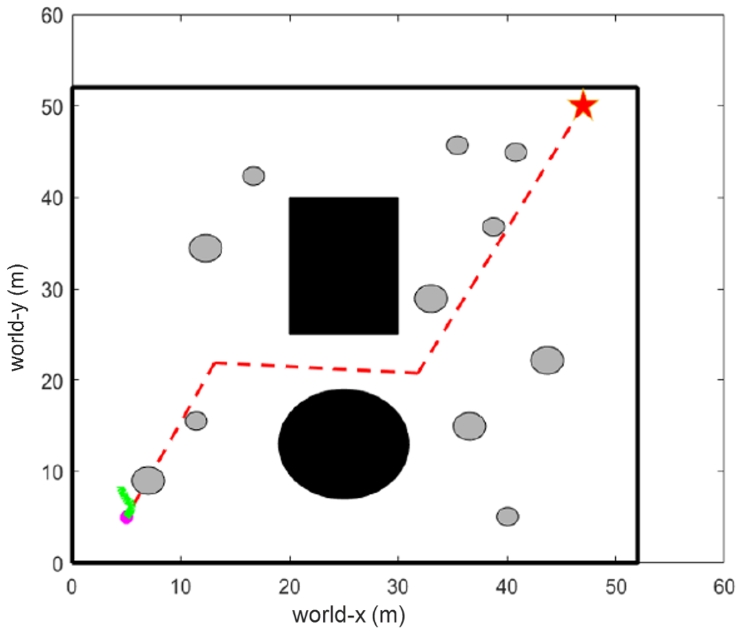}
		\caption{}
		\label{my1case31}
	\end{subfigure}
	\hfill 
	\begin{subfigure}[b]{0.24\textwidth}
		\includegraphics[width=\textwidth]{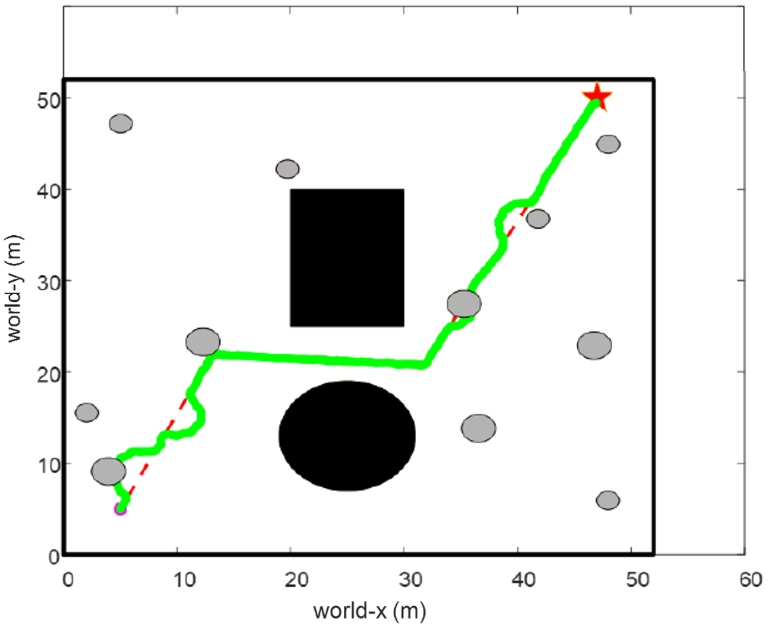}
		\caption{}
		\label{my1case32}
	\end{subfigure}
	\captionsetup{justification=centering}
	\caption{Simulation results of hybrid algorithm operated in a dynamic environment with random moving obstacles}
	\label{my1result2}
\end{figure}

\textit{Case 4: Comparison with hybrid and reactive method in two scenarios }

\begin{figure}[htbp]
	\centering
	\begin{subfigure}[b]{0.24\textwidth}
		\includegraphics[width=\textwidth]{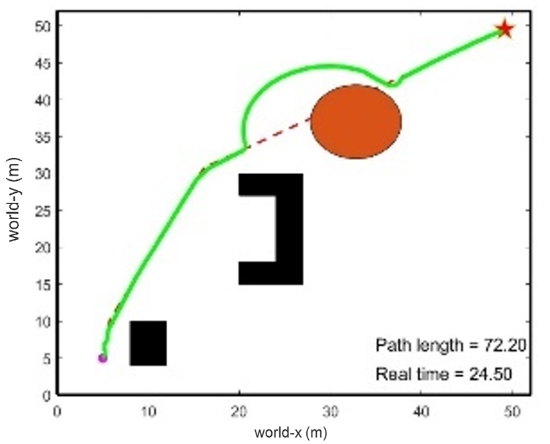}
		\caption{Hybrid method in simple \quad case}
		\label{my1case4_1}
	\end{subfigure}
	\hfill
	\begin{subfigure}[b]{0.24\textwidth}
		\includegraphics[width=\textwidth]{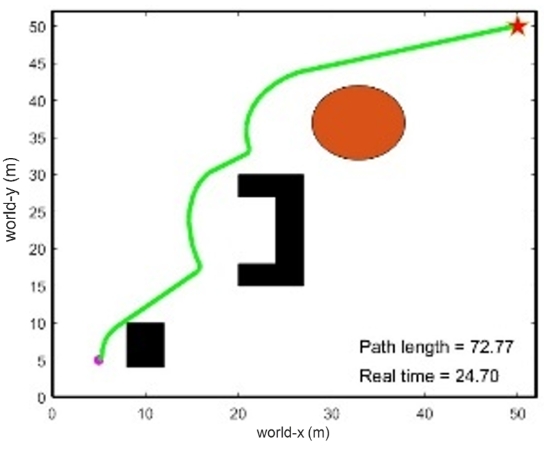}
		\caption{Reactive method in simple case}
		\label{my1case4_2}
	\end{subfigure}
	
	\vspace{0.5cm} 
	
	\begin{subfigure}[b]{0.24\textwidth}
		\includegraphics[width=\textwidth]{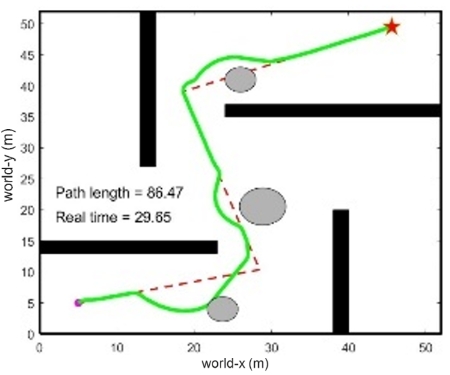}
		\caption{Hybrid method in complex case}
		\label{my1case4_3}
	\end{subfigure}
	\hfill
	\begin{subfigure}[b]{0.24\textwidth}
		\includegraphics[width=\textwidth]{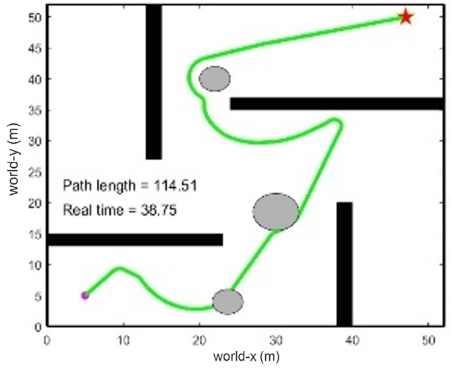}
		\caption{Reactive method in complex case}
		\label{my1case4_4}
	\end{subfigure}
	
	\captionsetup{justification=centering}
	\caption{The operations by hybrid method and reactive method in simple and complex environments\\((a) and (c) figures are hybrid method, (b) and (d) figures are reactive method)}
	\label{my1case4_main}
\end{figure}

In this case, the reactive path planning method and hybrid navigation strategy are performed in simple and complex environments, respectively, and the path length and found path speed will be used as the judging criteria for the algorithm. It is evident that the proposed method is superior and much more effective. In the simple scenario (see Fig.~\ref{my1case4_1} and Fig.~\ref{my1case4_2}), in which there is only one moving obstacle with speed $ [0.4,0] $ and a large distance between other obstacles. In this scenario, there is not a significant difference in path length and search speed between the two methods, because the complexity of the environment is relatively simple with only one moving obstacle (In reactive method, the path length is $72.77 \mathrm{~m} $ and time spent is $ 24.70 \mathrm{~sec} $. In hybrid method, the path length is $ 72.20 \mathrm{~m} $ and time spent is $ 24.50 \mathrm{~sec} $). Therefore, the hybrid method is only slightly better than the reactive method. However, for a slightly more complex environment (see Fig.~\ref{my1case4_3} and Fig.~\ref{my1case4_4}), the velocity of moving obstacles are $ v_1= [0.4,-0.3] $, $ v_2=[0.1,-0.15] $, $ v_3=[0.2,0] $ respectively). The duration time and path length in hybrid navigation method both greatly improved. (Reactive method: path length is $ 114.51 \mathrm{~m} $ and time spent is $ 38.75 \mathrm{~sec} $. Hybrid method: path length is $ 86.47 \mathrm{~m} $ and time spent is 29.65 $\mathrm{~sec}$).

\begin{figure}[htbp]
	\centering
	\begin{subfigure}[b]{0.24\textwidth}
		\includegraphics[width=\textwidth]{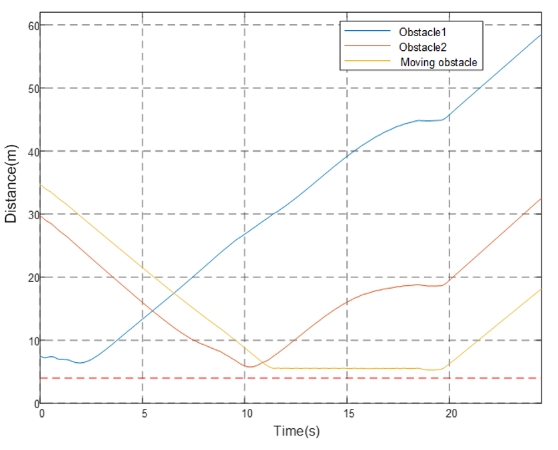}
		\caption{Hybrid method}
		\label{my1dis_1}
	\end{subfigure}
	\hfill 
	\begin{subfigure}[b]{0.24\textwidth}
		\includegraphics[width=\textwidth]{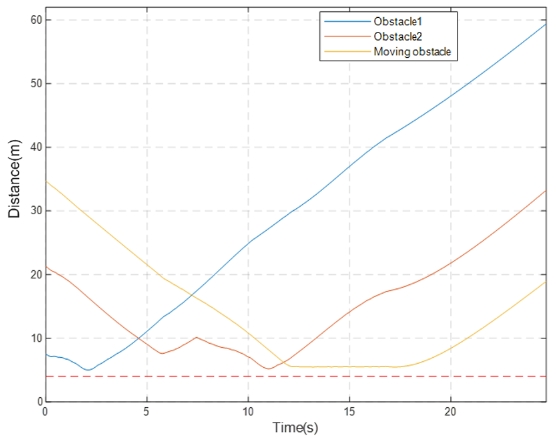}
		\caption{Reactive method}
		\label{my1dis_2}
	\end{subfigure}
	\begin{subfigure}[b]{0.24\textwidth}
		\includegraphics[width=\textwidth]{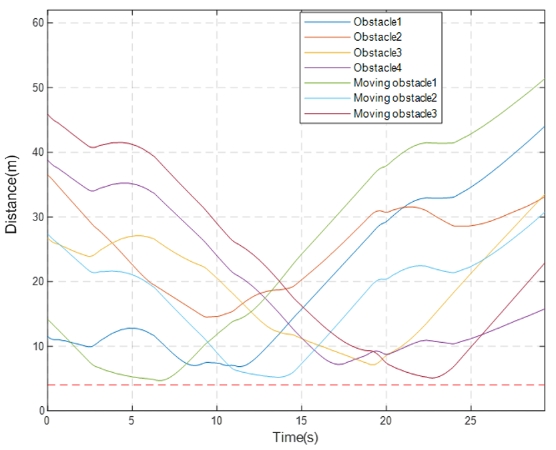}
		\caption{Hybrid method}
		\label{my1dis_3}
	\end{subfigure}
	\hfill
	\begin{subfigure}[b]{0.24\textwidth}
		\includegraphics[width=\textwidth]{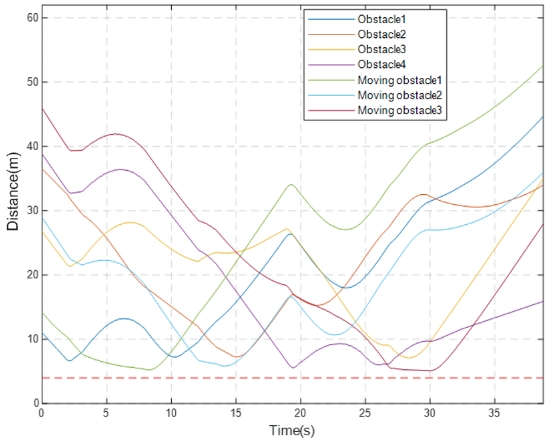}
		\caption{Reactive method}
		\label{my1dis_4}
	\end{subfigure}
	\captionsetup{justification=centering}
	\caption{The distance between each obstacles by hybrid method and reactive method in simple and complex scenarios. Figures (a) and (c) figures are hybrid method, (b) and (d) are reactive method. The red dashed line is the safety line}
	\label{my1dis_main}
\end{figure}

\begin{figure}[htbp]
	\centering
	\begin{subfigure}[b]{0.24\textwidth}
		\includegraphics[width=\textwidth]{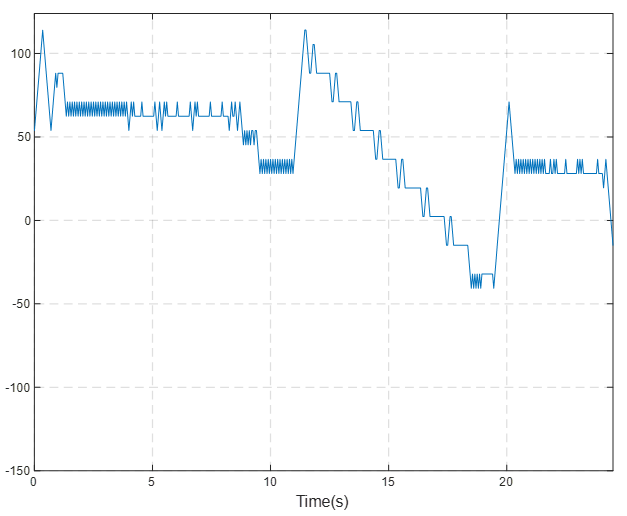}
		\caption{}
		\label{my1Ang_1}
	\end{subfigure}
	\hfill 
	\begin{subfigure}[b]{0.24\textwidth}
		\includegraphics[width=\textwidth]{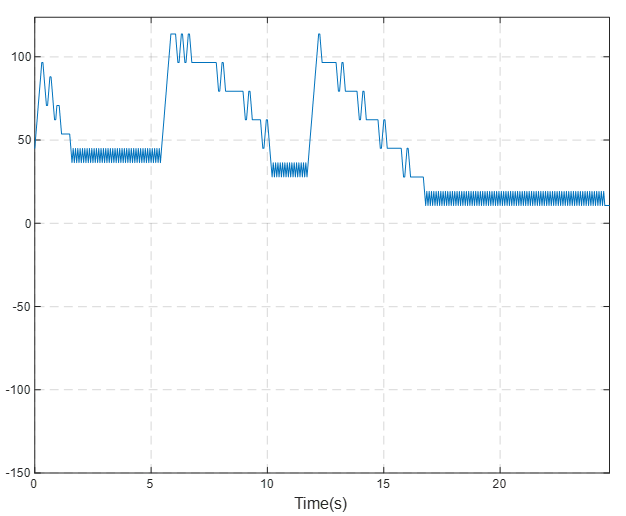}
		\caption{}
		\label{my1Ang_2}
	\end{subfigure}
	\begin{subfigure}[b]{0.24\textwidth}
		\includegraphics[width=\textwidth]{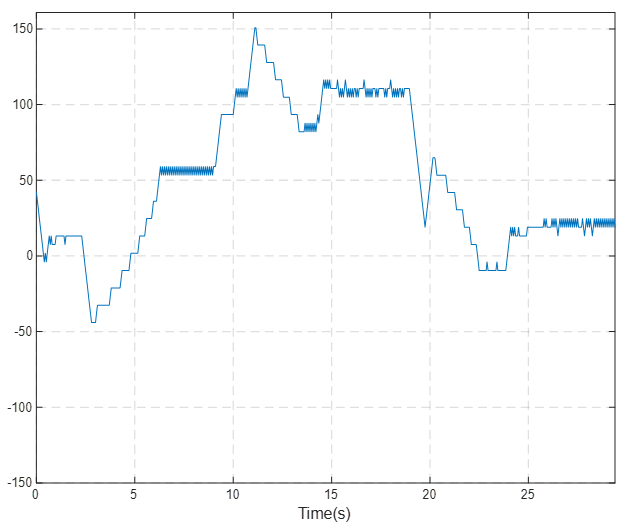}
		\caption{}
		\label{my1Ang_3}
	\end{subfigure}
	\hfill
	\begin{subfigure}[b]{0.24\textwidth}
		\includegraphics[width=\textwidth]{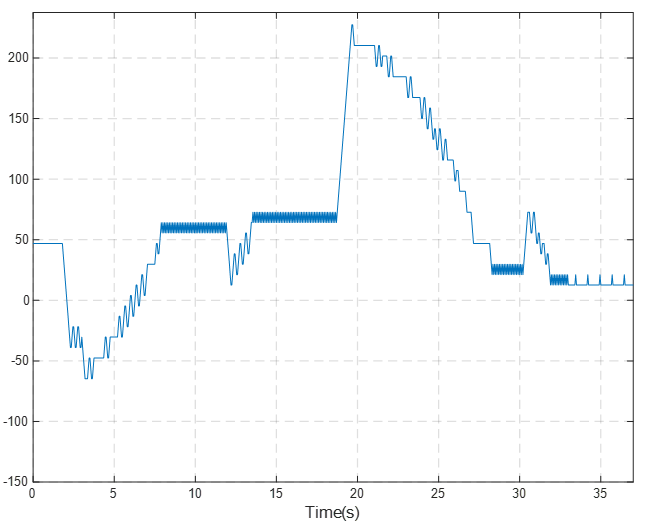}
		\caption{}
		\label{my1Ang_4}
	\end{subfigure}
	\captionsetup{justification=centering}
	\caption{The turning angle of the vehicle operated in case 4. Figures (a) and (c) figures are from hybrid method, (b) and (d) are from reactive method}
	\label{my1Ang_main}
\end{figure}

Additionally, the distance and the orientation of vehicle in these two scenarios are showing in Fig.~\ref{my1dis_main} and Fig.~\ref{my1Ang_main}.  Each figure (a,b) and (c,d) represents simple and complex scenarios respectively. Left figures come from hybrid method and right figures are from purely reactive method. In Fig.~\ref{my1dis_main}, it can be observed that, compared to simple case, the hybrid method exhibits more advantages in complex environment, where the reactive method to all obstacles is closer than hybrid method due to the lack of environmental knowledge. Fig.~\ref{my1Ang_main} shows that hybrid method in complex environments, the turning angle of the vehicle is smaller than reactive, especially the turning angle can exceed $ 200 $  degrees in reactive approach. Therefore, this section proposed collision-free navigation is a safe and energy-efficient method. 

\subsection{Conclusion}\label{my1s5}
This work presented a navigation algorithm combining global planner (RRT) and reactive planner for the safe navigation of mobile robot in dynamic environments. The approach implements the global path planning to provide the reference path in planner stage while biomimetic reactive approach is adopted to solve local obstacle avoiding in a quick manner. Besides, the proposed switcher control method improves the stability and success rate of hybrid navigation algorithm. The performance of the proposed navigation algorithm was validated using Matlab simulations with different cases. However, this section only verifies the theory with 2D regular obstacles, and in the future, this theory can be extended to 3D environments or changing the shape of obstacles, to develop collision-free method for deformable obstacles.

\section{A hybrid algorithm of UAV path planning for rescue in bushfire environments}\label{mypaper2}
As the Section \ref{mypaper1} proposed the hybrid navigation method on moving obstacles. This section introduces an enhanced algorithm designed to optimize forest fire rescue route planning. The method leverages a hierarchical structure to facilitate safe navigation, particularly in environments that are either partially known or entirely unknown. A key challenge addressed in this section is the dynamic nature of obstacles, with deformable obstacles used to simulate the spread of forest fires. Additionally, a decision-making layer integrates route replanning mechanisms to mitigate risks posed by dynamic obstacles. However, the approach focuses primarily on 2D environments.

\textbf{Some of the work is part of the paper:} \textbf{J Wei} and S Li, "A Hybrid Algorithm of UAV Path Planning for Rescue in Bushfire Environments," In 2023 IEEE International Conference on Robotics and Biomimetics (ROBIO), pp. 1-6, Samui, Thailand, Dec 2023.

\subsection{Introduction}

Safety navigation for mobile robots has became a crucial challenge, especially in unfamiliar or unpredictable terrain. The lack of prior information regarding the environments poses major difficulties in predicting navigation routes, such as the unpredictability of the movements of  obstacles on unknown terrains and the ambiguity of their shapes. Neither global nor local planning can adequately address these challenges due to the limitations of purely using one of them. Consequently, combining the assets of both planning approaches can offer a optimal solution for achieving safe navigation.

Global path planning encompasses two primary categories: search-based methods and sampling-based methods. Search-based techniques, such as A-star \cite{sathyaraj2008multiple}, Dijkstra \cite{wang2011application}, and D-star algorithms \cite{koenig2005fast}, involve traversing the entire environment to derive globally optimal paths. Notably, recent advancements have focused on optimizing these algorithms by refining heuristic functions and addressing concerns such as node redundancy and selection strategies \cite{sun2021mobile, ju2020path} . For instance, one notable study proposed a solution to the computational storage issue in A-star by intelligently expanding the domain and optimizing the selection strategy \cite{sun2021mobile}. Another study \cite{ju2020path} aimed to enhance the overall performance of A-star by prioritizing the shortest path as the optimization target. On the other hand, one sampling-based methods, exemplified by the Rapidly Exploring Random Tree (RRT) algorithm \cite{diankov2007randomized}, swiftly generate paths by exploring potential actions. RRT is particularly advantageous in complex and expansive terrains due to its relatively low computational cost. Nevertheless, it is worth noting that these navigation methods, while proficient in predefined environments, can encounter hurdles in unknown environments or in the presence of dynamic obstacles, potentially resulting in navigation failures or excessive computational demands.

Reactive navigation can generate the path planning process with collision avoidance in dynamic or unknown environment. So far, Model Predictive Control (MPC) has been widely recognized as an effective reactive approach capable of identifying the optimal trajectory by leveraging predictive models of motion  \cite{richards2006robust}. Typically, linear MPC relies on linear kinematic models utilize double integrator dynamics for more accurate forecasting. However, this method is not a common method can use for other models. In \cite{bai2019path}, the author proposes an advanced method called nonlinear MPC (NMPC), which is used for non-holonomic models to discretize the predictive model and update the state. Furthermore, certain reactive strategies rely on sensor techniques such as range sensors, video cameras, or optic flow sensors to gather precise data. Unlike global sensor-based planner methods, reactive approaches are limited to local knowledge and do not involve the construction of a comprehensive model of the environment. For example, the Dynamic Window Approach \cite{missura2019predictive,seder2005integrated}, curvature velocity \cite{simmons1996curvature} and velocity obstacles \cite{fiorini1998motion} approaches all use on-board sensors to observe a nearest obstacle of unknown terrain and update existing map information. In other words, these methods treat dynamic obstacles as static which can result in computational expensive.
Additionally, sliding mode control \cite{elmokadem2018hybrid} and boundary following approaches \cite{wang2018strategy,huang2019algorithm,matveev2011method,savkin2013simple} can also achieve the objective by detecting the distance to the nearest obstacle. Although, these distance-based methods can easy implement to deal with dynamic obstacles, they still have issues with low effectiveness and potential failure in complex or enormous environments.

To overcome the aforementioned limitations, researchers have undertaken additional investigations to develop hybrid techniques that combine both global and local methods. These strategies aim to establish a reliable path in partially known environments, effectively addressing the identified deficiencies. Some of the proposed hybrid algorithms are a combination of two improved global planning approaches. \cite{xu2020fast} combines both improved rapidly expanding random tree (RRT) algorithm and probabilistic roadmap (PRM) algorithm to separate the map, taking advantage of the fast response characteristics of the global planner to continuously update the environment to reach path planning. This approach and the method proposed in \cite{seder2005integrated} which combines D-star and DWA, have the same limitation of increasing computational effort and inefficiency. Some algorithms consider utilizing the Rapidly-Exploring Random Trees (RRT) and other reactive methods to handle the problem of navigating through both static and dynamic obstacles \cite{elmokadem2018hybrid}. In terms of global planning, the A-star method ensures route optimization but may be less efficient in finding solutions compared to the RRT algorithm. While, the DWA may lead to increased computational complexity.  
Therefore, when implementing hybrid methods, it is essential to consider the adaptability of the algorithms. 

Moreover, some hybrid strategies employ layered management systems to facilitate collaboration between global and reactive approaches. These layered management systems offer a structure for effectively incorporating the benefits of both global and reactive approaches, allowing for seamless coordination and decision-making in navigation tasks.
\cite{zhu2012new} propose the layers with the deliberative and reactive layers, both of them will be responsible for the decision making and execution of pathways respectively. \cite{niewenhuisen2016layered} plans a complete hierarchical mission, with an overall top-to-bottom increasing frequency navigation system that allows the vehicle to plan trajectories in an incomplete environmental model by implementing collision-free. In conclusion, the layered structure ensures that the hybrid algorithm operates in a more hierarchical and complete manner.	

This section presents a hybrid navigation algorithm designed to plan rescue routes in a simulated forest fire scenario using unmanned aerial vehicles (UAVs). The strategy permits rapid response, globally optimal trajectories, and the maintenance of a secure distance from the fire zone. By employing navigational planning to analyze deformable obstacles in a environment, the primary challenge posed by the fire's unpredictable spread and uneven shape is overcome. In contrast to previous research, this study takes into account moving obstacles that are not assumed to be rigid or solid and can deform, rotate, or creep in any manner. As rescue teams require coordination between ground vehicles and drones, the UAV will fly at a constant altitude. Therefore, the development of fire models is not the focus of this investigation. In this analysis, forest fires are assumed to be a deformable, wind-driven moving plane.

To ensure safe navigation, the movement of deformed obstacles needs to be taken into account. Unlike the previous movement of regular rigid objects, the uncertainty of deformation obstacles requires the addition of a safety domain around the obstacle. In previous research, \cite{matveev2012real} discussed the problem of patrolling the movement domain boundary with a given safety margin for the distance from the mobile robot to the boundary, addressing the problem of patrolling the movement and deformation boundary and considering a single robot scenario. In \cite{wang2020dynamics}, a safety area is defined around the robot so that obstacle avoidance can be performed during operation. Additional, convex optimization method is often applied to obstacle avoidance checks to achieve trajectory optimization \cite{schulman2014motion}. Influenced by the these, we will consider adding a safety plane to the deformation plane to ensure stability of the algorithm implementation.	

The content of this section is structured as follows. In section \ref{my2s2}, we present the system description and outline the navigation problem we aim to address. Section \ref{my2s3} is dedicated to introducing and discussing the navigation algorithm proposed in this study. The subsequent section presents the results of computer simulations conducted to evaluate the performance of the proposed method. Section \ref{my2s5} presents the brief conclusion.

\subsection{Problem Statement}\label{my2s2}
In this study, we consider a three-dimensional forest environment populated by a series of obstacles, both static and dynamic, denoted as $O = \{O_1, O_2, \dots, O_i\}$. These obstacles include natural elements such as rocks, lakes, caves, and bushes, which are naturally occurring within the environment. A non-holonomic mobile robot is tasked with navigating through this complex environment at a fixed altitude $h$, where $h > Z^{min}$, ensuring safe traversal while avoiding all obstacles to reach a designated target point $P_d$.

The dynamics of the robot's movement are governed by the following set of equations:
\begin{align}
	\dot{x}(t) &= v(t) \cos(\theta(t)), \\
	\dot{y}(t) &= v(t) \sin(\theta(t)), \\
	\dot{\theta}(t) &= u(t),
\end{align}
with constraints on the velocity and angular speed given by:
\begin{align}
	-V^{\max} \leq v(t) \leq V^{\max}, \quad -U^{\max} \leq u(t) \leq U^{\max}.
\end{align}
This model is typical of non-holonomic robotic systems and is widely applicable to various types of mobile machines, including wheeled robots, missiles, and underwater vehicles. Here, $v(t)$ and $u(t)$ denote the robot's linear velocity and angular velocity, respectively, while $\theta(t)$ represents the robot's heading in the horizontal plane at height $h$. The robot's position in the environment is given by the Cartesian coordinates $c_r = [x, y, h]^T$, and the velocity vector can be expressed as:
\begin{align}
	v_r(t) = \dot{c_r}(t).
\end{align}

The robot's minimum turning radius is given by:
\begin{align}
	R_{min}=\frac{V^{max}}{U^{max}}
\end{align} \label{my2eq1}

Additionally, the environment contains an unknown, moving domain $O_d(t)$. This domain is not necessarily rigid and may undergo various deformations, including stretching, curving, twisting, and relative motion with respect to other objects. For the purposes of this research, we treat this domain as a representation of a dynamically spreading forest fire. The spread of the fire is intentionally left undefined to avoid oversimplification.

The goal is to develop a navigation law that allows the robot to move safely within this environment, avoiding the spreading fire and reaching its destination. To maintain safety, the robot must keep a minimum distance $d_0$ from the boundaries of the moving domain $O_d(t)$. The distance between the robot and the nearest point on the boundary $\partial O_d(t)$ is defined as:
\begin{align}
	d(t) = \min_{C_r' \in \partial O_d(t)} \| C_r - C_r' \| \geq d_0, \quad \forall t.
\end{align}

The velocity of the domain boundary, $v_o(t)$, can be detected by the robot's sensors, which serve as the sole external information source. To ensure safe navigation, an expansion factor $\tau$ may be applied to the current boundary $\partial O_d(t)$, following the approach of previous studies. The boundary is segmented to allow for appropriate division between different regions. Furthermore, all obstacles are required to satisfy the velocity constraint $0 \leq \|v_o\| < V_{\max}$ to ensure that avoidance is feasible.

\subsection{Hybrid navigation method}\label{my2s3}
The proposed safety navigation approach consists of three layers: the observe layer, the execution layer and the decision layer. The observe layer utilizes a Global Planner with prior knowledge of the environment to generate a feasible reference path for the execute layer. Sensory perceptions provide information to the robot, allowing it to update the map as it travels through the environment. The global planner is then responsible for developing feasible pathways based on the obtain data. The created waypoints serve as the initial offline input in the execution layer, where the drone will start operating offline and follow the route if there are no unknown obstacles encountered during operation.  However, if an obstacle is detected along the current path, the decision layer takes over. It is important to note that the mobile robot has two motion modes at the execution layer: offline trajectory tracking mode and online dynamic obstacle avoidance mode (reactive control). The mode of operation is determined by the decision layer, while the execution layer is only affected by input parameters. The decision layer determines the movement mode based on the current distance and orientation angle.

Moreover, an additional rule will be incorporated into the decision-making process to facilitate re-planning. This replan approach effectively prevents the navigation algorithm from getting stuck in a dead loop or experiencing algorithm failures. During this stage, the current position serves as the starting point for conducting a new global path planning. This additional measure ensures the integrity of the algorithm by addressing the limitations of reactive methods, which may lead to the inability to find a path or result in an incorrect path. By implementing this hierarchical management system, the navigation tasks are separated, and a path is generated that is optimized in terms of distance, time, and adherence to specified constraints. The workflow described above can be visualized in the following figure:
\begin{figure}[h]
	\centering
	\includegraphics[width=0.5\textwidth]{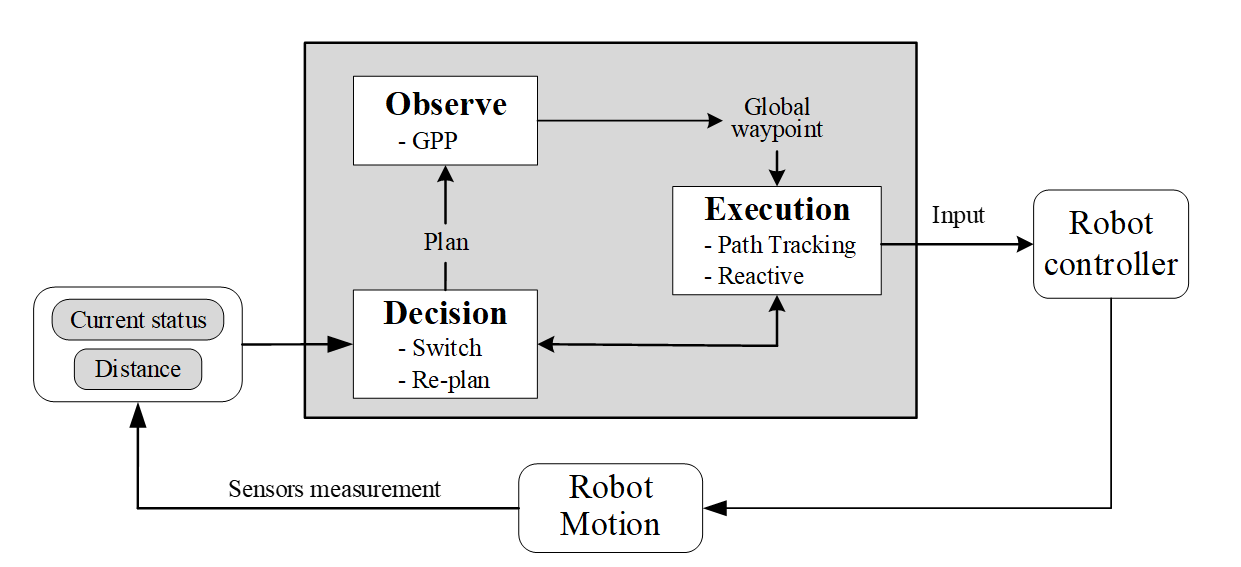}
	\caption{The principle of hybrid navigation strategy with hierarchical structure}
	\label{my2fig2}
\end{figure}
The proposed flowchart illustrates the robot's current state update process and the navigation method (represented by the layered structure in the gray blocks). This method provides inputs to the robot controller, which drives the robot's motion. Meanwhile, the robot's sensors provide new detection information at each time step to update the Robot Status. Within the layered structure, we introduce three layers: the Obstacle Layer, the Execution Layer, and the Decision Layer. These three layers interact and influence each other. Initially, the Robot Status module sends current information, such as the distance to obstacles, to the Decision Layer. The Decision Layer then determines whether to switch modes (Switch) or to re-plan (Re-plan). If the decision is to switch, the navigation mode in the Execution Layer will be updated accordingly. If re-planning is required, the system will use Global Path Planning within the Observe module to re-plan the reference path and output the Global Waypoint to the Execution Layer.

\subsubsection{Observe Layer}
The observe layer of the proposed safety navigation approach is primarily composed of the global path planning algorithm. We have opted for the Rapidly Exploring Random Tree (RRT) method among the available algorithms due to its time efficiency. RRT is a sampling-based search algorithm designed to swiftly identify collision-free paths in a certain environment. Initially, a tree space is established to store path nodes, with the starting point serving as the parent node in the tree. To ensure efficiency and accuracy, samples are generated randomly, and each point is compared with existing nodes to determine the nearest one. A collision-free path is identified if no obstacles are encountered by creating new nodes along the line connecting the nearest point and the random point at fixed step intervals. A set of waypoints within the tree $ \varpi = \varpi_{1}, \varpi_{2}, ,\cdot\cdot\cdot, \varpi_{n} $ represents the resulting reference path.

\begin{figure}[h]
	\centering 
	\includegraphics[width=.48\textwidth]{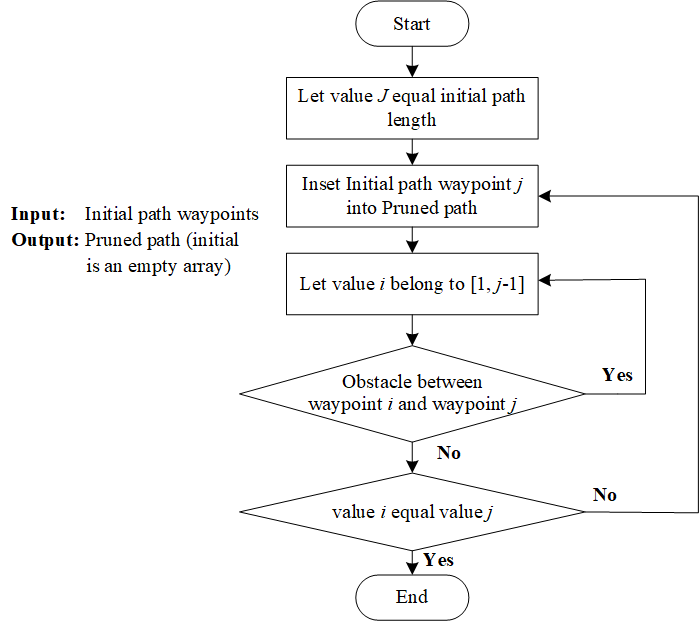}
	\caption{Pruning progress algorithm}
	\label{my2fig3}
\end{figure}

Considering that the reference path obtained from the previous step may contain redundant waypoints and may not represent the optimal route, we employ a pruning process method, as described in \cite{yang2010efficient}, to eliminate these redundancies and achieve the shortest path between the start point and the end point. The pruning process method can be summarized as follows (refer to Fig.~\ref{my2fig3}):

Once a pruned path is obtained, the task of the observe layer is considered complete. The pruned path, now serving as the reference path, is passed on to the execution layer, which is responsible for executing path tracking methods. The observe layer's role is to generate a viable global path by utilizing map information obtained from sensory measurements.

\subsubsection{Execution Layer}
Since this section is an extension of section \ref{my1s3}, the navigation modes used here are consistent with those discussed earlier. However, considering the impact of these modes on the robot model, we have rewritten and retained this section to ensure clarity and continuity.

\subsubsubsection{Path tracking mode} 
The path tracking execution mode requires the UAV to precisely follow the planned path until an obstacle is encountered, which can trigger online reactive mode. The distance to the unknown obstacle becomes the sole determining factor during the operation. In order to ensure accurate tracking, we utilize the orientation error fixed approach from \cite{elmokadem2018hybrid}, which ensures that the robot's motion consistently aligns with the next waypoint $\varpi_{n} $ on the planned path. Consequently, we evaluate the following pure path tracking algorithm:
\begin{align}
	\begin{array}{l}
		u(t) = U^{max}sgn(\theta_{fix}) \\
		v(t) = V^{max}
	\end{array}   \label{my2mode1}
\end{align}
In this described pure path tracking algorithm, the orientation error correction, denoted as $ \theta_{fix} $, is calculated as the absolute difference between $ \theta_{H} $, which represents the orientation from the current position to the next waypoint, and $ \theta_{v} $, derived from the model described in equation (\ref{my2mode1}).  $ sgn(\cdotp) $ denotes the standard $ sign $ function:
\begin{align}
	sgn(x) = 
	\left\{\begin{array}{ll}
		1, & if \quad x >0 \\
		0, & if \quad x = 0 \\
		-1, & if \quad x <0
	\end{array}\right.
\end{align}

\subsubsubsection{Reactive mode} 
In addition to offline mode, our approach includes an online obstacle avoidance mode. Inspired by biological systems, we employ a reactive control system based on local motion control principles \cite{savkin2013simple}. This approach utilizes distance information to enable collision-free motion around moving objects by identifying the smallest turning angle required for avoidance. 

The distance detected by the sensor is the sole necessary information to implement this method, enabling local collision-free navigation in dynamic situations. The controller can be summarized as follows:
\begin{align}
	\begin{array}{l}
		u(t) = -U^{max}f((v_{o}(t)+e_{o}^{i}(t)),v_{r}(t)) \\
		v(t) = V^{max}
	\end{array} \label{my2mode2}
\end{align}
where the vector $ v_{o}(t)+e_{o}^{i}(t), i = 1,2 $ represents the two directions to avoid the dynamic obstacle. It combines the velocity vector of the obstacle ($ v_{o}(t) $) with the occlusion lines $ e_{o}^i(t) $ to determine the avoidance directions. The occlusion lines $ e_{o}^i(t) $ can be calculated with the observation angle $ \alpha_{o}^i(t) $ and enlarged avoiding angle $ \beta_{o}^i(t), i = 1,2 $ : 
\begin{align}
	\left\{\begin{array}{l}
		\beta_{o}^i(t) = \alpha_{o}^i(t) \pm \alpha_{safe} , \\
		\angle l_1(t) =  \cos(\beta_{o}^i(t)), \\
		\angle l_2(t) =  \sin(\beta_{o}^i(t)), \\
		\bigtriangleup V = V^{max} - ||v_{o}(t)||, \\
		e_o^i = \bigtriangleup V[\angle l_1(t), \angle l_2(t)].
	\end{array}\right. 
\end{align}

To provide clarity, the enlarged avoiding angle $ \beta_{o}^i(t) $ is obtained by adding the observation angle $ \alpha_{o}^i(t) $ and the constant parameter $ \alpha_{safe} $ ($ 0< \alpha_{safe} < \frac{\pi}{2}$). The vector $ e_{o}^i(t) $ can  represent the boundaries of the vision cone from the vehicle to the obstacle, ensuring that the avoidance directions remain within this cone.

Here we introduce one function $ \varphi (\angle1,\angle2) $, which calculates the angle from vector 1 to vector 2 in the counter-clockwise direction. The index of the smallest angle $ \varphi $ is then used as an input to the function $ f(\cdotp,\cdotp) $.
\begin{align}
	Index = \min_{i = 1,2} \left| \varphi(\angle l_1(t), \angle l_2(t))
	\right|
\end{align}
Let $ N \in [1,2] $ represent the chosen index number. We can now introduce the function $ f(\cdot,\cdot) $.
\begin{align}
	f(l_1,l_2) = 
	\left\{\begin{array}{ll}
		0, & if \quad \varphi(l_1,l_2) =0 \\
		1, & if \quad 0< \varphi(l_1,l_2) \leq \pi \\
		-1, & if \quad -\pi < \varphi(l_1,l_2) < 0
	\end{array}\right.
\end{align}

The reactive navigation algorithm utilizes a biologically-inspired simple control method to achieve local obstacle avoidance in dynamic environments. This approach determines the distance between the vehicle and detected obstacles to take appropriate actions. When the detected distance becomes too small, the obstacle avoidance mode is triggered, and the vehicle transitions out of its current control state until it is safe to proceed. By incorporating these conditions, the reactive navigation method enables the vehicle to dynamically avoid obstacles.

\subsubsection{Decision Layer}
The UAV operates in two modes during operation: the routine operation mode (offline mode) denoted as $MO_1$, and the local/reactive escape operation mode (online mode) denoted as $MO_2$. In the path tracking mode ($MO_1$), the vehicle implements the control method described by equation (\ref{my2mode1}). On the other hand, $MO_2$ is designed to apply a reactive method to avoid dynamic or unknown obstacles, with the control method derived from equation (\ref{my2mode2}).

In the decision layer, the distance to the closest obstacle obtained from the sensors serves as the sole external criterion for decision-making. As the current distance $d(t)$ reduces to $ d_0 $, the UAV will trigger a switch from $MO_1$ to $MO_2$. The UAV will switch back from $MO_2$ to $MO_1$ only when the UAV's orientation is directed towards the next target point.

To ensure proper operation of $MO_2$ in avoiding moving obstacles, an additional time-observer condition, denoted as $T_{ob}$, should be introduced. During $MO_1$, $T_{ob}$ remains constant at -1. However, when transitioning to $MO_2$, $T_{ob}$ is reset to 0, and each time obstacle avoidance is performed, it is incremented. This time condition ensures that the mobile robot can operate $MO_2$ effectively to avoid moving obstacles. Consequently, $T_{ob}$ should have a periodic value to prevent erroneous switching back to $MO_1$ due to angular errors during $MO_2$.

In the navigation algorithm, we also introduce an additional re-plan rule, which is triggered only in $MO_1$. The purpose of this re-plan rule is to address inefficiencies that may arise in complex and highly dynamic environments. By employing a re-plan method, we can avoid excessive computational load and ensure quick responses to perceived obstacles.

To determine the need for re-planning, we stipulate that if the UAV's turning radius does not satisfy the condition specified in Equation (\ref{my2eq1}), the re-planning stage will be initiated. This criterion helps optimize the navigation process by allowing the UAV to adapt its path and avoid potential obstacles more effectively.

To sum up, the decision layer can be summarized as follow:

\textbf{DL1:} Switching from mode $MO_1$ to reactive mode $MO_2$ when distance $d(t)$ reduces to the value F, i.e., $ d(t) = d_0$ and $ \dot{d(t)} < 0 $.

\textbf{DL2:} Switching back to routine operation mode $ MO_1 $  at any time $ t^{*}$ when vehicle is oriented toward to the target, i.e., $ \theta_{fix}=0 $, and $ T_{ob} $ is large than a period time (i.e., $ 1.5 sec $).

\textbf{DL3:} In mode $MO_2$, replan is activated if the turning radius $ R < R_{min}  $.

\subsection{Computer Simulations}\label{my2s4}
In this section, we conduct computer simulations to evaluate the performance of the proposed approach that combines the RRT algorithm and reactive method. The simulations are carried out in a forest fire scenario where different terrain features such as rock caves, a lake, and bushes are represented by grey, blue, and green figures, respectively. This information is prior known and provided in the map. The initial red region indicates the location of the fire. The starting point represents the entrance for the rescue vehicle, and a feasible rescue path  is generated using the RRT algorithm. Once the mobile robot starts moving, it initially follows the predefined path (red dashed line) while the sensors continuously measure the detection distance of the surroundings. It is noticed that the UAV operation line denotes as green line. The decision layer utilizes this information to determine the current mode of operation. Replanning is incorporated to prevent potential path search failures. To highlight the advantages of the hybrid algorithm, we compare its performance to a purely reactive approach. The simulations consider parameters such as maximum linear velocity, angular speed, safe scape angle, sample time, and obstacle speeds. Specifically, the maximum linear velocity is  $ V^{max} =  4 m/s $, the maximum angular speed is $ U^{max} = 2 rad/s $, the safe scape angle $ \alpha_{safe} $ is $\frac{\pi}{5}$, and the sample time is 0.05 seconds. The obstacle speeds are defined as initially spreading in all directions with speeds ranging from $ 1.1 m/s $ to $ 1.2 m/s $, and then spreading downwards at a speed of $ 2.2 m/s $. The drone is capable of real-time detection of the surface speed of these deforming obstacles to perform local obstacle avoidance. The red area represents the spreading fire, which is a deformable obstacle that expands randomly along the \(xy\) axes at a variable speed (with a maximum speed less than \(V^{max}\)). All the initial parameters in Table~\ref{my2table} and all simulations are implemented in MATLAB.

\begin{table}[h]
	\captionsetup{labelfont={color=black}, textfont={color=black}} 
	\caption{\textcolor{black}{Setting of Parameters for Simulation}}
	\label{my2table}
	\begin{center}
		\begin{tabular}{c||c|c}
			\hline
			\multicolumn{1}{c||}{case/parameters} & Hybrid method & Reactive method \\			
			\hline
			$ V^{\text{max}}(\, \mathrm{m/s}) $ & 4 & 4\\
			$ U^{\text{max}} (\, \mathrm{rad/s}) $ & 2 & 2\\
			$ \alpha_{\text{safe}} $ & $ \frac{\pi}{5} $ & $ \frac{\pi}{5} $\\[0.5ex]
			$ d_{0} (\, \mathrm{m})$ & 4.5 & 4.5 \\
			$T_{sample}( \, \mathrm{sec} $) & 0.05 & 0.05 \\
			\hline
		\end{tabular}
	\end{center}
\end{table}

\begin{figure}[h]
	\centering 
	\includegraphics[width=0.48\textwidth]{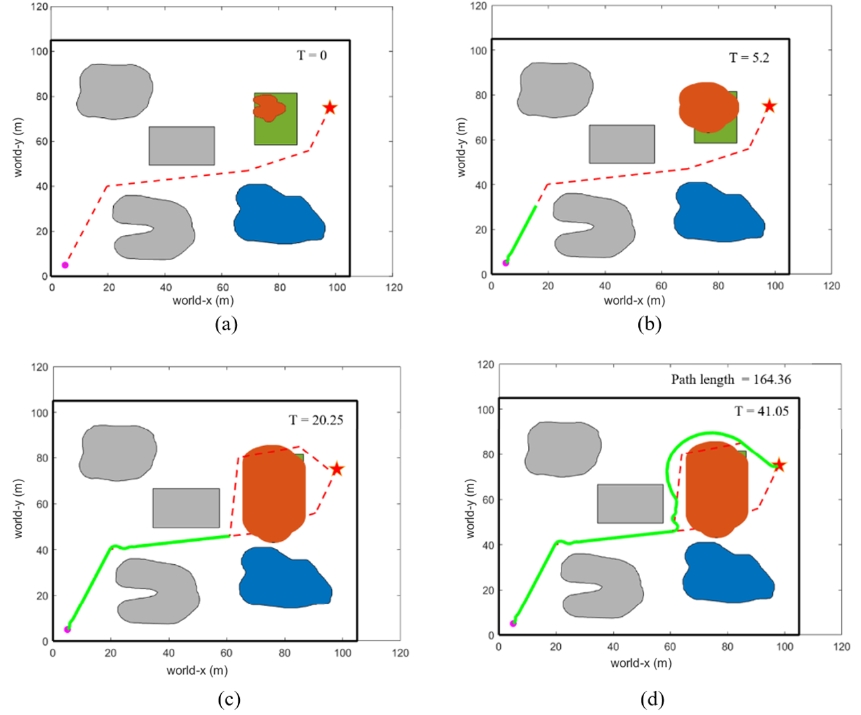}
	\caption{UAV navigated by the hybrid algorithm operates in a dynamic environment, red objective denotes the deformable obstacle and others are stationary obstacles. (a) is the initial map with reference path from observe layer, the UAV actual path denote with green line as can be seen in (b). In the time of 20.25 sec in (c), the replan rule is trrigerd. (d) represents the ultimate status of UAV}
	\label{my2fig4}
\end{figure}

\begin{figure}[h]
	\centering 
	\includegraphics[width=0.48\textwidth]{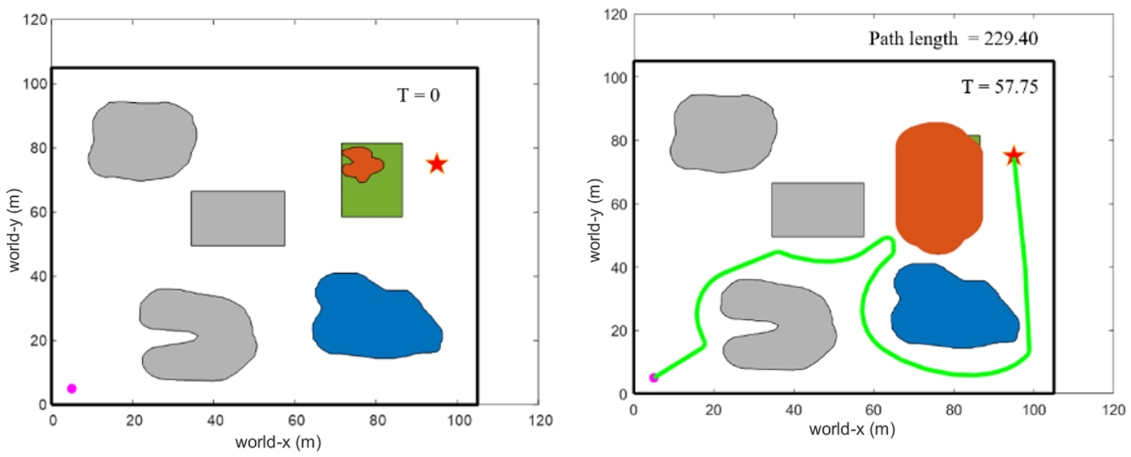}
	\caption{UAV navigated by the purely reactive algorithm operates in a dynamic environment}
	\label{my2fig5}
\end{figure}

Fig.~\ref{my2fig4} displays the simulation results using the hybrid algorithm proposed in this section. In Fig.~\ref{my2fig4} (a) and (b), the drone executes in routine operation mode $MO_1$, following a predefined path. However, when an obstacle is encountered and the detection angle exceeds a certain threshold (shown in Fig.~\ref{my2fig4}(c)), replanning is triggered.   The RRT algorithm is recalculated, and the drone follows a different path. Since the red obstacle represents a dynamic and unknown fire, the newly planned path may approach the fire area.
As a result, the local obstacle avoidance mode $MO_2$ is activated along the subsequent path.
The final path has a length of $ 164.36 m $ and it takes  41.05 seconds  to complete. Compared to a purely reactive approach (see in Fig.~\ref{my2fig5}), the proposed algorithm significantly improved the path length and overall performance. In reactive approach, the lack of prior knowledge of the entire map causes the drone to choose smaller angles for obstacle avoidance, resulting in a less efficient path.

\begin{figure}[h]
	\centering 
	\includegraphics[width=.48\textwidth]{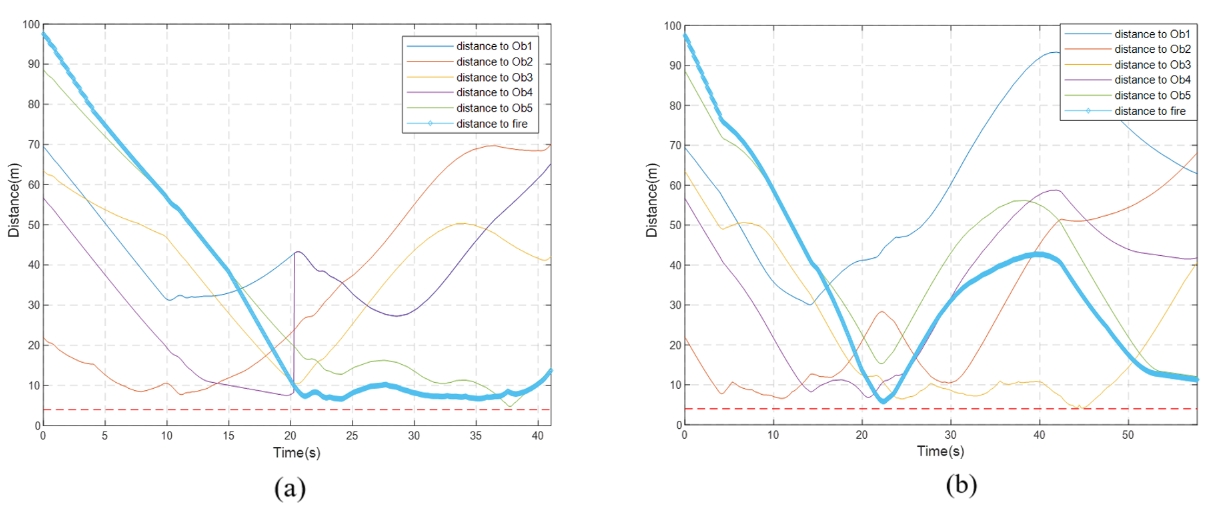}
	\caption{Distance to obstacles, (a) is UAV operated by hybrid method, (b) is UAV operated by reactive method}
	\label{my2fig6}
\end{figure}

\begin{figure}[h]
	\centering 
	\includegraphics[width=0.48\textwidth]{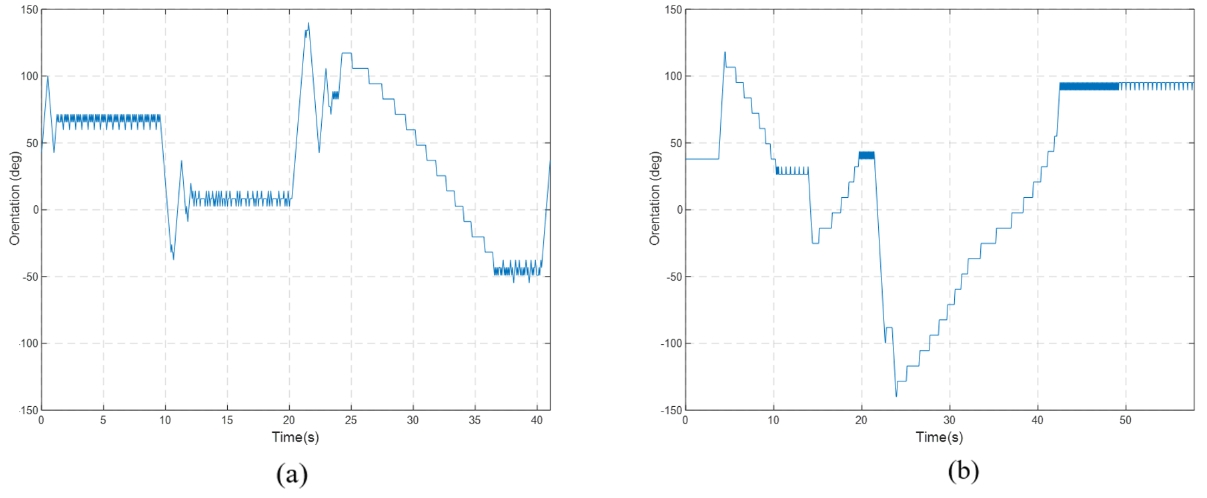}
	\caption{Orientation of UAV during the simulation (a) is UAV operated by hybrid method, (b) is UAV operated by reactive method}
	\label{my2fig7}
\end{figure}

In Fig.~\ref{my2fig6} and Fig.~\ref{my2fig7}, A comparison is performed between the distances of the UAV from each obstacle and the UAV's turning angle during simulation for the two methods. The proposed method in this section ensures a safe distance is maintained from all static obstacles at the start, while replanning in the decision layer ensures that the drone does not need to use excessively large turning angles during operation, thereby reducing energy consumption.

\subsection{Conclusion}\label{my2s5}
This research presented a navigation algorithm combining a global planner (RRT) and reactive planner for the mobile robot's safe navigation in a simulated forest fire scenario. The global pre-defined path generated by RRT is applied in execute layer. To handle local obstacles and ensure a quick response, a boundary-following reactive method is employed. The proposed navigation algorithm's performance is evaluated through Matlab simulations, using specific cases tailored to the forest fire scenario. These simulations serve to validate the effectiveness and efficiency of the algorithm in navigating through the simulated environment.

\section{A Method for Collision-free  UAV Navigation around Moving Obstacles over an  Uneven Terrain}\label{mypaper3}
We concentrate on the obstacle avoidance algorithm in a two-dimensional environment in section \ref{mypaper1} and \ref{mypaper2}. The section further extends the approach into 3D environment. We present a novel 3D reactive navigation method for UAV in uneven terrain with unknown moving obstacles. This proposed method effectively reduces computational burden while ensuring safe and efficient UAV operations in complex environments. 

\textbf{Some of the work is part of the paper:} \textbf{J Wei} and S Li, "A Method for Collision-free UAV Navigation around Moving Obstacles over an Uneven Terrain." In 2023 IEEE International Conference on Robotics and Biomimetics (ROBIO), pp. 1-6, Samui, Thailand, Dec 2023.

\subsection{Introduction}
In recent years, unmanned aerial vehicles (UAVs) have emerged as cost-effective and highly maneuverable tools for various applications, such as environmental surveillance, target localization, way-point tracking, and wireless sensor networks. These applications not only mitigate the risks associated with complex tasks performed by humans but also significantly enhance efficiency. However, one notable challenge in these tasks involves achieving automated safe navigation to prevent collisions, particularly in environments characterized by unknown tunnels, bushfires, or uneven terrains. Consequently, the development of collision avoidance techniques for unknown or moving obstacles holds great promise. An important hindrance to achieving collision-free path planning lies in the lack of prior knowledge, which results in unreliable cooperative communication between UAVs.

The aforementioned article explores classical techniques for comprehensive environmental analysis through data information. 
The sample-based method and the travel-based method are two separate types of path planning.  However, this type method has drawbacks of high computational cost and inefficiency in partially known environment.  \cite{zhang2020novel} proposes a dynamically enhanced A-star algorithm, which has been applied in complex dynamic three-dimensional environments. However, this method still suffers from the issues present in the traditional A-star algorithm, such as excessive computational complexity that hinders quick path planning.  An algorithm combining probabilistic roadmap (PBM) and D*lite has been proposed to safely navigate through obstacle-rich environments \cite{hrabar20083d}. This method reduces planning time but does not yield the optimal path. Due to the stochastic nature of the probabilistic roadmap approach, the generated path incurs certain costs in terms of searching for a viable route. The RRT* algorithm can rewire nodes to optimize the path cost towards a specific node, resulting in a path that approximates the optimal theoretical path. However, RRT* is not the most flexible method for application in dynamic environments. \cite{mohammed2021rrt} proposes an algorithm that combines the properties of RRT* and is applicable to 3D environments. It first generates a primary path from the start point to the goal point and then extends this path by adding intermediate points to navigate around encountered obstacles. This method is not suitable for complex environments and tends to generate paths that do not adhere to the non-holonomic property, requiring further path optimization.

However, an alternative method that does not rely on prior knowledge is the reactive path planning algorithm. This approach, also known as behavior-based control or map-less algorithm, generates motion plans based on sensory input and desired objectives. Unlike traditional methods that depend on precomputed maps and extensive exploration, reactive path planning focuses on real-time responsiveness to the environment during task execution. Reactive motion planning can generally be classified into velocity-based, boundary-based and sensor-based methods. As a classic curvature control algorithm, \cite{simmons1996curvature} presents a local obstacle avoidance algorithm with velocity-space constraints. This algorithm enables robots to achieve safe and efficient navigation by incorporating physical limitations and environmental constraints. Based on this foundation, the Dynamic Window Approach (DWA) algorithm has been improved. By analyzing the motion model and velocity constraints at the current moment, DWA identifies a velocity window that best aligns with the desired velocity for the next time step \cite{chang2021reinforcement}. In \cite{li2017obstacle}, the author proposes an enhanced dynamic window approach that addresses the problem of local minima and accomplishes an optimal path by analysing the relationship between the free space and obstacles surrounding the mobile robot.  Furthermore, one sliding mode control method proposed in \cite{matveev2011method} can be applied to boundary patrolling and obstacle avoidance problems. It belongs to the category of boundary-based dynamic obstacle avoidance algorithms, similar to the approaches presented in \cite{savkin2013simple} and \cite{wang2018strategy}. In addition to that, some methods have been developed to achieve real-time obstacle avoidance in human-robot interaction scenarios by leveraging sensors and model predictions.  The research presents in \cite{huang2019algorithm} proposed an innovative algorithm for the 3-D arrangement of interconnected UAVs, emphasizing proactive collision prevention to enhance surveillance efficiency. This approach harnesses instantaneous data for adaptive maneuvering of UAVs in a three-dimensional environment, maximizing observation scope while circumventing barriers. Besides, the integration of sensor data and advanced planning algorithms allows \cite{ullah2012sensor} and \cite{ullah2011sensor} to generate paths without collision. This combination of sensor fusion and intelligent path planning enables these approaches to tackle the challenges of navigation in positional environments effectively.

The purpose of this section is to address the difficulty of secure obstacle avoidance navigation in the 3D environment. To address this issue, a reactive obstacle avoidance algorithm for nonholonomic mobile robots is proposed. The most important benefit of this method is its low computational cost, as only the distance to the adjacent obstacle is required for quick analysis and calculation. In contrast to conventional methods, this real-time navigation approach generates efficient obstacle avoidance trajectories without the need for map updates.

This section is structured as follows. Section \ref{my3s2} formulates the problem of 3D navigation in an uneven environment. The suggested navigation strategy is then proposed with a thorough analysis in section \ref{my3s3}. This strategy is then validated using computer simulation in section \ref{my3s4}. In section \ref{my3s5}, we will provide a brief summary.	

\subsection{Problem Formulation}\label{my3s2}
\subsubsection{System Model}
Consider a typical navigation problem where  a single UAV, denoted as $ E $, needs to navigate safely through an uneven terrain represented by the three-dimensional space $ \mathbb{R}^3 $. This environment contains various unidentified hills $ \chi_{terrain} $, as well as other unknown flying obstacles, represented by $ \chi = \{\chi_{1},\chi_{2},\dots,\chi_{n}\} $. These obstacles and uneven terrain can be considered as unforeseen elements within the environment. The primary objective is to guide a UAV securely to a stationary target area $ e_{target}\subset \mathbb{R}^3$, while ensuring obstacle avoidance. 

The drone $ E $ commences route planning mission with an initial map of the environment containing only the boundary $ B \subset \mathbb{R}^3$, starting location $ e_{0}\subset \mathbb{R}^3 $, and target location  $ e_{target} \subset \mathbb{R}^3 $. As the drone traverses the map, any encountered obstacles $ \chi_{n} $ and uneven terrain $ \chi_{terrain} $  are treated as unknown knowledge. It should be noticed that these coordinates are defined relative to a ground coordinate frame. 	In this dissertation, the term $F_{g}$ refers to the ground coordinate frame, while $F_{u}$ represents the UAV coordinate frame, also known as the body coordinate frame. Let $ \boldsymbol{e} (t)   := [x(t),y(t),z(t)]^T$ denotes the UAV $ E $ Cartesian coordinates in frame $ F_{g} $. Now we can summarise the  explanation of the nonholonomic kinematic model as follows, based on the given information:
\begin{align}
	& \dot{e}(t) = V(t) \widetilde a (t), \\
	&\dot{\widetilde a}(t) = u(t),  \\
	&(\widetilde a (t),u(t)) = 0 \label{my3robotmodel}
\end{align}
Here,  $ \widetilde a (t) \in \mathbb{R}^3 $ represents the orientation of the vehicle $ E $. The scalar product $(\cdot,\cdot) $ denotes that the direction vector should always be perpendicular to the control input $ u(t) $. The orientation vector is required to maintain a normal length of $ ||\widetilde a (t)|| = 1 $. For all time $ t$, UAV's velocity $ V(t) \in \mathbb{R}^3$ and angular speed $ u(t) \in \mathbb{R}^3 $ are considered as control inputs to the model. Both of these inputs must adhere to the following constraints:
\begin{align}
	&-U_{max} \leqslant u(t) \leqslant U_{max}, \\
	& 0 \leqslant V(t) \leqslant V_{max}
\end{align}
Additionally, the minimum turning radius of the UAV is given by
\begin{align}
	R_{min} = \dfrac{V_{max}}{U_{max}} 
\end{align}
We assume that the UAV $ E $ is equipped with sensors that provide visibility of the obstacle portions within its sensor range $ R_{sensor} $. Therefore, it can provide the current shortest Euclidean distance  $ d(t) $  to the closest obstacle. Moreover, to ensure the safety navigation, we need determine a safe margin $ d_{\varepsilon} > 0 $ from boundary of unknown obstacles according to the following:
\begin{align}
	d(t) :=  \min \limits_{e'(t) \in \chi} || \boldsymbol{e}(t) - \boldsymbol{e}'(t)|| \geq d_{\varepsilon} \forall t
\end{align}

\subsubsection{Rotation Transformation}
\begin{figure}[htbp]
	\centering
	\resizebox{0.3\textwidth}{!}{\includegraphics{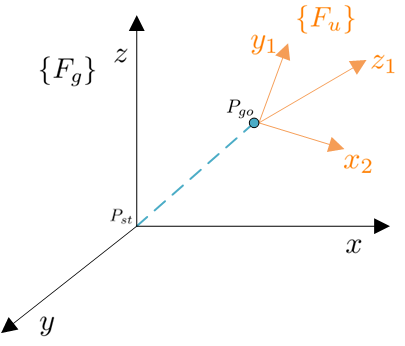}}
	\caption{Rotation transformation }
	\label{my3fig2}
\end{figure}

Fig.~\ref{my3fig2} illustrates the process of rotational transformation from the $F_{g}$ frame to the $F_{u}$ frame.

In the figure, $P_{st}$ represents the starting point, and $P_{go}$ represents the end point. The coordinate conversion matrix $D$ is given by:

\begin{align}
	D=
	\begin{bmatrix}
		\sin \alpha \cos \beta & \cos \alpha \cos \beta & \sin \beta \\
		-\sin \alpha & \cos \alpha & 0 \\
		-\sin \alpha \sin \beta & -\cos \alpha \sin \beta & \cos \beta
	\end{bmatrix}
\end{align}
We use $\alpha$ to represent the angle between the $x_0$ axis and the $x_1$ axis, while $\beta$ represents the angle between the $x_1$ axis and the $x_2$ axis. The conversion method between the resulting coordinate system and the initial coordinate system is given by:

\begin{align}
	\begin{bmatrix}
		\begin{array}{c}
			P_{i} \\
			\hdashline  1
		\end{array}
	\end{bmatrix}
	= \begin{bmatrix}
		\begin{array}{c:c}
			{ }_g^u D & { }^g P_{st} \\
			\hdashline 0 & 1
		\end{array}
	\end{bmatrix}
	\times
	\begin{bmatrix}
		\begin{array}{c}
			{ }^u P_{\text {go}} \\
			\hdashline  1
		\end{array}
	\end{bmatrix} \label{my3Rotation}
\end{align}

where$P_{i} = \left[x_{st}, y_{st}, z_{st}\right]^T$ represents the point in the original coordinate system, and ${ }^u P_{\text {go}} = \left[x, y, z\right]^T$ represents the point after the conversion.

\textbf{Assumption 1:} The velocity of obstacles can be denoted as $ v_{x} $, it must satisfy the following constraints: $0 \leqslant ||v_{x}|| < V_{max}$. Failure to meet these constraints renders avoidance impossible.

\textbf{Assumption 2:} Despite the obstacle being unknown to the UAV, the on-board sensors are capable of detecting the surface velocity $v_{x}$, of unknown obstacles and can provide the relative velocity, denoted as $\bigtriangleup v$.

\subsection{Motion Planning Strategy in 3D}\label{my3s3}

This section proposed navigation algorithm for UAV in 3D environment is inspired idea from \cite{wang2018strategy,elmokadem2021hybrid}. As an extension of the research in Chapters 3 and 4, we have upgraded the original 2D algorithm to a 3D algorithm. When the UAV performs obstacle avoidance in the air, a "collision avoidance plane" is formed. This plane is defined by the UAV's current position, the coordinates of the nearest point on the obstacle, and the coordinates of the target point. At any given moment, the UAV executes reactive path planning on this plane. The details on how to choose the collision avoidance plane is provided in the following sections.

There are two important points to note: 1.The plane changes with each obstacle avoidance event. 2.To avoid the situation where the three points that define the plane are collinear, we additionally select two nearby points on the obstacle to determine the collision avoidance plane more effectively.

We assume here is one unknown obstacle $ \chi_{i} $, we will define a collision avoidance plane $ H_a (t) $ going through the UAV current position and obstacle. This surface can be constructed by robot's direction $\widetilde a (t_{*})$ and $ \mathop{T} \limits ^{\rightarrow}$. Where the $\mathop{T} \limits ^{\rightarrow} $ is  a vector from the robot's coordinates tangent to nearby obstacle at the given time $ t_{*} $.
The avoidance maneuver starts at the time $ t_{} $, and the direction of the normal vector of the plane $ \mathop{n} \limits ^{\rightarrow} H_a (t{})$ is followed by the expression.
\begin{align}
	\mathop{n} \limits ^{\rightarrow}  H_a (t_{*})
	= \mathop{T} \limits ^{\rightarrow} \times  \widetilde a (t_{*})
\end{align}

\begin{figure}[htbp]
	\centering
	\resizebox{0.4\textwidth}{!}{\includegraphics{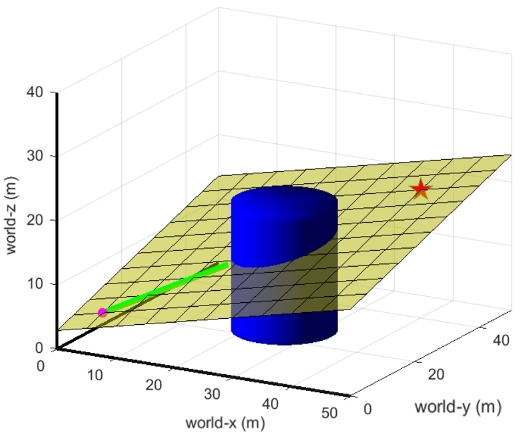}}
	\caption{Choosing a collision avoidance plane when the UAV needs to avoid obstacles}
	\label{my3fig1}
\end{figure}

Within this plane, it is possible to identify the surface points of the obstacle. By applying a rotational transformation (\ref{my3Rotation}), these points can be converted to the coordinate frame of the UAV $ F_{u} $.

Now, we can introduce the navigation method for obstacle avoidance:
\begin{align}
	\begin{array}{l}
		V(t) = ||v_{\Gamma}(t)||,\\
		u(t) = -U_{max}f(v_{\Gamma}(t),v_{u}(t)) \mathop{i_{n}} \limits ^{\rightarrow} (t) \\
		\mathop{i_{n}} \limits ^{\rightarrow} (t)  = 
		\widetilde a (t) \times 
		\mathop{n} \limits ^{\rightarrow}  H_a (t)
	\end{array} \label{my3M1}
\end{align}

where $ \mathop{i_{n}} $ is an unit vector perpendicular to $ \widetilde a (t) $ directing away from the surface of obstacle. The vector 	 $ v_{\Gamma}(t) $ denotes the desired velocity of UAV. It can be denote as:
\begin{align}
	v_{\Gamma}(t) = v_x(t) + l^{i}(t), \forall i = 1,2
\end{align} 
This equation $V(t) = ||v_{\Gamma}(t)||  $ from (\ref{my3M1}) can replace the velocity of nonholonomic robot model $ \dot{e} = V(t) $. Besides, the vector $ v_{\Gamma}(t) , \forall i =1,2$ shows the two directions to avoid the obstacle, which combines the velocity vector of the obstacle $ v_x(t) $ and the two occlusion lines $ l^{i}(t) $. In here, the occlusion lines $ l^{i}(t) $ are related with robot's velocity $ \dot{e}(t) $, the observation angle $ \alpha^i(t) $ and enlarged avoiding angle $ \beta^i(t), i = 1,2 $ . Some specific formulas are as follows:
\begin{align}
	\left\{\begin{array}{l}
		\beta^i(t) = \alpha^i(t) \pm \alpha_{safe} , \\[1ex]
		\angle L1(t) =  \cos(\beta^i(t)), \\[1ex]
		\angle L2(t) =  \sin(\beta^i(t)), \\[1ex]
		\bigtriangleup V(t) = V_{max} - ||v_{x}(t)||, \\[1ex]
		l^i(t) = \bigtriangleup V(t)[\angle L1(t), \angle L2(t)].
	\end{array}\right.  \label{my3controlmethod}
\end{align}

The intuitive approach to obstacle avoidance strategies is greatly influenced by \cite{savkin2013simple,wang2018strategy} . These sources propose a locally-based, biologically-motivated path planning algorithm.  In this research, we improve this algorithm to represent detected obstacle as geometric structures, such as cylinders. Besides, we define a avoid plane $ H_a (t) $ and rotate the coronate in UAV frame $ F_u $ for safe navigation. These representations serve as constraints that need to be avoided during the path planning process.

This strategy can be divided into two modes in the framework of UAV operations. The first mode, avoidance mode($ M1 $), uses the proposed algorithm (\ref{my3M1}) to perform local obstacle avoidance based on distance and angle considerations. In this phase, the UAV takes the precautions necessary to avoid obstacles based on the information it has gathered about them. The second mode is the normal mode of the UAV ($ M2 $), in which it perpetually faces the target direction at maximum speed. In $ M2 $, the unmanned aerial vehicle does not execute local obstacle avoidance manoeuvres and instead flies directly towards the target. The switch rule borrowed by \cite{wang2018strategy} between $ M1 $ and $ M2 $ as follow:

\textbf{\textit{R1}}: Switching from mode $ M2 $ to mode $ M1 $ occurs at any time $t_0$ when the distance from the robot to the obstacle $i$ reduces to the value $C$, i.e., $d_i\left(t_0\right)=C$ and $\dot{d}_i\left(t_0\right)<0$

\textbf{\textit{R2}}: Switching from mode $ M1 $ to mode $ M2 $ occurs at any time $t_*$ when $d_i\left(t_*\right) \leq  d_{\varepsilon} + \kappa (\kappa >0)$ and the vehicle is oriented towards the final destination point, i.e., the vehicle velocity vector $v(t)$ is co-linear to the heading to the final destination point $H\left(t^*\right)$.

\textit{Remark 3.1.} the enlarged avoiding angle $ \beta^i(t) $ is obtained by adding the observation angle $ \alpha^i(t) $ and an avoiding constant angle. The term $ \alpha_{safe} $ ($ 0< \alpha_{safe} < \frac{\pi}{2}$) is used here to refer to a constant avoiding angle. The vector $ l^i(t) $ can  represent the boundaries of the vision cone from the vehicle to the obstacle, ensuring that the avoidance directions remain within this cone.

\textit{Remark 3.2.} To determine the value of $ i $ in equation (\ref{my3controlmethod}) that will satisfy the requirements of the obstacle avoidance navigation algorithm described in Remark 3.1, we introduce one function $ \zeta (\angle a_1,\angle a_2) $, which calculates the angle from vector $ a_1 $ to vector $ a_2 $ in the counter-clockwise direction. The index of the smallest angle $ \zeta $ is then used as an input to the function $ f(\cdotp,\cdotp) $ in control method (\ref{my3M1}).
\begin{align}
	Index = \min_{i = 1,2} \left| \zeta(\angle a_1,\angle a_2)
	\right|
\end{align}
Let $ N \in [1,2] $ represent the chosen index number. We can now introduce the function $ f(\cdot,\cdot) $.
\begin{align}
	f(l_1,l_2) = 
	\left\{\begin{array}{ll}
		0, & if \quad \varphi(l_1,l_2) =0 \\
		1, & if \quad 0< \varphi(l_1,l_2) \leq \pi \\
		-1, & if \quad -\pi < \varphi(l_1,l_2) < 0
	\end{array}\right.
\end{align}

\textit{Remark 3.3.} In a particular scenario, let us consider the case where the robot encounters a significantly large obstacle, such as a wall, where the presence of an edge is not detectable. In such instances, the avoidance plane normal $ \mathop{n} \limits ^{\rightarrow} H_a (t{})$ can be arbitrarily chosen to follow the boundary of the obstacle until a collision-free trajectory can be achieved.

\begin{figure}[htbp]
	\centering
	\resizebox{0.35\textwidth}{!}{\includegraphics{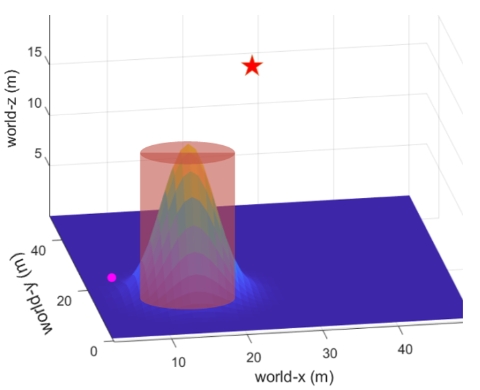}}
	\caption{One probably  method for determining the avoiding plane is to treat the unknown obstacle as a cylinder}
	\label{my3fff3}
\end{figure}

\textit{Remark 3.4.} There are various methods available for determining $ \mathop{T} \limits ^{\rightarrow} $ and $ \mathop{n} \limits ^{\rightarrow} H_a (t)$. Taking inspiration from \cite{elmokadem2021hybrid}, the author adopted a strategy wherein a bounding shape for the obstacles is established using distance measurements obtained from perception systems, such as an ellipsoid. In this research, the obstacles occurring in uneven terrain can be treated as cylinders, see Fig.~\ref{my3fff3} and the angle can be determined by identifying the tangent line.

\subsection{Simulation Results}\label{my3s4}
In this section, we delineate a comprehensive set of case studies executed in a simulated forest environment with varying topographical complexities. These studies aim to rigorously assess the performance and robustness of the proposed navigation algorithm, particularly in the context of search and rescue missions in unfamiliar and challenging terrains. Utilizing MATLAB for simulation and performance evaluation, we present the results in the following analysis, emphasizing salient findings and insights.

The proposed navigation algorithm was implemented in MATLAB, where the 3D environment was modelled using a grid-based elevation map to represent the terrain's topography. Obstacles, both static and dynamic, were modelled as cylinders with varying radii and heights to mimic real-world scenarios. The UAV's initial and target positions were defined in 3D space, and its motion was governed by a discrete-time kinematic model, updated at each simulation step. The entire uneven terrain was also considered an obstacle for the UAV, making the environment more challenging. Simulation results demonstrated that our proposed reactive path planning method is fully applicable in complex environments, successfully achieving collision-free navigation in the presence of both uneven terrain and dynamic aerial obstacles. To validate the effectiveness of the algorithm, we utilized Case 1 and Case 2, which represent a simple environment and a complex environment, respectively, as test scenarios.

A critical aspect of the implementation was the real-time obstacle avoidance strategy, which relied on dynamically recalculating a "collision avoidance plane" at each time step. This plane was formed based on the UAV's current position, the closest point on a detected obstacle, and the target position. When the UAV detected an obstacle within a predefined threshold, the algorithm triggered an avoidance manoeuvrer by switching to a reactive path planning mode on this plane. The UAV's path was continuously adjusted to maintain a safe distance from obstacles while progressing towards the target. In cases where the path was obstructed, the algorithm initiated a replanning process, using global path planning techniques to generate a new reference path for the UAV to follow.

\begin{figure}[htbp]
	\centering
	\begin{subfigure}{0.48\columnwidth}
		\includegraphics[width=\textwidth]{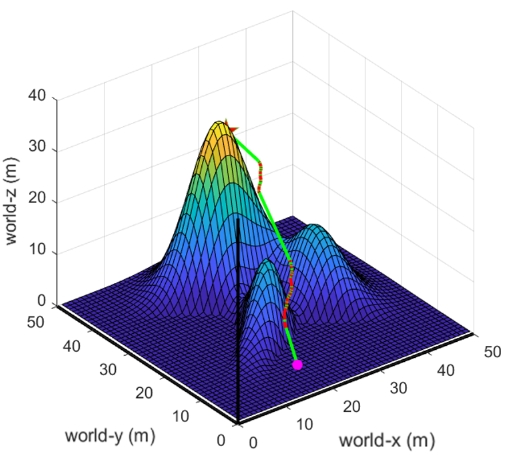}
		\caption{Environment without moving obstacles}
	\end{subfigure}%
	\hfill
	\begin{subfigure}{0.48\columnwidth}
		\includegraphics[width=\textwidth]{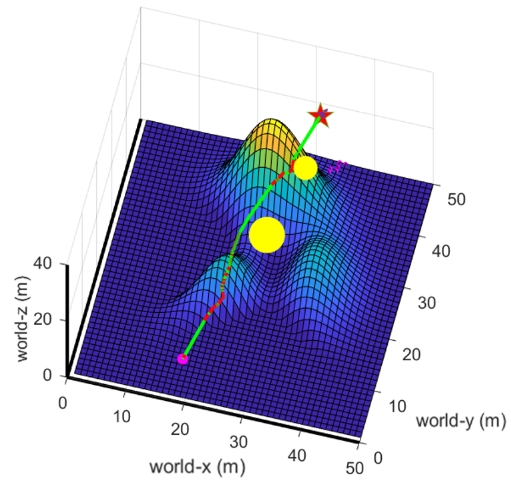}
		\caption{Environment with moving obstacles}
	\end{subfigure}
	\caption{Case 1: Simulation result with multiple unknown peaks ground.Yellow spheres are unknown moving obstacles}
	\label{my3case1}
\end{figure}

In the first simulation, a three-dimensional environment with multiple unknown peaks was considered. Each encountered obstacle along the path was modeled as a cylinder (described in Section III). The navigation algorithm was employed to generate feasible and safe paths based on the selection of avoidance planes. The design parameters employed in the simulation were derived from Case 1 as shown in Table~\ref{my3table1}. With an initial position $ e_0 = [18, 7, 2]^T$ and a target position $ e_{target} = [32,45,25]^T $, Fig.~\ref{my3case1} on the left demonstrates the UAV can achieve obstacle avoidance in an unknown environment. In addition, we add some moving obstacles in this case with the speeds of the moving obstacles:  $v_{x_{1}} = [0,1,1]$ and $v_{x_{2}} = [-1,1,0]$, and the result can be seen on the right of Fig.~\ref{my3case1}. It is now clear that the simulation demonstrated the successful achievement of the target without any collisions.

\begin{figure}[htbp]
	\centering
	\resizebox{0.48\textwidth}{!}{\includegraphics{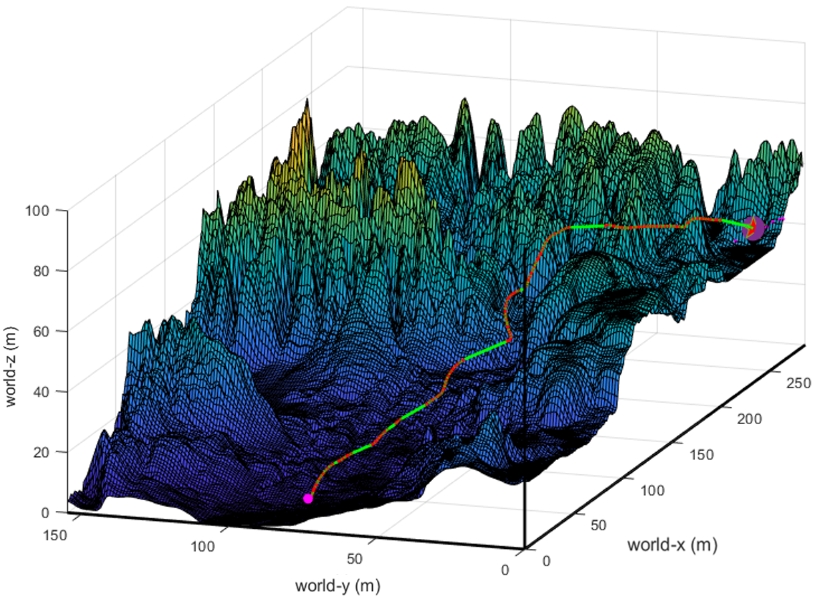}}
	\caption{Case 2: Simulation results with no unknown obstacles uneven terrain region spanned from latitude (44.4-43.5) and longitude (-111.7,-110)}
	\label{my3case2_1}
\end{figure}

In Case 2, while keeping the other parameters unchanged, we extracted elevation data for a specific forest map from USGS. The selected region spanned from latitude 44.4 to 43.5 and longitude -111.7 to -110, with a sampling resolution of 6 arc-seconds. We choose the UAV starting point as $e_{0} = [23,81,5]^T$. The ultimate objective was to reach a target point $e_{target} = [258,4,45]^T$. Since this map represented a larger environment compared to Case 1, we maintained the remaining parameters unchanged but only change some parameter. Specifically, we set $d_{\varepsilon} = 10 m$, $C = 3d_{\varepsilon}$, and $R_{sensor} = 30 m$. The simulation result of this uneven terrain is depicted in Fig.~\ref{my3case2_1}, in where the red line represents the UAV's trajectory while operating in mode $M1$, and the green line represents UAV operating in mode $M2$.

\begin{figure}[htbp]
	\centering
	\begin{subfigure}{0.48\columnwidth}
		\includegraphics[width=\textwidth]{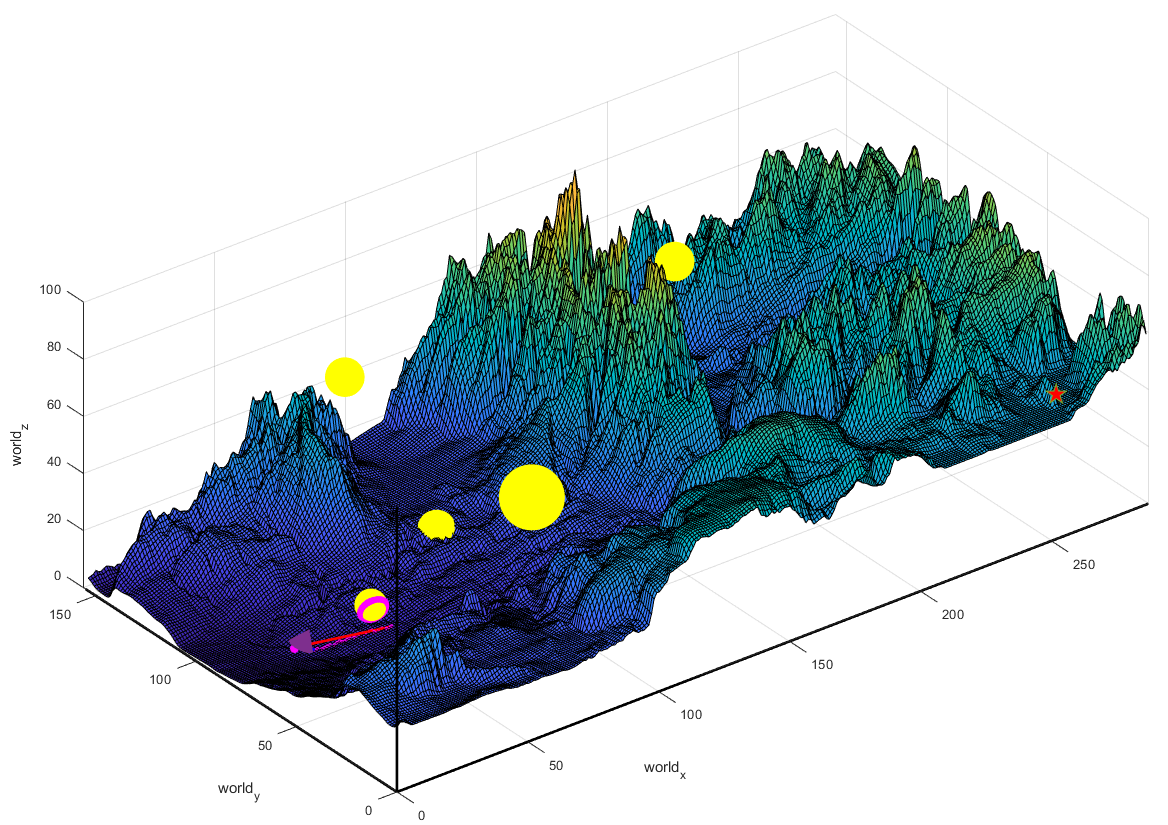}
		\caption{$ T = 0.2\, \mathrm{sec}  $}
	\end{subfigure}%
	\hfill
	\begin{subfigure}{0.48\columnwidth}
		\includegraphics[width=\textwidth]{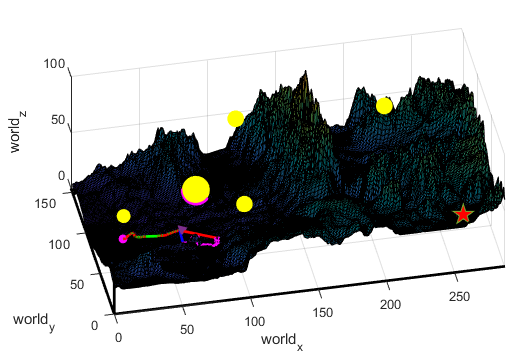}
		\caption{$ T = 14.25\, \mathrm{sec} $ }
	\end{subfigure}\\[0.5ex]
	\begin{subfigure}{0.48\columnwidth}
		\includegraphics[width=\textwidth]{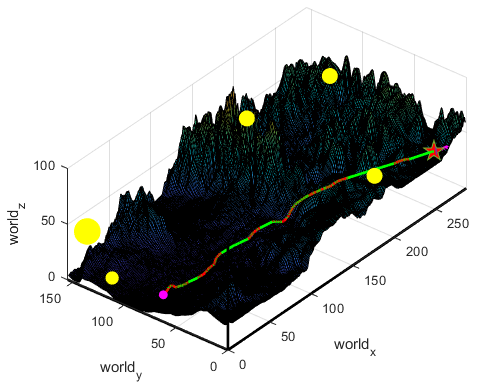}
		\caption{$ T = 98\, \mathrm{sec} $ }
	\end{subfigure}%
	\hfill
	\begin{subfigure}{0.48\columnwidth}
		\includegraphics[width=\textwidth]{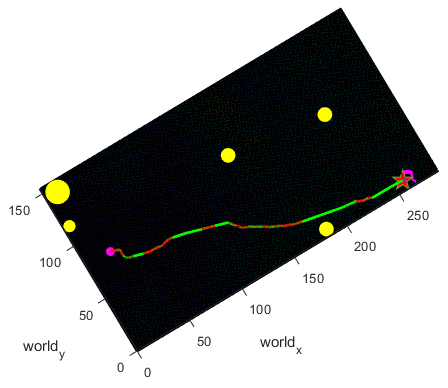}
		\caption{$ T = 98\, \mathrm{sec} $ }
	\end{subfigure}
	\caption{Simulation results in adding multiply moving obstacles (yellow spheres) in complex uneven terrain. $ T $ is the total time}
	\label{my3case2_2}
\end{figure}

In this case (see Fig.~\ref{my3case2_2}), we also trying to add some flying obstacles in the environment. We randomly chose five moving obstacles with velocities $v_{x_{1}} = [-1, 1, 0]$, $v_{x_{2}} =[1, -1, 1]$, $v_{x_{3}} =[1, -1, 1]$, $v_{x_{4}} =[1, -1, 0.5]$ and $v_{x_{5}} =[-1, 1, 0]$ respectively. The figure below (Fig.~\ref{my3case2_2}) illustrates the process of dynamic obstacle avoidance by a UAV in a reshaped terrain map. Similar to Case 1, the UAV's planned route towards the target destination was represented by a green line, while the red line depicted the UAV's obstacle avoidance mode when encountering unknown obstacles. This visualization allowed for a clear understanding of the UAV's navigation behaviour in response to obstacles, and the pink line shows the detected range.

\begin{figure}[htbp]
	\centering
	\begin{subfigure}{0.48\columnwidth}
		\includegraphics[width=\textwidth]{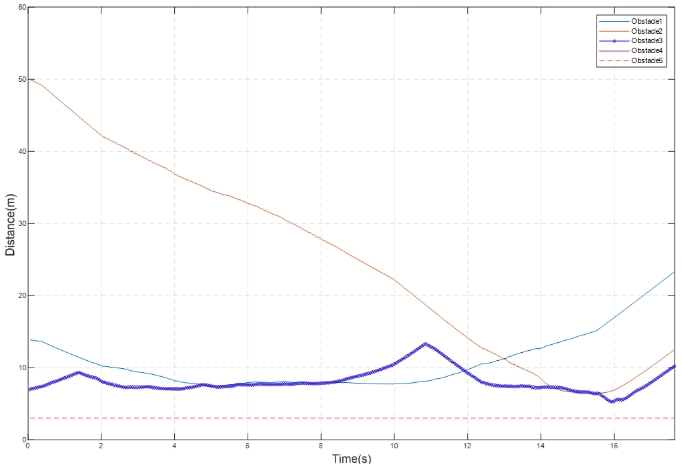}
		\caption{case1}
	\end{subfigure}%
	\hfill
	\begin{subfigure}{0.48\columnwidth}
		\includegraphics[width=\textwidth]{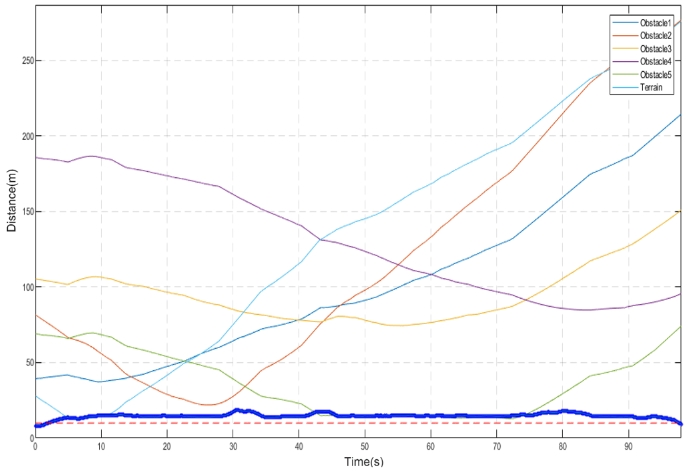}
		\caption{case 2}
	\end{subfigure}
	\caption{Distance to the ground and each obstacles in two cases. The red dash line is the safety line defined}
	\label{my3datanysis}
\end{figure}

Additionally, we conducted an analysis of the real-time distance between the UAV and all moving obstacles in the map (see in Fig.~\ref{my3datanysis}), as well as the distance to the ground in both Case 1 and Case 2. In the figure, the red dashed line indicates the safe distance $d_{\varepsilon}$ set at the beginning for the UAV, and the thick blue line represents the distance of the UAV to the ground. As can be seen from the figure, the reactive algorithm is able to perform safe dynamic obstacle avoidance navigation in complex maps.

To further validate the feasibility of the algorithm, we conducted analysis on Case 3. In Case 3, we selected two different geographic regions in the United States for verification. The first region chosen for simulation and analysis was a rugged mountainous area with longitude ranging from -110.83 to -110.03 and latitude ranging from 44.27 to 44.80. The second region selected was characterized by multiple hills, with longitude ranging from -110.74 to -110.21 and latitude ranging from 44.27 to 44.58. The parameters for these simulations were set according to Case 3, as outlined in Table~\ref{my3table1}.

\begin{figure}[htbp]
	\centering
	\begin{subfigure}{0.48\columnwidth}
		\includegraphics[width=\textwidth]{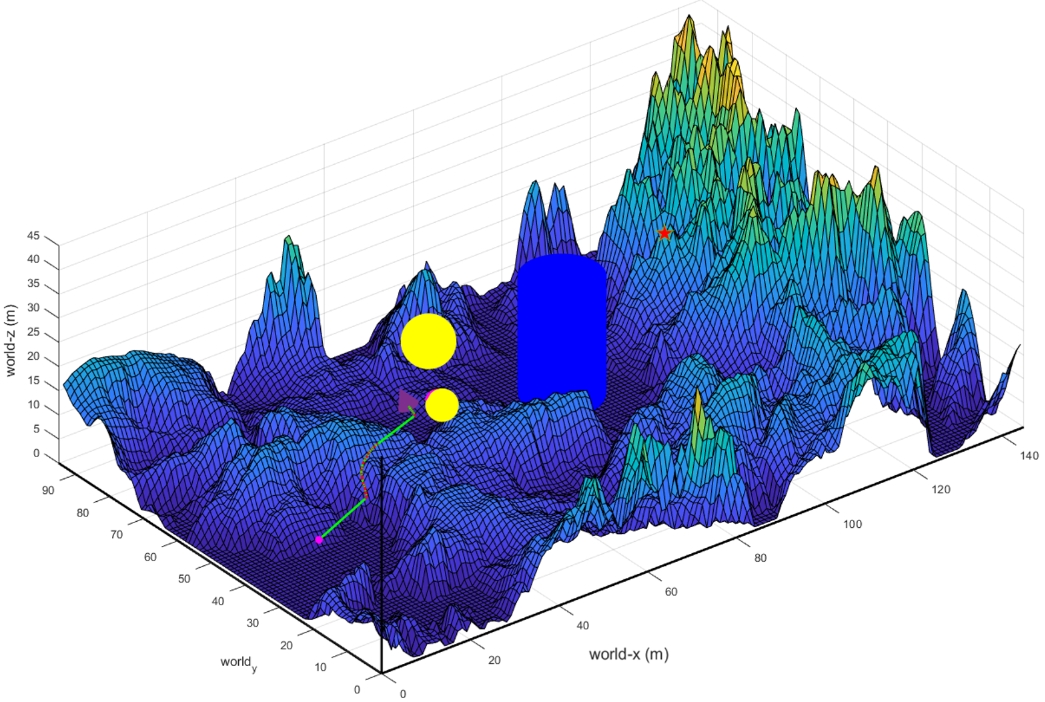}
		\caption{}
	\end{subfigure}%
	\hfill
	\begin{subfigure}{0.48\columnwidth}
		\includegraphics[width=\textwidth]{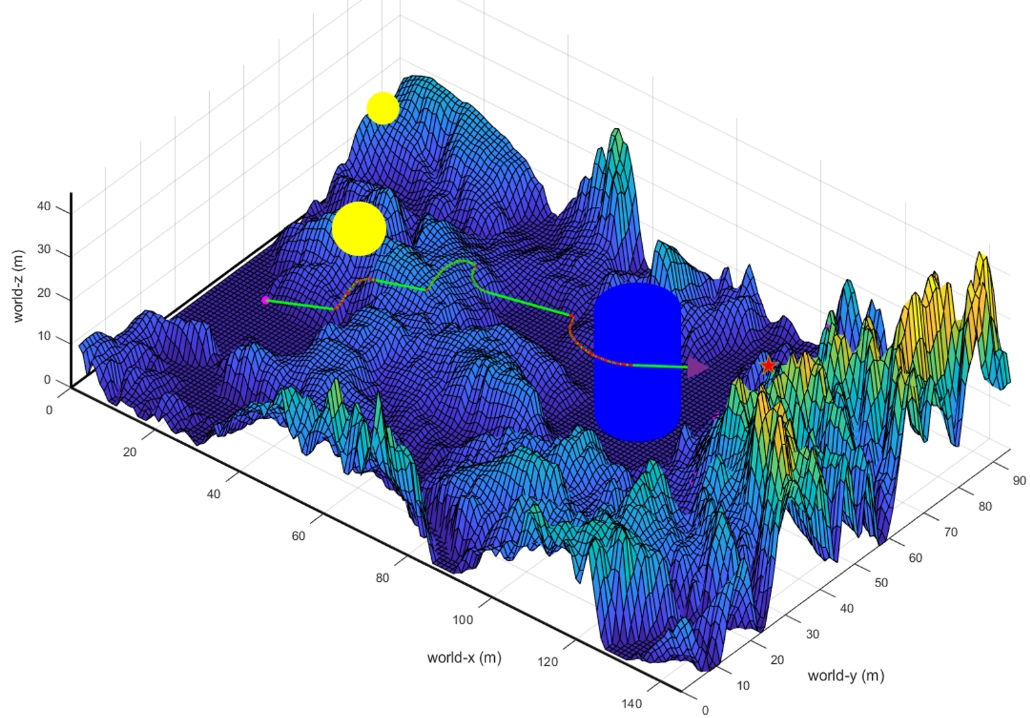}
		\caption{}
	\end{subfigure}\\[0.5ex]
	\begin{subfigure}{0.48\columnwidth}
		\includegraphics[width=\textwidth]{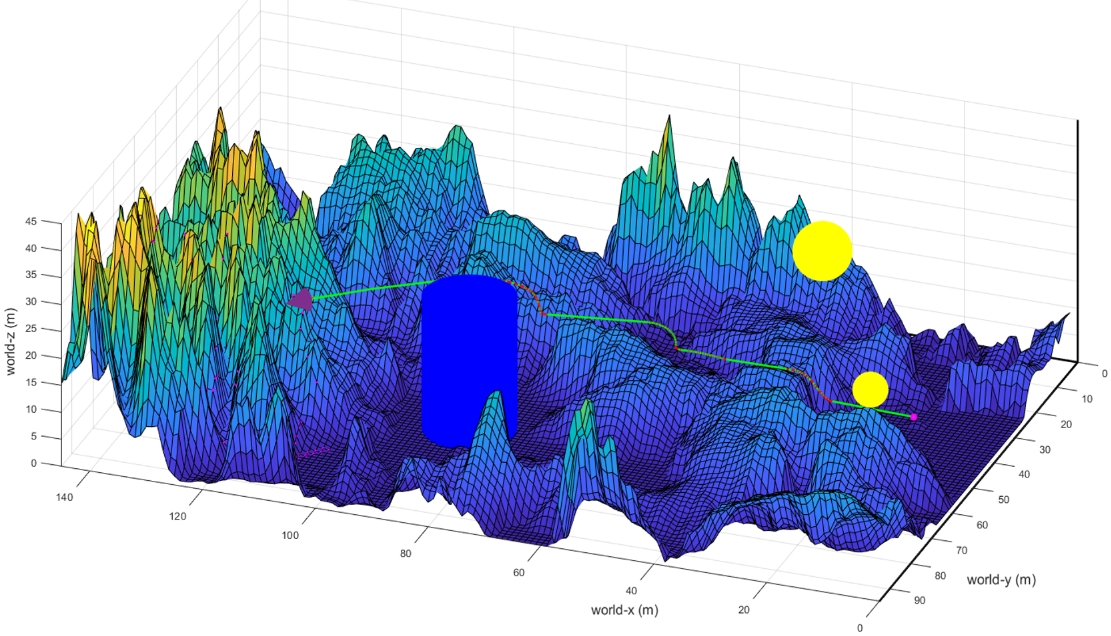}
		\caption{}
	\end{subfigure}%
	\hfill
	\begin{subfigure}{0.48\columnwidth}
		\includegraphics[width=\textwidth]{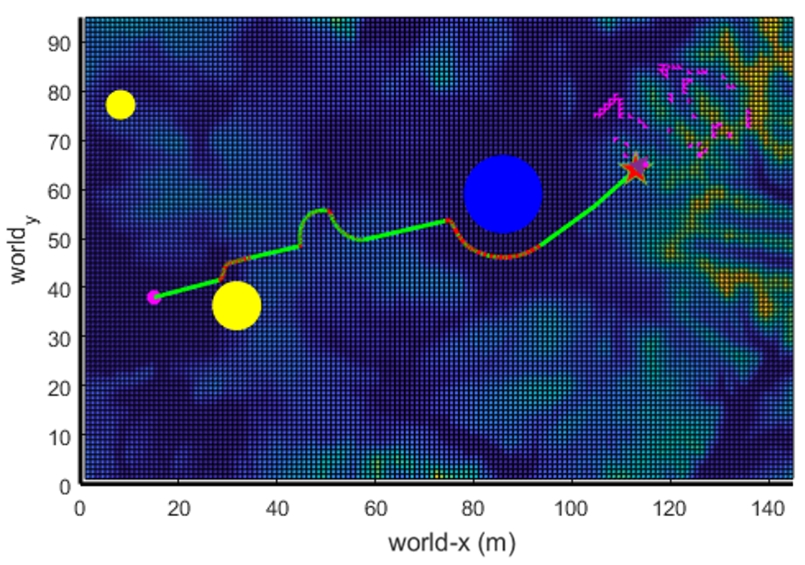}
		\caption{}
	\end{subfigure}
	\caption{Case 3(1): Simulation results with region spanned from latitude (44.27-44.80) and longitude (-110.83,-110.03). (a) is the operational status of the UAV when the total time reaches $  12.4\, \mathrm{sec}  $, (b) represents the operational status of the UAV when the total time reaches $  32.65\, \mathrm{sec} $, (c) and (d) are the UAV status at total runtime of $  39\, \mathrm{sec} $ from 3D view and top view}
	\label{my3case3_1}
\end{figure}

\begin{figure}[htbp]
	\centering
	\begin{subfigure}{0.48\columnwidth}
		\includegraphics[width=\textwidth]{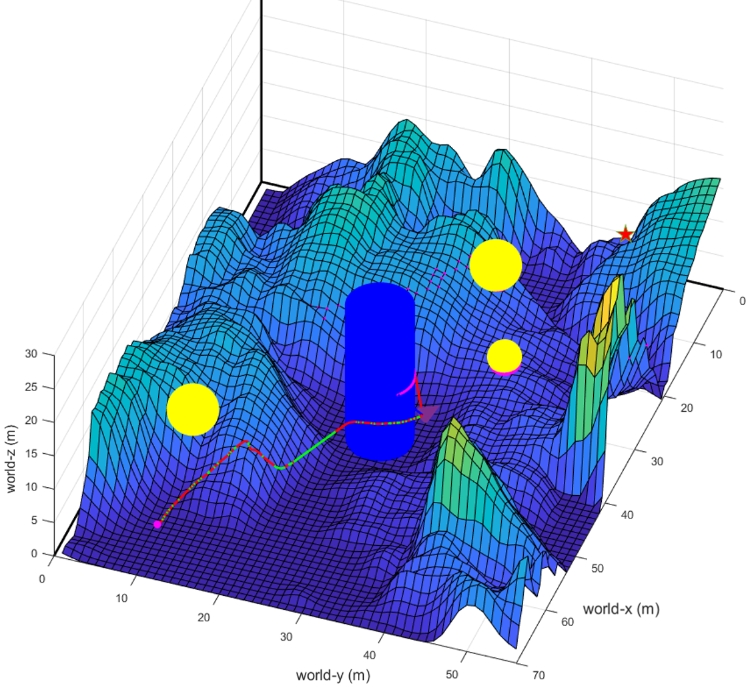}
		\caption{}
	\end{subfigure}%
	\hfill
	\begin{subfigure}{0.48\columnwidth}
		\includegraphics[width=\textwidth]{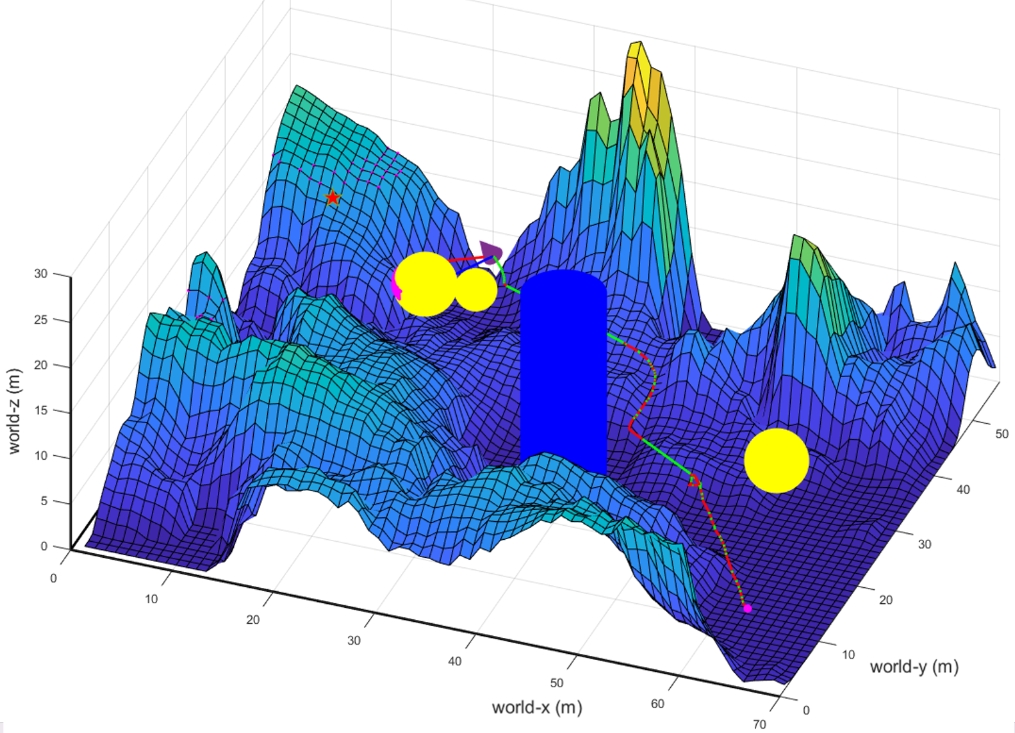}
		\caption{}
	\end{subfigure}\\[0.5ex]
	\begin{subfigure}{0.49\columnwidth}
		\includegraphics[width=\textwidth]{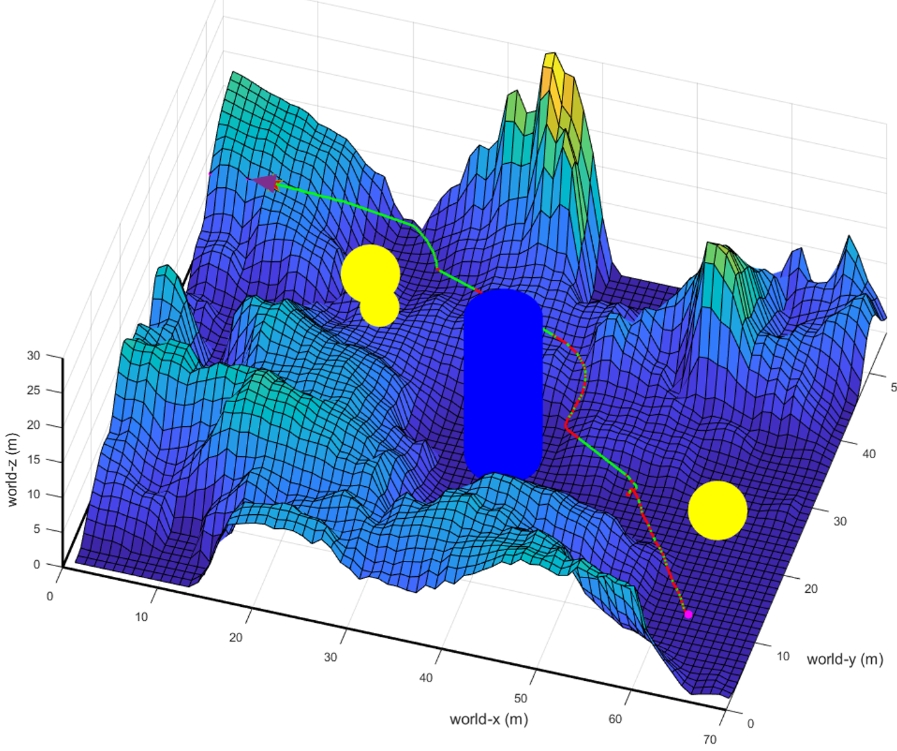}
		\caption{}
	\end{subfigure}%
	\hfill
	\begin{subfigure}{0.48\columnwidth}
		\includegraphics[width=\textwidth]{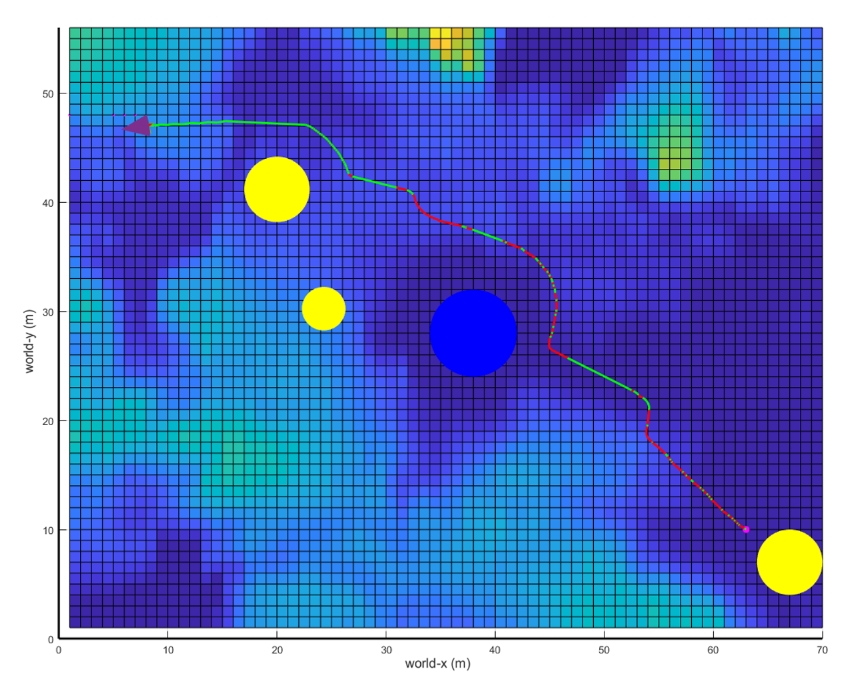}
		\caption{}
	\end{subfigure}
	\caption{Case 3(2): Simulation results with region spanned from latitude (44.27-44.58) and longitude (-110.74,-110.21). (a) denotes the UAV operational status in $ 14.8\, \mathrm{sec}  $, (b) is the status in $ 26.2\, \mathrm{sec} $, (c) and (d) are the represents the final status in  $30.9\, \mathrm{sec} $   in both 3D view and top view}
	\label{my3case3_2}
\end{figure}

The results of the third case are depicted in Fig.~\ref{my3case3_1} and Fig.~\ref{my3case3_2}. In Fig.~\ref{my3case3_1} illustrates the simulation scenario of Case 3 (1), where yellow spheres represent moving obstacles with velocities of v$ _{x_{1}} = [-1.5,1,0.5] $ and $ v_{x_{2}} = [-1,-1,0.5] $. The blue cylindrical structure represents a large static obstacle. The UAV successfully completes the mission, even in the presence of a narrow space formed by the two obstacles. In Fig.~\ref{my3case3_2} (case 3 (2)), the three yellow spheres have velocities of $ v_{x_{1}} =[-0.65,-0.8,0.2] $, $ v_{x_{2}} =[0,0.4,-0.4] $, and $ v_{x_{3}} =[1,-0.8,0.5] $, respectively, while the blue cylindrical obstacle remains stationary. The UAV efficiently reaches the target point, demonstrating the effectiveness of the proposed navigation algorithm in accurately avoiding obstacles while achieving the mission objectives.	

\begin{table}[h]
	\caption{Setting of Parameters for Simulation}
	\label{my3table1}
	\begin{center}
		\begin{tabular}{c||c|c|c|c}
			\hline
			\multicolumn{1}{c||}{case/parameters} & Case1 & Case2 & Case3(1) & Case3(2) \\
			\hline
			$ V_{\text{max}}(\, \mathrm{m/s}) $ & 4 & 4 & 4 & 4\\
			$ U_{\text{max}} (\, \mathrm{rad/s}) $ & 4 & 4 & 4 & 4\\
			$ \kappa (\, \mathrm{m})$ &  4  & 4 & 4 & 4\\
			$ \alpha_{\text{safe}} $ & $ \frac{\pi}{5} $ & $ \frac{\pi}{5} $ & $ \frac{\pi}{5} $ & $ \frac{\pi}{5} $\\[0.5ex]
			$ d_{\varepsilon} (\, \mathrm{m})$ & 3 & 8 & 5 & 4\\
			$ C (\, \mathrm{m})$ & $ 1.1d_{\varepsilon} $ & $ 3d_{\varepsilon} $ & $ 3d_{\varepsilon} $ & $ 1.1d_{\varepsilon} $\\
			$ R_{sensor} (\, \mathrm{m})$ & 10 & 20 & 20 & 10 \\
			$T_{sample}( \, \mathrm{sec} $) & 0.025 & 0.025 & 0.025 & 0.025 \\
			\hline
		\end{tabular}
	\end{center}
\end{table}

\subsection{Conclusion}\label{my3s5}

In conclusion, our research demonstrates a novel 3D reactive navigation method  to effectively guide the UAV to complete missions in complex environments. This lays a solid foundation for safe, efficient, and reliable operations in the field of UAV applications. Future research can further explore optimization of path planning algorithms and more accurate sensor technologies to enhance the performance and adaptability of UAV navigation.

\section{A Method for UAV  Collision-Free Path Planning in Forest Fire Rescue Missions over an Uneven Terrain}\label{mypaper4}
We elaborate on a reactive 3D uneven terrain navigation algorithm in section \ref{mypaper3}. In this section, we offer a fusion strategy that combines the suggested reactive method with an RRT based on non-uniform sampling to address the difficulties of performing mountain fire rescue missions in uneven terrain. Furthermore, the integration of a sophisticated 3D simulated fire spread model is proposed, designed to be utilized within this navigational environment. 

\textbf{Some of the work is part of the paper:} \textbf{J Wei}, Z Fang and S Li, "A Method for UAV Collision-Free Path Planning in Forest Fire Rescue Missions over an Uneven Terrain", In 2024 16th International Conference on Computer and Automation Engineering (ICCAE), pp. 599-604. IEEE, Melbourne, 2024.

\subsection{Introduction}
UAV technology advancements have extended their applications in missions like surveillance and delivery in challenging environments \cite{savkin2020bioinspired,savkin2022joint,huang2019algorithm,xu2022multi}. Obstacles and varied terrains in these settings impact UAV visibility and communication, influencing mission success. Current research aims to enhance UAV efficiency in such environments, often through optimization considering constraints like noise and energy consumption.

The challenges of UAV path planning in uneven terrains are substantially addressed in \cite{de2012path}, which validates the use of A*-based and Theta* algorithms for developing dependable and efficient navigation strategies across diverse landscapes. Furthermore, \cite{mittal2007three} presents a novel 3D path planning technique for UAVs, utilizing NSGA-II and NURBS, to provide a balanced solution of efficiency and safety suitable for uneven terrains. The study in \cite{wei2021predicting} applies a neural network model to predict and optimize energy consumption for robots on uneven grounds, offering insights that, while directly applicable to ground robots, may only serve as heuristic guidance for UAV operations. Hailong's research \cite{888} provides a navigation algorithm for UAVs to improve surveillance by keeping a stable line-of-sight and reducing travel distance. However, it lacks in optimizing for obstacle avoidance, which may affect the flight path efficiency.

Optimizing UAV deployment is also critical for tackling the challenges of uneven terrain. This entails identifying the best deployment locations to mitigate the effects of uneven terrain, as discussed in \cite{zhang2022path,savkin2019proactive}. The research in \cite{zhang2022path} introduces a hierarchical path planning approach for wheeled robots, improving navigation in uneven terrains through the A* algorithm and Q-learning. Meanwhile, \cite{savkin2019proactive} tackles a 3D gallery problem to ascertain the minimum number of UAVs needed for complete coverage, ensuring constant monitoring and thereby enhancing surveillance quality, while minimizing the impact of uneven landscapes.

UAVs are increasingly utilized for their advanced surveillance in various missions, notably in bush fire monitoring due to their rapid spread and the difficulty of timely human response. A bushfire model detailed in \cite{zheng2017forest} suggests that flames may serve as both surveillance targets and obstacles affecting flight paths. To navigate these, path planning algorithms that ensure both task completion and obstacle avoidance are essential, as seen in \cite{savkin2020navigation} and \cite{wei2023hybrid}. These sources introduce a reactive algorithm adept at creating smooth paths and avoiding obstacles during bush fires. Nonetheless, they do not address the complications posed by uneven terrain.

In this section, we integrate findings from prior research to present a UAV Collision-Free Path Planning method for Forest Fire Rescue Missions on Uneven Terrain. We employ a hybrid path planning algorithm combining an enhanced RRT algorithm with a reactive model. This hybrid approach enables UAVs to avoid obstacles while effectively tracking the progression of a mountain fire over time, ensuring a smooth flight path. To align our research with real-world scenarios, we detail a mountain fire spread model. We frame the effectiveness of UAV mission execution as an optimization problem, solving it under various constraints to showcase the proposed algorithm's superiority. Simulation results, obtained using computer software, confirm the feasibility of our algorithm.

The structure of this section is as follows: Section \ref{my4s2} introduces the system model and problem statement. Section \ref{my4s3} details the proposed navigation algorithm. Computer simulations that validate the approach's performance are presented in section \ref{my4s4}. Finally, section \ref{my4s5} concludes the section.

\subsection{Proposed approach overview}\label{my4s2}
Consider a challenge navigation problem where a UAV $ Q $ need to execute a rescue mission in the forest fire spread situation. In this section, a hybrid algorithm is introduced, enabling the deployment of UAV $ Q $ in forest fire rescue route planning through a hierarchical control approach. The fundamental terrain of the forest $ D_{terrain} $, the initial positions of the UAVs $ c(t) $, and the rescue objectives $ \varpi_{T} $ are pre-configured in the UAV system as known parameters. The hierarchical strategy architecture under consideration integrates parts derived from both the global planner and reactive navigation concepts are designed to the UAV. This approach is based on an idea known as perception-planning-reaction. In this framework, path planning and the creation of a local environment model by the higher control layer occur simultaneously. In addition, computer simulation of forest fires spread can be achieved through proposed fire spread model in the following section \ref{my4Forestfirespreadmode}.	

In our study, a simulated forest hill environment represented by the 3D space $ \mathbb{D}^3 $ includes various uncharted hills $ D_{terrain} $ and unidentified obstacles $ D_{i} $ $ \{i = 1,2,...,n\} $. These obstacles can be either static or flying  which are regarded as unknown knowledge. Furthermore, the forest fire spread model $ D_{fire} $ is proposed and considered as a deformable obstacle within the environment. The main objective is to steer the UAV through a forest fire scenario for the purpose of executing an essential rescue mission. This involves generating a feasible, obstacle-avoidant trajectory originating from the predetermined initial point $ \varpi_{O} $ and terminating at the designated rescue target $ \varpi_{T} $. It is crucial to emphasize that both of these spatial coordinates are pre-specified locations on the existing map.

\subsubsection{nonholonomic kinematic model}
This study considers the nonholonomic dynamical model of a  UAV, which is denoted as $Q$ while disregarding effects such as rotor drag.  To elucidate our model, two coordinate frames are utilized: the ground coordinate frame $F_g = \{x_g, y_g, z_g\}$, with $z_g$ oriented upwards, and the body coordinate frame $F_b = \{x_b, y_b, z_b\}$, centered at the UAV’s center of mass. The coordinate orientation matrix $ R $ is employed for the conversion between coordinate frames. For a clearer understanding, we will exemplify the transformation relationship. Refer to equation (\ref{matrix}), where the orientation matrix $ { }_g^b R $ represents the transformation from the body coordinate frame $ F_b $ to the ground coordinate frame $ F_g $. The angle between these two coordinate frames is denoted by $\alpha$ and $ \beta $. Subsequently, the transformation matrix $ { }_g^b T $ is derived using the origin point $ { }^g P_{b(org)} $ of the coordinate frame $ F_b $. If a point coordinate $ { }^b P_{go} $ is given in frame $ F_b $, we can calculate its equivalent in frame $ F_g $ using equation (\ref{conversion}).
\begin{align}
	{ }_g^b R =
	\begin{bmatrix}
		\sin \alpha \cos \beta & \cos \alpha \cos \beta & \sin \beta \\
		-\sin \alpha & \cos \alpha & 0 \\
		-\sin \alpha \sin \beta & -\cos \alpha \sin \beta & \cos \beta
	\end{bmatrix} \label{matrix}
\end{align}
\begin{align}
	{ }_g^b T = \begin{bmatrix}
		\begin{array}{c:c}
			{ }_g^b R & { }^g P_{b(org)} \\
			\hdashline 0 & 1
		\end{array}
	\end{bmatrix}\label{matrix2}
\end{align}
\begin{align}
	{ }^g P_{go} = { }_g^b T \times { }^b P_{go} \label{conversion}
\end{align}
Above this,	a nonholonomic mathematical model can be established (see in equation (1-3)).  In this model, the position of UAV $Q$ is expressed as $ p(t) = [x(t), y(t), z(t)]^T $ within the ground coordinate frame. Correspondingly, the UAV's initial and target points in this frame are represented as $ P_O $ and $ P_T $, respectively.  Additionally, the linear velocity of the UAV is represented by $ v_{q}(t) $, with its direction of motion being articulated through two angles: the heading angle $ \phi $ and the flight path angle $ \gamma $. The associated nonholonomic kinematic model is delineated in equation (\ref{my4model}).
\begin{align}
	&\dot{x}(t)  = v_{q}(t) \cos \beta(t) \cos \alpha(t) \\
	&\dot{y}(t)  = v_{q}(t) \sin \beta(t) \cos \alpha(t) \\
	&\dot{z}(t)  = v_{q}(t) \sin \alpha(t) \\
	&\dot{\widetilde a}(t) = u(t)\\
	&(\widetilde a (t),u(t)) = 0 \label{my4model}
\end{align}
where the UAV $ Q $ uses the linear speed $ v_{q}(t) $, angular velocity $ u(t) $ as control input. Due to physical limitations on any mechanical, these control inputs should satisfy:  
\begin{align}
	& 0  \leq v_{q}(t) \leq V^{max}\\
	& \| u(t) \| \leq  U^{max}
\end{align}

This model represents a general kinematic framework, applicable not only to UAVs but also to a variety of vehicles operating in 3D spaces.

Following the coordinate system principles outlined in equations (1), (2), and (3), our collision-free navigation method relies on using three key points to define an avoidance plane. The plane's orientation is determined by its normal vector, which also defines the rotation angle in the matrix ${}_g^b R$. This method integrates three critical components: the current position of the Unmanned Aerial Vehicle (UAV) $Q$, represented as $c(t)$; the next waypoint target; and the nearest obstacle's closest surface point. These components together create the vertices of the avoidance plane, as established in our prior research\cite{wei2023method}.

\subsubsection{Forest fire spread} \label{my4Forestfirespreadmode}
Numerous scholarly inquiries have been conducted to develop models for the spread of forest fires. When attempting to simulate fire spread as a deformable obstacle in a 3D environment, we found it quite challenging. After reviewing several studies, we discovered that some existing fire spread models, such as those discussed in \cite{you2022real,wu2022simulation,himoto2008development,green1983fire,duane2016integrating}, were not suitable for the specific map setup in our research. Inspired by the work in \cite{himoto2008development}, which incorporates elements like temperature rise due to heat transfer, wind speed influence, and critical thermal energy thresholds, we developed a simplified fire spread control model. This model aims to mimic real-world fire spread as closely as possible while being adaptable to the deformable nature of obstacles in the environment. In \cite{duane2016integrating}, the proposed MedSpread model is highlighted as a widely applicable model for simulating various types of fire spread patterns, but it was not suitable for the environment addressed in this section, where a fire model needs to be deformable and capable of detecting dangerous edge points. Therefore, we developed our own model, represented by equations (6.11)-(6.13), which considers both wind and no-wind conditions, with fire spreading at random speeds in each direction of the $ xyz $ axes. In order to describe the fire model applied in this section more clearly, we will summarize and define the symbols used in the equations.

\begin{table}[h]
	\centering
	\caption{Frequently used symbols and their meanings}
	\begin{tabular}{c|l}
		\hline 
		\multicolumn{1}{c|}{Symbol} & \multicolumn{1}{c}{Meaning}  \\ 
		\hline
		$\mathbf{T}(t)$ & Temperature vector at time $ t $. \\
		$T_{\text{trigger}}$ & Temperature threshold to trigger a fire. \\
		$\bigtriangleup T$ &  Increase in temperature at each time step. \\
		$\mathbf{V}_{\text{wind}}(t)$ & Wind speed vector at time $ t $. \\
		$ \kappa $ &  Influence factor for fire spread without wind. \\
		$ \rho $ & Influence factor for fire spread with wind. \\
		$ N(i) $ & Set of 26 neighboring points around location $i $. \\
		\hline
	\end{tabular}
\end{table}

The forest fire spread model uses a temperature vector $\mathbf{T}(t)$ to represent the temperature at various points in space over time. Initially, we clarify the mathematical formulation that prescribes the fire spread model.

\begin{align}
	\frac{d \mathbf{T}_i(t)}{dt} = \sum_{j \in \mathbf{N}(i)} H_{ij}(t)
\end{align}
where,

\begin{align}
	H_{ij}(t) = 
	\begin{cases} 
		\mathbf{T}_j(t) \times W_{ij}(t) \times \rho, & \text{if } \mathbf{T}_j(t) > T_{\text{trigger}} \\ 
		0, & \text{otherwise} 
	\end{cases}
\end{align}

Here, $\mathbf{T}_j(t)$ represents the temperature at the neighboring point $j$, which influences the temperature change at point $i$. This term is crucial because the fire can only spread from point $j$ to point $i$ if the temperature at point $j$ exceeds the trigger temperature $T_{\text{trigger}}$.

and,

\begin{align}
	W_{ij}(t) = \mathbf{V}_{\text{wind}}(t) \cdot \mathbf{D}_{ij} + \kappa
\end{align}

In this model:

\begin{itemize}
	\item $\mathbf{T}(t)$ is a temperature vector, where each $\mathbf{T}_i(t)$ represents the temperature at point $i$ at time $t$.
	\item $W_{ij}(t)$ is the time-dependent weight influenced by the wind vector $\mathbf{V}_{\text{wind}}(t)$ and the direction vector $\mathbf{D}_{ij}$. The term $\mathbf{V}_{\text{wind}}(t) \cdot \mathbf{D}_{ij}$ represents the dot product between these two vectors, indicating how wind influences the spread of fire along the direction from point $i$ to point $j$.
\end{itemize}

$ \textit{Remark} $: In this research, every dynamic obstacle and the progression of fire spread are characterized by distinct velocity attributes. The UAV is outfitted with sensors that can detect the velocity of proximate obstacles within its operational range. It is crucial for the UAV's maximum speed to surpass that of any moving dynamic obstacle, satisfying the condition $0 \leqslant ||v_{wind}|| < V^{max}$. Failure to meet these constraints renders the UAV incapable of avoiding the obstacle.

\subsection{Motion Planning Strategy in 3D}\label{my4s3}
In this section, we introduce a hybrid navigation control algorithm that operates through Deployment and Execution layers to enhance navigation safety. To create deformable obstacles in a 3D environment, we developed a fire spread model as described in Section 6.2.2. Although the fires in this model spread randomly depending on the wind speed and initial settings, the model allows the boundary coordinates of the fires to be determined. When an unmanned aerial vehicle (UAV) approaches, it can detect the peripheral coordinates of these fires and treat them as deformable obstacles for obstacle avoidance. The Deployment layer is responsible for generating global waypoints based on current environmental data, while the Execution layer focuses on real-time collision avoidance and strategy development. The algorithm leverages an improved RRT-connect (Rapidly-exploring Random Tree) method, inspired by prior work \cite{wang2023manipulator}, to iteratively optimize the UAV’s path towards the most effective waypoints. The core function of the Execution layer is to navigate these waypoints, ensuring real-time obstacle avoidance through the UAV’s onboard sensors, which detect and respond to dynamic obstacles.

\begin{figure}[htbp]
	\centering
	\resizebox{0.45\textwidth}{!}{\includegraphics{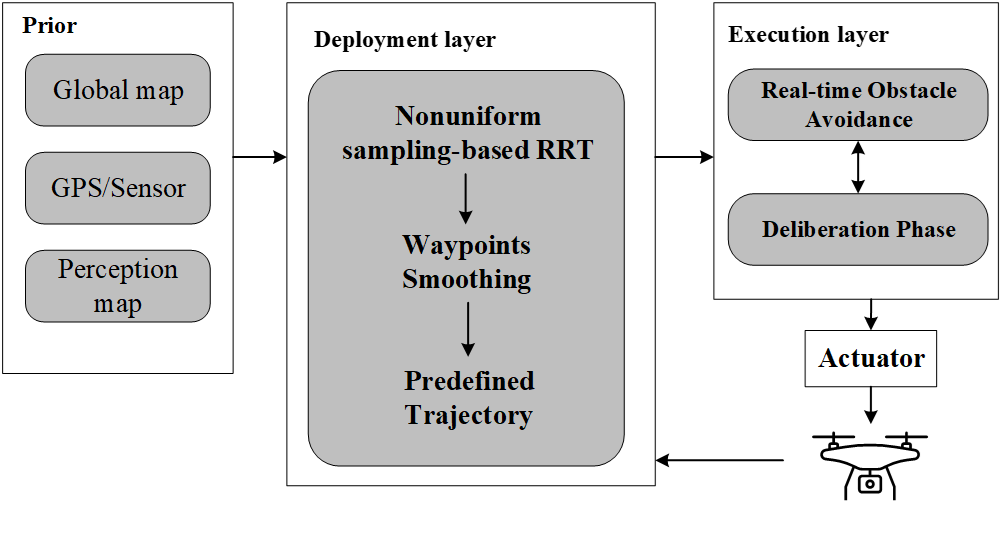}}
	\caption{Motion Planning Strategy}
	\label{methodf1}
\end{figure}

\subsubsection{Deployment layer}\label{A}	
In the Deployment layer, the algorithm begins by utilizing prior knowledge to generate a feasible route. The UAV's initial position, target destination, and surrounding terrain data are defined as key variables in this process. The primary objective is to identify a viable path and then optimize it using the initial solution. To achieve this, we introduced a non-uniform sampling method that enhances the efficiency of the RRT-connect algorithm by directing the sampling process towards areas more likely to yield an optimal path.

In this context, the key variables in the optimization problem include:

- Start Position \(P_0\): The initial coordinates of the UAV.

- Target Position \(P_T\): The final destination coordinates.

- Sampling Points \(P_{sample}\): Intermediate points generated during the RRT process.

- Path Segments \(\tau_i\): Segments of the path that connect consecutive sampling points.\\
The decision variables in the optimization problem are:

- \textbf{\(P_{new}\)}: A newly generated sampling point aimed at improving the current path.

- \textbf{\(J_1\)}: The first cost function, which seeks to minimize the total path length.

- \textbf{\(J_2\)}: The second cost function, which smooths the path using B-spline curves to minimize abrupt changes in direction.

\begin{figure}[htbp]
	\centering
	\resizebox{0.49\textwidth}{!}{\includegraphics{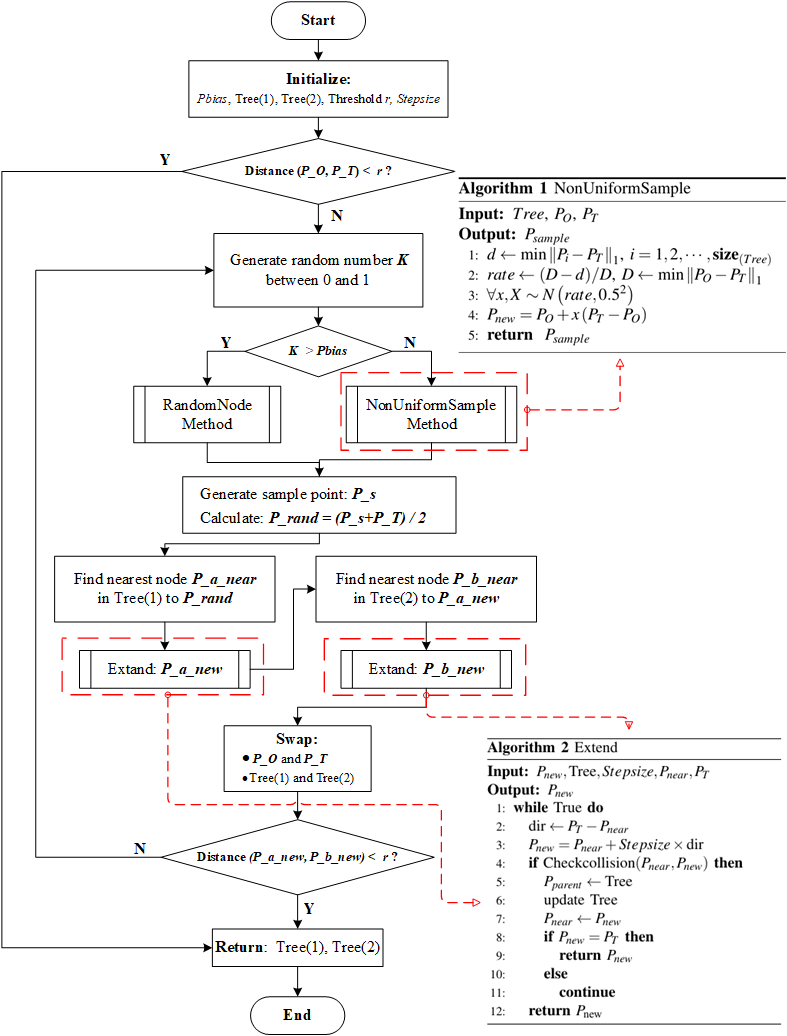}}
	\caption{Flowchart of the  non-uniform sampling based RRT algorithm}
	\label{fig1}
\end{figure}  

The optimization process works by first minimizing the path length (\(J_1\)) by reassessing and potentially replacing nodes in the Deployment layer with new ones if a shorter, collision-free path is identified. Then, the path is further refined through \(J_2\), which minimizes the sum of angles between consecutive path segments to ensure a smooth trajectory. The final objective function, therefore, aims to strike a balance between path length and smoothness. The first cost function, \( J_1 \), seeks to minimize the path length by reassessing nodes from the deployment layer, replacing the original path points with new ones if a shorter collision-free path to the target is found. The optimization for \( J_1 \) is expressed as:
\begin{align}
	&J_1=\underset{\tau \in C_{\text {free }}}{\operatorname{argmin}}\left\{\sum_{i=1}^{\operatorname{Size}(\tau)}\left\|p_i-p_{i-1}\right\|_1  \text{ and } L_{new} < L_{orig}  \right\}	
\end{align}
In this equation:
\begin{itemize}
	\item \( \tau \) represents the path trajectory, consisting of a sequence of nodes \( p_i \).
	\item \( C_{\text{free}} \) denotes the set of all collision-free paths.
	\item The term \( \|p_i - p_{i-1}\|_1 \) calculates the Manhattan distance between consecutive nodes, which is summed over the entire path.
	\item The decision variable in this optimization is the path \( \tau \), and the objective is to minimize the total path length while ensuring that the new path \( L_{\text{new}} \) is shorter than the original path \( L_{\text{orig}} \).
\end{itemize}
After optimizing the path length with \( J_1 \), the second cost function, \( J_2 \), is employed to smooth the trajectory for efficient and continuous motion. \( J_2 \) uses B-spline curve methods to reduce sharp turns in the path by minimizing the sum of angles between consecutive path segments. The optimization for \( J_2 \) is expressed as:
\begin{align}
	J_2= \sum_{i=2}^{n-1}\left|\theta\left(\tau_{i-1}, \tau_i, \tau_{i+1}\right)\right| \label{cost2}	
\end{align}

Here:
\begin{itemize}
	\item \( \theta (\tau_{i-1}, \tau_i, \tau_{i+1}) \) represents the angle between consecutive path segments \( \tau_{i-1} \), \( \tau_i \), and \( \tau_{i+1} \).
	\item The goal is to minimize the sum of these angles to smooth out the trajectory, reducing the sharpness of turns for more fluid motion.
\end{itemize}

\begin{figure}[htbp]
	\centering
	\resizebox{0.49\textwidth}{!}{\includegraphics{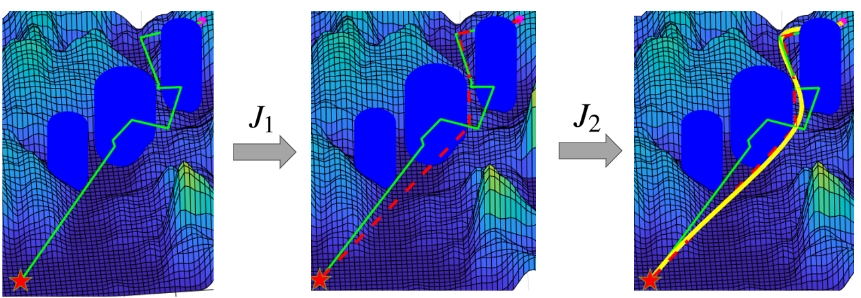}}
	\caption{Sequential Visualization of Pathway Optimization: Initial Non-Uniform Sampling via RRT Algorithm (Green Line), Intermediate Optimal Stage $ J_1 $ (Red Dash Line), and Concluding Optimal Stage $ J_2 $ (Yellow Line)}
	\label{my4optimal}
\end{figure}

By addressing both path length and smoothness, these two cost functions ensure that the final path is both efficient and suitable for real-time execution in the given environment (see in Fig.~\ref{my4optimal}).

\subsubsection{Execution layer}\label{reactive}

The Execution layer strategy encompasses a deliberation phase aimed at deciding whether real-time obstacle avoidance should be carried out. It's essential to introduce the two distinct execution modes. It's worth noting that this aspect of the work has already been completed in prior research efforts.

\subsubsubsection{Real-time Obstacle Avoidance}
We assume that the real-time obstacle avoidance begins at a specific moment, denoted as $ t_* $ (not from the initial moment). For each unknown obstacle, it is necessary to define a collision avoidance plane that intersects the current position of the drone and the obstacle. This plane will be constructed based on the $ x_b $ axis and $ y_b $ axis of the newly defined body coordinate system $ F_b $. This implies that the real-time obstacle avoidance operation will take place within the plane formed at each time interval $ \triangle t $.

The method for selecting an appropriate plane relative to the obstacle is critical for introducing the first execution mode. During the Real-Time Obstacle Avoidance Phase (\(OP_1\)), both the UAV and the unknown obstacles' coordinates are represented in the \(F_b\) frame. The UAV’s velocity \(v(t_*)\) and control input \(u(t_*)\) at time \(t_*\) are computed within the \(F_b\) coordinate system after transformation from the global frame \(F_g\).

Now we can introduce the navigation strategy of the Real-Time Obstacle Avoidance Phase ($ OP_1 $):
\begin{align}
	\begin{array}{l}
		u(t) = -U_{max} \Gamma (v_{\tau}(t),v_{q}(t)) \mathop{i_{n}} \limits ^{\rightarrow} (t) \\
		v_q(t) = ||v_{\tau}(t)||
	\end{array} \label{M1} 
\end{align}

In equation (\ref{M1}), $ \mathop{i_{n}} $ and function $ \Gamma(\cdot,\cdot) $  is to determine the sign of control input $ u(t) $ which calculated from occlusion value $ \theta(l^{(1)},l^{(2)}) $. $ \theta $ denotes the angle between the vector $ l^{(1)} $ and $ l^{(2)}  $ measured from $ l^{(1)} $ in the counter-clockwise direction. In the function \( \Gamma(\cdot,\cdot) \), the terms \( v_{\tau}(t) \) and \( v_{q}(t) \) represent the calculated velocity and the current velocity of the UAV, respectively.

The calculated velocity \( v_{\tau}(t) \) combines the detected surface velocity vector of the unknown obstacle \( v_d(t) \) with two occlusion lines \( l^j(t) \), where \( j = 1,2 \) indicates the two directional strategies for avoiding unknown obstacles. These occlusion lines \( l^j(t) \) are derived from the observed angles \( \beta^j(t) \), which are computed as follows:

\begin{align}
	\left\{\begin{array}{l}
		\beta^j(t) = \alpha^j(t) \pm \alpha_{O} , \\[1ex]
		\bigtriangleup V(t) = V^{max} - ||v_{\tau}(t)||, \\[1ex]
		l^j(t) = \bigtriangleup V(t) \cdot [ \cos(\beta^j(t)), \sin(\beta^j(t)) ] 
	\end{array}\right.  \label{eqop}
\end{align}

In these equations, \( \alpha_{O} \) is a constant that defines a safe avoidance angle, constrained within \( (0,\frac{\pi}{2}) \). The term \( \alpha^j(t) \) represents the angles detected by the UAV from the unknown obstacle, with \( j = 1,2 \) corresponding to the two detection directions.

\subsubsubsection{Deliberation Phase}
In the Deployment layer, this mode is denoted as \( OP_2 \). We initiate with the waypoints tracking mode at time \( t_0 \). In this tracking mode, a \( \text{sign} \) function is employed to ascertain the drone's relative orientation \( \theta_{\text{toward}} \) in the subsequent action, ensuring it is always directed towards the next waypoint. Now, we introduce the control method for mode \( OP_2 \) as follows. It is important to note that all coordinates in mode \( OP_2 \) should be based on the ground coordinate frame \( F_g \).

\begin{align}
	\begin{array}{l}
		u(t) = U^{max}sgn(\theta_{toward}) \\
		v_q(t) = V^{max}
	\end{array}   \label{mode1}
\end{align}

\textit{Remark 1}: In UAV operations, the complexity of tasks demands multi-objective optimization. Prioritizing a singular goal such as the shortest path may compromise other aspects like smoothness or safety. Our proposed optimization problem, denoted as $Q$, integrates three distinct objectives—$J_1$ for the shortest path, $J_2$ for path smoothness, and $J_3$ for minimizing environmental risks, thus ensuring UAV safety in adverse conditions. $Q$ aims to achieve a balanced solution by tuning parameters $\alpha, \beta, \gamma$ to find the best compromise among path length, smoothness, and safety. The optimization function combines these three criteria, subject to the UAV's dynamic constraints and a safety distance constraint to avoid obstacles. The weighting factors $\alpha $, $ \beta $, $ \gamma $, which add up to 1 and vary between 0 to 1, provide the flexibility to shift the optimization focus according to the mission's specific requirements.

\begin{align}
	\begin{gathered}
		\underset{\tau, \alpha, \beta, \gamma}{\operatorname{minimize}} \quad Q=\alpha J_1(\tau)+\beta J_2(\tau)+\gamma J_3(\tau) \\
		\text { subject to } 
		\tau \in C_{\text {free }}, \text{ }
		\alpha+\beta+\gamma=1, \text{ } 
		\alpha, \beta, \gamma \in[0,1]
	\end{gathered} \label{optimalcost}
\end{align}

\textit{Remark 2}: The UAV should follow mode $ OP_2 $ until it encounters an unknown obstacle. The evaluation criteria are determined based on the distance that can be detected by the sensors equipped on the drone. When this detection distance is less than a preset safety value, the drone will switch to Mode One for Real-time Obstacle Avoidance. To sum up, the changing rules can be summarized as follow:

\textbf{RL1:} Switching from mode $OP_2$ to reactive mode $OP_1$ at time $ t_* $ when distance $d(t_*)$ reduces to the value $ D $, i.e., $ d(t_*) = d_0$ and $ \dot{d}(t_*) < 0 $.

\textbf{RL2:} Switching back to routine operation mode $ OP_2 $ at any time when vehicle is oriented toward the target, i.e., $ \theta_{fix}=0 $, and the time spent in $ OP_1 $ is larger than a set duration (e.g., 1.5 seconds).

\subsection{Simulation Results}\label{my4s4}
In this section, we simulate forest fire spread on uneven terrains using MATLAB, testing a hybrid navigation algorithm for rescue route planning in disaster zones. The effectiveness of this algorithm for complex terrains is evaluated, focusing on key metrics like path length, planning time, and complexity, compared to earlier reactive methods in \cite{wei2023method}. Further details and insights from these simulations will be explored in upcoming analysis.

\begin{figure}[htbp]
	\centering
	\begin{subfigure}{0.5\columnwidth}
		\includegraphics[width=\textwidth]{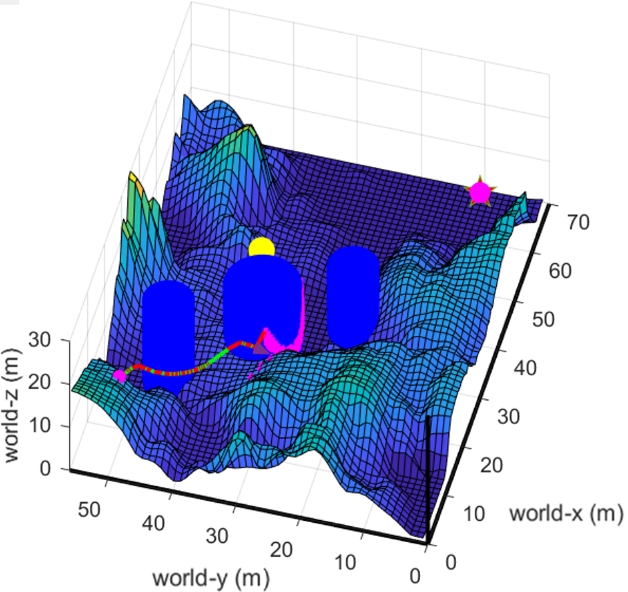}
		\caption{$ T = 10.9\, \mathrm{sec}  $}
	\end{subfigure}%
	\hfill
	\begin{subfigure}{0.45\columnwidth}
		\includegraphics[width=\textwidth]{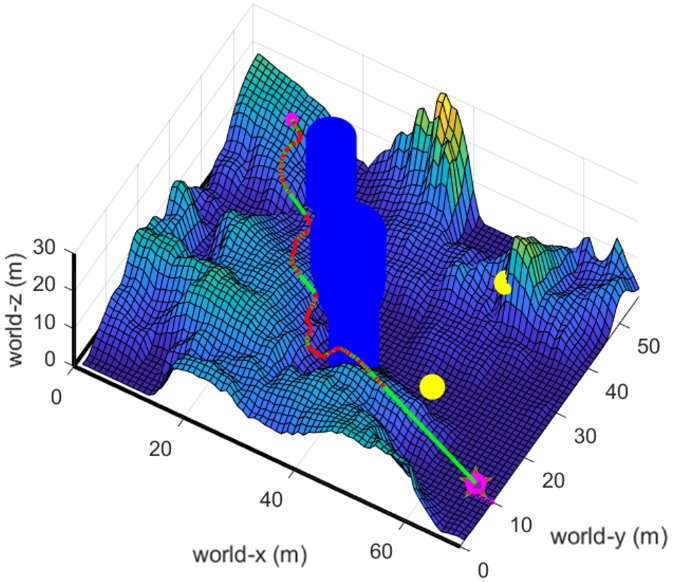}
		\caption{$ T = 35.10\, \mathrm{sec} $ }
	\end{subfigure}\\[0.5ex]
	\begin{subfigure}{0.45\columnwidth}
		\includegraphics[width=\textwidth]{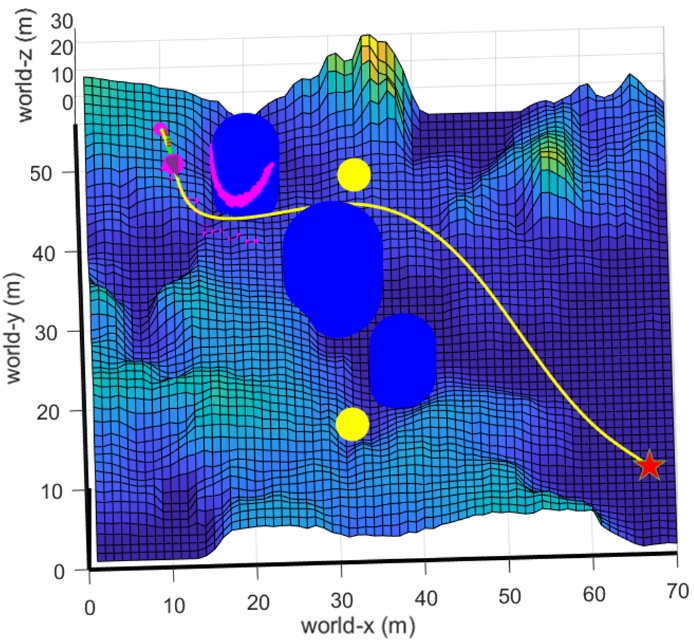}
		\caption{$ T = 1.35\, \mathrm{sec} $ }
		\label{my4case1image3}
	\end{subfigure}%
	\hfill
	\begin{subfigure}{0.5\columnwidth}
		\includegraphics[width=\textwidth]{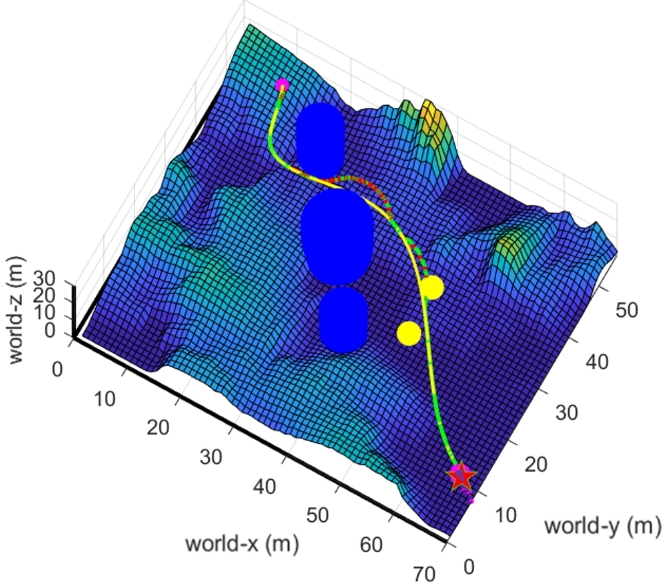}
		\caption{$ T = 26.15\, \mathrm{sec} $ }
	\end{subfigure}
	\caption{The case in forest environment without fire spread: (a) and (b) are Reactive method and (c) and (d) are hybrid algorithm. The UAV predefined trajectory denotes as yellow line, Actual path from UAV denotes as green line and red line.}
	\label{my4case1}
\end{figure}

In this study, we evaluated an algorithm for handling deformable obstacles in two scenarios. The first experiment involved common dynamic obstacles, while the second dealt with deformable obstacles mimicking fire spread. Fig.~\ref{my4case1} shows blue cylinders as static and yellow spheres as moving obstacles. The algorithm uses only start point (pink dot in figure), destination shown as pentagram in the map, and map boundary information in the Reactive approach. Our Path Planning Strategy adds terrain and static obstacles as inputs. Fig.~\ref{my4case1}(c) shows the Deployment Layer's design for a navigation strategy with an optimal path for faster rescue route deployment. The second scenario's fire spread adds complexity to the algorithm. Our previous study \cite{wei2023method} confirmed the effectiveness of a reactive approach in uneven terrains during fires. This experiment compares the time efficiency of our algorithm and its ability to address proximity issues to ground and obstacles, a common challenge in reactive methods. Table \ref{my4table} lists key performance metrics, evaluating the algorithm's flexibility and effectiveness.

In our experiments, we detail the UAV's basic settings. Start points are marked with pink dots, and destinations by pentagrams on the map. Initial, the yellow line in the figures denotes the predefined trajectory from non-uniform sampling RRT algorithm. The UAV's actual path is shown in green and red lines, indicating operation mode $OP_2$ (green) and obstacle avoidance mode $OP_1$ (red). The UAV has a maximum angular speed of $4~\text{rad/s}$ and a maximum linear speed of $4~\text{m/s}$. Safety threshold and sensing radius are set at $3~\text{m}$ and $5~\text{m}$, respectively, with $\alpha_{safe}$ at $\frac{\pi}{5}$. Starting and ending coordinates are [10 51 12] and [67 10 3]. Two mobile obstacles, represented as yellow spheres, are initially at [27.7, 41.2, 9] and [30, 13, 10], with velocities [0.65, -0.8, 0] and [0.65, 0.8, 0]. For Case Study 2, parameters include a fire model with the initial ignition point at [42, 31, 5], wind speed vectors [-8, -8, 0, 1], and fire influence factors $\kappa = 0.05$ and $\rho = 1$.

\begin{figure}[htbp]
	\centering
	\begin{subfigure}{0.5\columnwidth}
		\includegraphics[width=\textwidth]{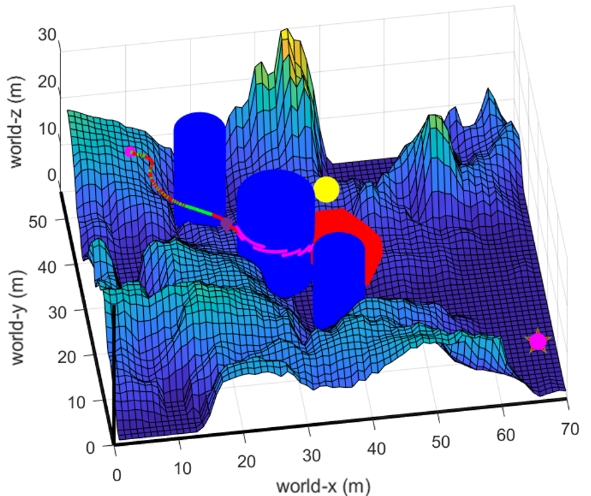}
		\caption{$ T = 9.8\, \mathrm{sec}  $}
	\end{subfigure}%
	\hfill
	\begin{subfigure}{0.45\columnwidth}
		\includegraphics[width=\textwidth]{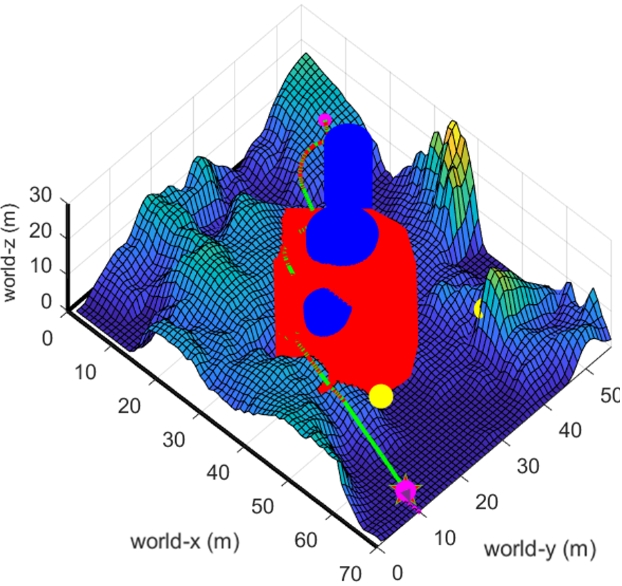}
		\caption{$ T = 35.6\, \mathrm{sec} $ }
	\end{subfigure}\\[0.5ex]
	\begin{subfigure}{0.45\columnwidth}
		\includegraphics[width=\textwidth]{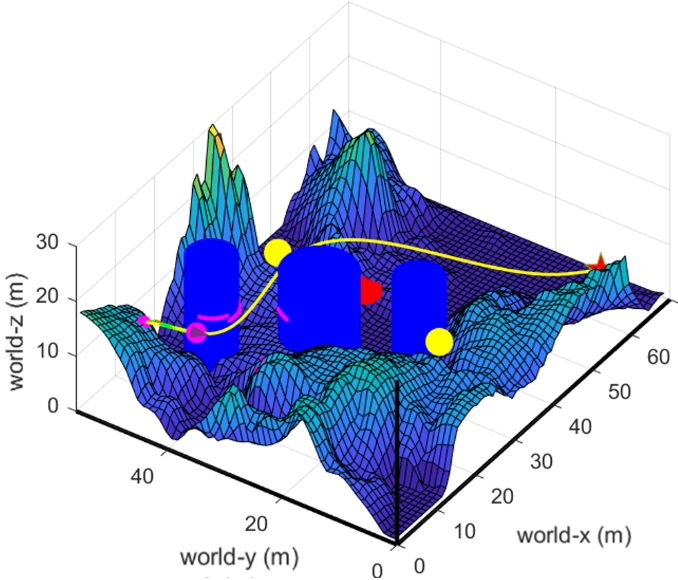}
		\caption{$ T = 1.85\, \mathrm{sec} $ }
	\end{subfigure}%
	\hfill
	\begin{subfigure}{0.5\columnwidth}
		\includegraphics[width=\textwidth]{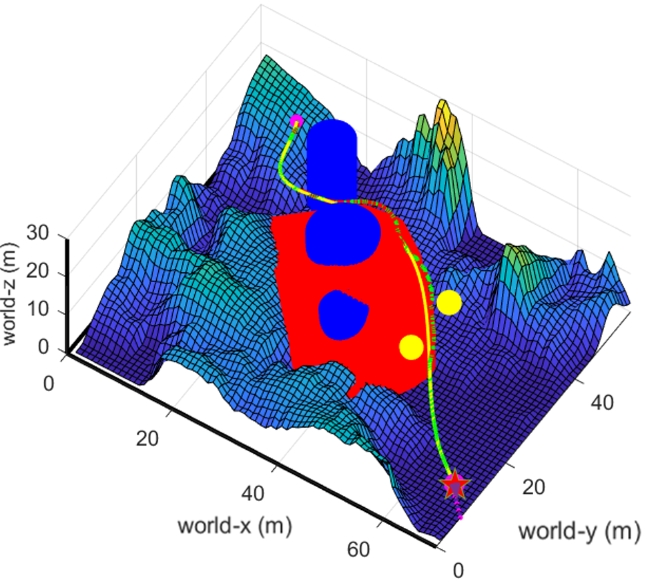}
		\caption{$ T = 28.10\, \mathrm{sec} $ }
	\end{subfigure}
	\caption{The case in forest environment with fire spread: (a) and (b) are Reactive method and (c) and (d) are hybrid algorithm. The UAV predefined trajectory denotes as yellow line, Actual path from UAV denotes as green line and red line.}
	\label{my4case2}
\end{figure}

As illustrated in Fig.~\ref{my4case1}, in a dynamic environment without fire spread, the comparative performance of the purely reactive approach and the proposed path planning algorithm is demonstrated. Table \ref{my4table} illustrates improvements in path length and search time achieved by the hybrid method, along with a significant reduction in computational complexity. This enhancement primarily results from integrating a global path planning component, which lessens the reliance on obstacle avoidance and refines the path efficiency. Fig.~\ref{my4case2} simulates a forest fire scenario, representing a highly complex environment. Despite the increased complexity compared to non-fire situations, there is a noticeable improvement in path length and search time with our method, especially when contrasted with the purely reactive approach.
\begin{table}[h]
	\centering
	\caption{Hybrid and Reactive method analysis}
	\begin{tabular}{c|c|c|c|c}
		\hline
		\multirow{2}{*}{Method} & \multirow{2}{*}{Case} & Path Length & Search Time & \multirow{2}{*}{Complexity} \\ 
		&  & (m) & (sec) &  \\ 
		\hline
		Reactive & Without Fire & 89.80 & 35.10 & 64.00 \\
		Hybrid & Without Fire & 82.87 & 26.15 & 52.13 \\
		Reactive & Fire & 96.4 & 35.60 & 89.36 \\
		Hybrid & Fire & 86.28 & 28.10 & 81.48 \\
		\hline
	\end{tabular} \label{my4table}
\end{table}

\subsection{Conclusion}\label{my4s5}
In conclusion, this study advances the field by introducing a hybrid algorithm built upon previous 3D reactive method research, ameliorating the limitations associated with solely utilizing local algorithms. Experimental results reveal that as the environmental complexity increases, the singular reactive method from prior research exhibits a rising temporal complexity. The hybrid algorithm introduced in this study optimizes this aspect. Moreover, as the environment becomes increasingly intricate, the advantages of the proposed method in terms of path length and search time become progressively more pronounced. Additionally, the section presents a fire propagation model adaptable to uneven terrains, capable of undergoing undefined deformations. Future research may focus on the collaborative operations of multi-unmanned aerial vehicle (UAV) systems, with a particular emphasis on enhancing adaptability, response to contingencies, and the study of fire propagation modes. Additionally, the practical application of theoretical algorithms in this field represents another significant area of interest. Given that these algorithms are primarily intended for safe obstacle avoidance in specific environments, the requirements for external input conditions are relatively limited, making theoretical simulation and replication entirely feasible. However, further research is necessary for the experimental application of these algorithms in actual machines.

\section{An Algorithm for Optimized Navigation of a Network of Cooperating UAVs and UGVs for Bushfire Surveillance and Disaster Relief Missions}\label{mypaper5}
This section aims to address the challenge of rapid bushfire evacuation. Based on the investigation of Chapter 6, guided single UAV to disaster area for safety rescue still exist resource wastage problem and lack og pre-disaster warning. Therefore, in this section, we propose an algorithm for monitoring and navigating catastrophe areas by employing a network of UAVs and unmanned vehicles. This algorithm is specifically built for bushfire alerts, disaster evacuation, and rescue operations. It also incorporates fire modelling to enhance its effectiveness.

\textbf{The partly work is presented in:}
\textbf{J Wei} and Z Fang, An optimal UAV and UGV Cooperative Network Navigation Algorithm for Bushfire Surveillance and Disaster Relief*, In 2024 16th International Conference on Computer and Automation Engineering (ICCAE), pp. 636-641. IEEE, Melbourne, 2024.

\subsection{Introduction}
In recent years, Unmanned Aerial Vehicles (UAVs) have gained prominence in various sectors such as communication, search and rescue (SAR), surveillance, and logistics \cite{li2021networked,savkin2021navigation,savkin2022joint}. Significantly, UAVs are taking over roles traditionally held by humans in hazardous environments. Particularly, in the realm of disaster management and more so in bushfire mitigation scenarios in Australia, the contributions of UAVs are undeniable. Such bushfires not only lead to significant economic implications but also inflict irreparable damages to natural habitats. The deployment of UAVs for surveillance emerges as a pioneering method for the early detection and consequential mitigation of these fire's catastrophic effects. Studies such as \cite{zheng2017forest} and \cite{moinuddin2018simulation} explore these dynamics further. The former leverages cellular automata (CA) for simulation, whereas the latter examines the correlation between the rate of flame spread (RoS) and determining factors like wind speed and grass height.\\
A strategically deployed network of UAVs can offer continuous surveillance over forested areas that are at high risk. Coverage Path Planning (CPP) stands out as an optimal framework for such UAVs, ensuring thorough monitoring, particularly in domains like communication and Search and Rescue (SAR). There is documented success of these algorithms in executing search and rescue operations in disaster-affected areas. For instance, a study by \cite{wan2022accurate} introduces an algorithm anchored in an improved ant colony optimization. This technique is designed to strengthen both global and local search capabilities. The said algorithm refines based on the total flight path length and the associated risk level of the terrain, equipping UAVs to carry out comprehensive searches in disaster zones. Another contribution \cite{savkin2020navigation} delves into the deployment of aerial drones for tracking the evolving boundaries of disaster areas. The paper proposes a computationally efficient sliding mode control algorithm that guides drones. Evidently, this algorithm adeptly tracks the dynamic borders of disaster zones, realizing global improvements in the optimization challenges undertaken. However, achieving both efficient coverage path planning and a rapid response time using a solitary planning algorithm often proves challenging. This has led to the persistent proposition of a hybrid layered algorithm. As said in previous research \cite{wei2023hybrid,elmokadem2018hybrid,ravankar2020hpprm,gul2022implementation}, the employed boundary-following methods for dynamic obstacle avoidance enable drones to navigate safely even in unknown environments. This comprehensive overview highlights the adaptability and effectiveness of local path planning algorithms, particularly in dynamic and unpredictable settings such as disaster zones. \cite{wang2020dynamics} introduces a Dynamic Global-Local (DGL) hybrid path-planning approach that seamlessly integrates global path planning with local hierarchical structures. This results in the optimal generation of sparse waypoints under strictly defined constraint conditions. The methods presented in \cite{wei2023hybrid} and \cite{elmokadem2018hybrid} both adopted the hybrid structure that integrates global and reactive path planning techniques. To cater to the unpredictability and uncertainties inherent in dynamic environments, a local hierarchical structure is integrated. Notably, while this study proficiently addresses the multi-objective optimization dilemma, its accentuation on surveillance and feedback tends to downplay the importance of relocating disaster victims.\\
Evacuating disaster zones necessitates the guidance of victims. Since bushfires are extremely spreading, they can involve a large area in a very short period of time. Previous literature has proposed the use of UAVs to monitor early bushfires, but for evacuation of victims, UAVs are of limited use. Therefore air-ground coordinated autonomous systems can provide inspiration for evacuation of victims. The study from Changsheng introduces a collaborative air-ground target Searching system\cite{shen2017collaborative}. The aerial robot detects and looks for multiple targets in the area and then guides the ground robot to the target location. Another study from Gilhuly converted the UAV-vehicle collaboration problem into a path-planning problem by which UAVs plan paths for unmanned vehicles to reach their destinations\cite{gilhuly2020aerial}. Although numerical optimization is proven to some extent, the significance of their algorithms on a practical level is not addressed and taken into account. However, these studies have inspired the synergistic system that we will propose. \\

In this section, we present an algorithm for disaster area surveillance and rescue navigation based on collaborative work with UAVs and unmanned vehicles in a network designed for bushfire alerts, disaster evacuation and rescue operations with integrated fire modeling. Our UAVs utilize Coverage Path Planning (CPP) within the mission area to ensure an exhaustive assessment of fire dynamics in each sub-area. Once a fire is detected in a high-risk area, the UAV assigned to that area activates a broadcast system to warn people in the neighborhood. The UAV then constructs a path to a pre-determined safe area, expertly guiding the unmanned vehicle to bring threatened people to safety. At this point, the UAVs deployed in each sub-area will be sorted and assigned based on the physical distance of the victims from the flames, planning a pick-up and drop-off path for the unmanned vehicle, which uses the REACTIVE algorithm in the offline positioning space to avoid known and unknown obstacles and evacuate the victims from the fire zone. Meanwhile, we conceptualize the entire surveillance and rescue process as an optimization problem subject to a set of constraints. \\ 

The remainder of this section is arranged as follows: Section \ref{my5s2} will introduce the system model and the problem statement. Section \ref{my5s3} will detail the proposed navigation algorithm. Computer simulations validating the performance of the approach are reported in Section \ref{my5s4}. Lastly, Section \ref{my5s5} will conclude our work.

\subsection{Problem Statement}\label{my5s2}
In this section, each UAV in a UAV network consisting of $k$ UAVs is identical. It is described here using a generalized kinematic model.In three-dimensional space, the direction of motion of the UAV is determined by the $x,y,z$ axis.The following equation describes a single UAV dynamics model with upper and lower bound constraints.
\begin{equation}
	\left\{\begin{array}{l}
		S_u(t)=(x(t), y(t), z(t))\\
		\dot{S_u}(t)=|\vec{v}(t)| \omega(t)
	\end{array}\right. \label{my5modelUAV}
\end{equation}

In this context, $S_u(t) = (x(t), y(t), z(t))$ represents the UAV's location in 3D space, while $v(t)$ and $\omega(t)$ denote the speed and direction of the UAV's movement, respectively. Here:
\begin{itemize}
	\item $S_u(t)$ is a vector in $\mathbf{R}^3$, indicating the UAV's position in three-dimensional space.
	\item $v(t)$ is a velocity vector in $\mathbf{R}^3$ with components $v_x(t), v_y(t), v_z(t)$ representing the UAV's speed in the $x$, $y$, and $z$ directions, respectively.
	\item $\omega(t) \in \mathbf{R}^3$ is a unit vector that indicates the direction of motion, satisfying $\|\omega(t)\| = 1$, where $\|\cdot\|$ denotes the Euclidean vector norm.
\end{itemize}

The constraint on the altitude $z(t)$ of the UAV is given by:

\begin{equation}
	\left\{\begin{array}{l}
		0 < Z_1 \leq Z_2\\
		z_{\min} \leq z(t) \leq z_{\max}
	\end{array}\right.
\end{equation}

Here, the position of the UAV in 3D space, based on a horizontal plane parallel to the ground, is described using $x(t)$ and $y(t)$, while $z(t)$ expresses the height of the UAV relative to the ground.

\begin{figure}[!t]
	\centering
	\includegraphics[width=0.35\columnwidth]{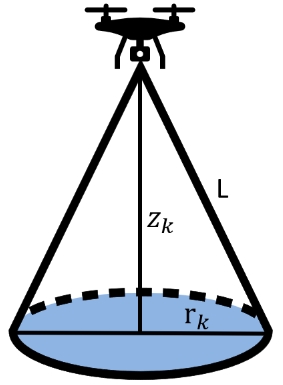}
	\caption{ UAV Model}
	\label{my5fig:subfig1}
\end{figure}

Each UAV is equipped with a camera capable of detecting fire spots. Because the purpose of this section is to develop optimal algorithms to improve the quality of bushfire surveillance as well as the efficiency of personnel rescue, the details of the cameras will not be examined in depth, and by default they are capable of functioning at a safe distance from a bush fire.In addition the UAV is equipped with a horn for vocalization to inform and evacuate the crowd. Again, it has been simplified in the model.Fig.~\ref{my5fig:subfig1} shows the UAV model.

The UAV's field of view takes the shape of a circle parallel to the ground, and the radius of the circle is the UAV's perceived radius, which is the same as the altitude when the observation angle is fixed. The relationship is given by:
\begin{equation}
	\tan{\alpha}=\frac{{r_k}}{{z_k}}
\end{equation}

In this study, UGVs and UAVs are integrated into a single model, with $ S_g $ representing UGV's location in 3D space, applying the formula (\ref{my5modelUAV}) as a substitute for the unmanned terrain vehicle model. The $ z-axis $ of the UGV is determined by the current terrain height, and we also apply the coordinate transformation method from Chapter 6 to align the UGV’s body coordinate system with the ground coordinate system.

In this section, the rescue of a bushfire is interpreted as a hierarchical optimization problem. In the proposed mountainous area, the region is pre-divided into $m$ sub-areas, each of which is covered by a UAV, so that the mathematical expression of the area is $D^m_{k_{n}}$, where $m \in M$, and $n = 1,\dots,k$. The $k$-th UAV performs full-coverage path planning in the area it is responsible for and accomplishes maximum coverage to surveil bushfires that may occur in the area.

\begin{figure}[!t]
	\centering
	\includegraphics[width=0.8\columnwidth]{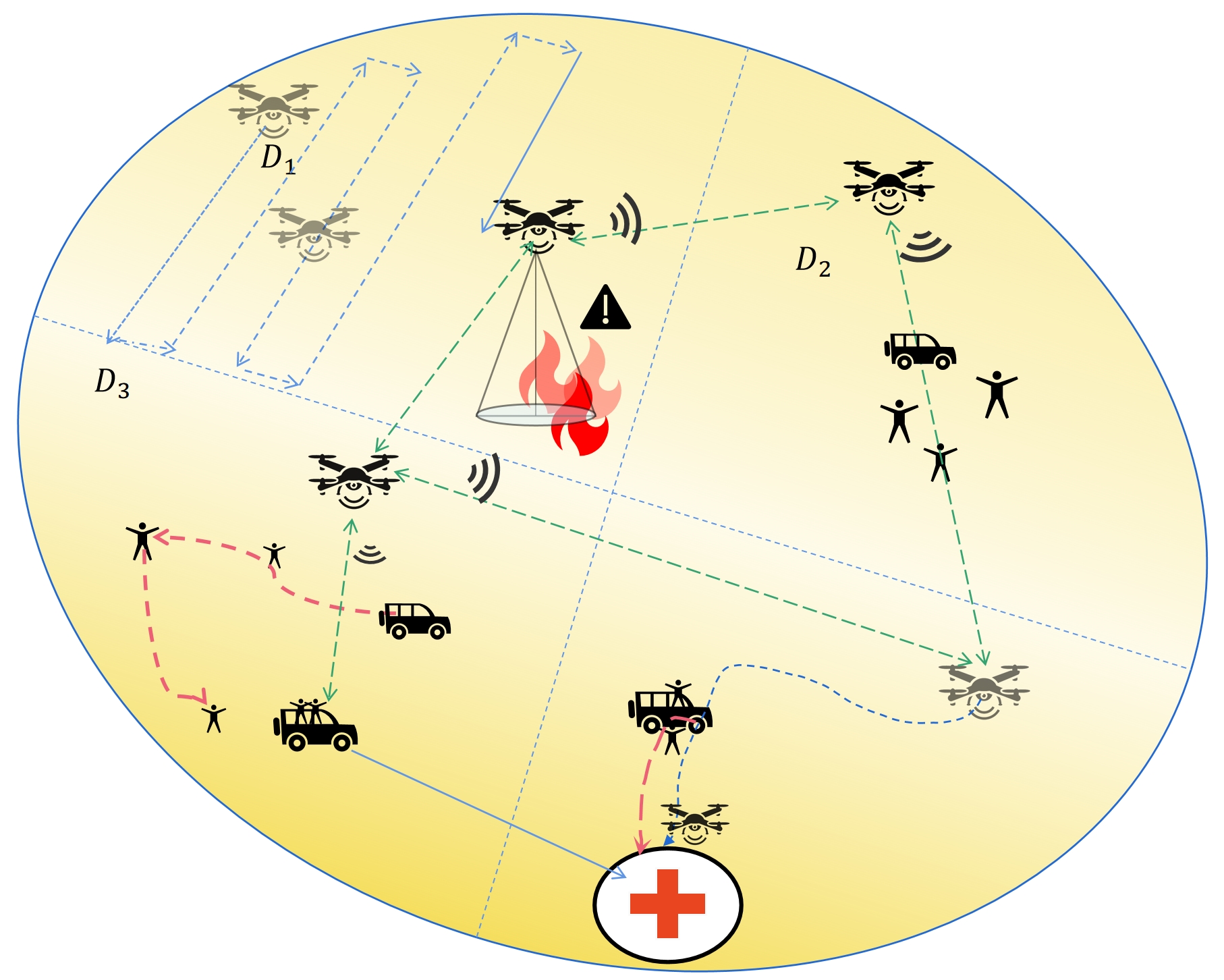}
	\caption{Bushfire Surveillance and Evacuation}
	\label{my5fig:subfig}
\end{figure}

When any UAV detects a bush fire in the sub-area $D^m_{k_{n}}$ for which it is responsible, that UAV not only needs to pass the fire information to other UAVs in a timely manner, but also needs to immediately move to the second tier of the mission - communicating and transporting. The UAV first transmits the gathered information to the base station, which then determines the rescue mission for the UGV. The UGV will use reactive planning paths to transport people to the predetermined safety zones as efficiently as possible. The execution of this hierarchical task is illustrated in detail by Fig.~\ref{my5fig:subfig}.

This dual requirement poses several challenges, we construct a mathematical optimization framework that takes into account UAV flight constraints, safety requirements, and surveillance objectives.

We introduce \(d_{s,k}\) as the distance from the victims within the area covered by the \(k\)-th UAV to the safety zone, and \(z_k\) represents the altitude of the \(k\)-th UAV, which corresponds to the sensing radius \(r_k\) of the UAV at that time. The optimization problem can be described as follows:

\textbf{Minimizing UAV Energy Consumption \(Q_k\):}

The main objective is to minimize the UAVs' energy consumption during the mission. This is achieved by optimizing the flight paths and operational decisions, such as altitude and trajectory planning. The decision variables include the UAV’s flight path and operational strategy, with the aim of finding the most energy-efficient routes that cover the required areas while avoiding unnecessary maneuvers and reducing energy expenditure. Mathematically, the total energy consumption \(Q_k\) for each UAV \(k\) can be expressed as:

\begin{align}
	Q_k = \int_{0}^{T_k} P(\text{path}_k(t), z_k(t)) \, dt
\end{align}

where \(P(\text{path}_k(t), z_k(t))\) represents the power consumption function dependent on the flight path and altitude over time, and \(T_k\) is the total mission time for UAV \(k\). The goal is to minimize total energy consumption by optimizing these variables:

\begin{align}
	\text{minimize} \quad \sum_{k=1}^{K} Q_k
\end{align}

where minimization in (7.5) is done over all possible flight paths of UAVs that satisfy the operational constraints, such as coverage requirements and no-fly zones" aligns with typical UAV operation constraints.

In summary, the focus is on minimizing UAV energy consumption to ensure that UAVs can sustain continuous surveillance and operational efficiency throughout the mission. This optimization forms a critical component of the overall emergency response strategy.

\subsection{The Algorithm for Navigation of a Team of UAVs and UGVs}\label{my5s3}
In this section, we describe the navigation algorithm for the UAV-UGV team. In this system architecture, the ground station serves as the central control node for the collaborative network of UAVs and UGVs. The ground station is responsible for coordinating the collaboration between UAVs and UGVs, providing environmental information to UGVs via UAVs patrolling the map. During rescue missions, the ground station not only assigns rescue tasks to UGVs, but also determines the pairing between UAVs and UGVs to ensure efficient task execution.

The environment is divided into multiple segments using the K-means method (e.g., dividing environment into four parts in Case 3 of section \ref{my5s742}). The K-means algorithm was employed to divide the monitoring area based on terrain features and task requirements. The K-means algorithm minimizes the variance within each cluster (in this case, sub-regions), effectively partitioning the map into several areas. Each sub-region is assigned to a UAV and a UGV, ensuring that any detected fire can be responded to quickly.   Each segment being assigned to one UAV and one UGV for joint operation. The ground station is essential for facilitating data exchange, as the UGVs rely on the UAVs for real-time environmental and target information. During patrols, the UAV applies the coverage path planning method from section \ref{my5s731} to perform collision-free patrols across the map. During this time, the UAV collects ground information and is equipped with sensors to detect potential fire hazards in the environment. Once a fire is detected, the UAV sends information (such as rescue point coordinates) to the ground station, and the UGV will utilize the reactive path planning method from section \ref{my5s732} to perform rescue operations for the known rescue targets on the ground.

Additional, the environmental model required in this section also relies on the forest fire spread model, which remains a crucial component. To ensure the accuracy and consistency of the simulation, this section applies the same mathematical model that was thoroughly introduced in section \ref{mypaper4}, Section \ref{my4Forestfirespreadmode}. Since the principles and applications of the model remain unchanged in this context, the detailed mathematical derivation and description will not be repeated here. Readers are referred to section \ref{mypaper4}, Section \ref{my4Forestfirespreadmode} for more information. This approach helps maintain the coherence of the content while minimizing unnecessary repetition.

This simple simulation method, based on area division and priority allocation, effectively optimizes the UGVs' evacuation time and path length while ensuring continuous monitoring of the entire area by the UAVs. Although this approach simplifies the complexity of a multi-UAV and multi-UGV system, it provides a sufficient basis for the initial validation of our strategies and models.

\subsubsection{Coverage Path Planning}\label{my5s731}
The Coverage path planning algorithm proposed in this paper employs a similar idea to the lawnmower algorithm to ensure full coverage of the region $D_k$ under its responsibility (inspired by \cite{savkin2021asymptotically}). Where the region $D_k$ is a closed bounded region with continuity. The flight path $p_i$ of the UAV is a straight line parallel to the ground. However, considering the uneven forest terrain, different flight altitudes of the UAV may lead to obstacles during patrol, so we use the reactive method described in section \ref{my5s732} for obstacle avoidance. The flight path $p_i$ divides the region $D_k$ into smaller closed regions $D_{k_n}$. The flight path $p_i$ of the UAV moves along the boundary of the region $D_k$, and when it reaches the bottom vertex it reorients itself and moves 90 degrees laterally to the next boundary point, and then later on makes a 90 degree rotation to continue flying in a straight line segment. This is repeated until the area $D_k$ is completely covered. When the UAV finds the fire point, then it quickly communicates with other UAVs with the UGV for the rescue phase of the mission. The UAV will create a direct path to the safe zone for the UGV. The UGVs will pick up the victims in order of danger level and then follow the path created by the UAVs and implement lower-control reactive path planning for obstacle avoidance and rapid evacuation.

\subsubsection{Rective method}\label{my5s732}
In this section, the reactive method introduced is applied to both UAV and UGV models. For UAVs, obstacles may arise during patrols due to varying flight altitudes. When employing the reactive method described in this section for obstacle avoidance, the next waypoint is dynamically set as the target, and the UAV continues avoiding obstacles until the path is clear. Once the obstacles are cleared, the UAV reverts to the patrol method outlined in section 7.3.1. For UGVs, path planning is carried out by prioritizing victims based on the severity of their danger level, with each victim sequentially set as a target point, and the final target being the designated safe zone. To avoid complexity, this section primarily focuses on the UGV when explaining the method.

The reactive methodology draws inspiration from the works presented in \cite{elmokadem2018hybrid} and \cite{wei2023hybrid}. Specifically, the research in \cite{wei2023hybrid} has successfully implemented the reactive component for optimal boundary path generation in a 2D environment. The methodology proposed in \cite{elmokadem2018hybrid} suggests the feasibility of extending this reactive approach to transform the coordinate frame into the obstacle plane. We proceed to delineate this enhanced reactive method in the subsequent sections.

In the process of reactive path following, two operational modes are defined. Mode 1 is designed to maintain the UGV's orientation consistently directed toward the target, while Mode 2 ensures collision-free navigation during execution. Upon activation of Mode 2 at time \( t_{*} \), a rotational transformation becomes essential to map the ground coordinate frame into the body coordinate frame. The matrix for coordinate conversion and the associated transformation technique are articulated in equations (\ref {my5rotation1}) and (\eqref{my5rotation2}), respectively.

\begin{align}
	M=
	\begin{bmatrix}
		\sin \alpha \cos \beta & \cos \alpha \cos \beta & \sin \beta \\
		-\sin \alpha & \cos \alpha & 0 \\
		-\sin \alpha \sin \beta & -\cos \alpha \sin \beta & \cos \beta
	\end{bmatrix} \label{my5rotation1}
\end{align}
\begin{align}
	\begin{bmatrix}
		\begin{array}{c}
			R_{i} \\
			\hdashline  1
		\end{array}
	\end{bmatrix}
	= \begin{bmatrix}
		\begin{array}{c:c}
			{ }_g^u M & { }^g R_{st} \\
			\hdashline 0 & 1
		\end{array}
	\end{bmatrix}
	\times
	\begin{bmatrix}
		\begin{array}{c}
			{ }^u R_{go} \\
			\hdashline  1
		\end{array}
	\end{bmatrix} \label{my5rotation2}
\end{align}
In equation (\ref{my5rotation1}), the variables \( \alpha \) and \( \beta \) denote angles originating from distinct \( x \)-axes. The notation in equation (\ref {my5rotation2}) signifies coordinates within different reference frames. For instance, \( R_{i} = \left[x_{st}, y_{st}, z_{st}\right]^T \) indicates the current position in the ground coordinate system, while \( {}^u R_{go} = \left[x, y, z\right]^T \) represents the spatial location post-conversion in the body coordinate system.

Now, we will encapsulate the functionalities of Mode 1 and Mode 2 in the context of the reactive method. Mode 1 is engineered with the primary goal of ensuring that the UGV perpetually moves in the direction of the target. The mathematical formulation for Mode 1 is presented in Equation (\ref{my5mode1}), as delineated below. 
\begin{align}
	\begin{array}{l}
		u(t) = U^{max}sgn(\theta_{\Delta }) \\
		v(t) = V^{max}
	\end{array}   \label{my5mode1}
\end{align}
In this mode, the variable $ \theta_{\Delta } $ represents the difference between the current orientation of the UGV and the orientation towards the target point. Serving as the input to the $ sign $ function.  This ensures that the UGV can constantly correct its direction to always face the target point.

In mode 2, it will follow this real-time obstacle avoiding law to do reactive.
\begin{align}
	\begin{array}{l}
		u(t) = -U_{max}f(v_{new}(t),v_{u}(t)) \mathop{i_{n}} \limits ^{\rightarrow} (t) \\
		v(t) = ||v_{new}(t)||,\\
	\end{array} \label{my5mode2}
\end{align}

where $ \mathop{i_{n}} $ is an unit vector perpendicular to $ \widetilde a (t) $ directing away from the surface of obstacle. The velocity of $ v_{new} (t) $ is the desired velocity in the next moment. It comes from:
\begin{align}
	\left\{\begin{array}{l}
		\angle \zeta 1(t) =  \cos(\beta^i(t)), \\[1ex]
		\angle \zeta 2(t) =  \sin(\beta^i(t)), \\[1ex]
		h^i(t) = \bigtriangleup V(t)[\angle \zeta 1(t), \angle \zeta 2(t)], \\[1ex]
		v_{new}(t) = v_x(t) + h^{i}(t)
	\end{array}\right.  \label{my5calculatemode2}
\end{align}
where the $ \beta^i(t), i = 1,2 $ represents the enlarged avoidance angle. The constant enlarged angle $ \alpha_0 $ is introduced to guarantee safe evasion.
The variable $ \bigtriangleup V(t) $ represents the difference between the current speed of the unmanned ground vehicle and the surface speed of the detected unknown obstacle. $ h^i(t) $ shows the two collision lines from the obstacle. Finally, we will calculate our angular velocity $ u(t) $ in equation (\ref{my5mode2}) by means of a deformed sign function $ F(\dot,\dot) $.
\begin{align}
	f(a_1,a_2) = 
	\left\{\begin{array}{ll}
		0, & if \quad \varphi(a_1,a_2) =0 \\
		1, & if \quad 0< \varphi(a_1,a_2) \leq \pi \\
		-1, & if \quad -\pi < \varphi(a_1,a_2) < 0
	\end{array}\right.
\end{align}
In addition, there is a switch rule to decide the execution mode. In the framework of this conversion rule, both the steering angle of the unmanned ground vehicle and the distance at which obstacles are detected serve as crucial parameters for evaluation. The principal components of the conversion rule are detailed below.

\textbf{\textit{R1}}: Switching from mode 2 to mode 1 occurs at any time $t_0$ when the distance from the robot to the obstacle $i$ reduces to the value $C$, i.e., $d_i\left(t_0\right)=C$ and $\dot{d}_i\left(t_0\right)<0$

\textbf{\textit{R2}}: Switching from mode 1 to mode 2 occurs at any time $t_*$ when $d_i\left(t_*\right) \leq d_{\varepsilon} + \kappa (\kappa >0)$ and the vehicle is oriented towards the final destination point, i.e., the vehicle velocity vector $v(t)$ is co-linear to the heading to the final destination point $H\left(t^*\right)$.

\subsection{Simulation Result}\label{my5s4}
\subsubsection{Simulation with Single UAV and  UGV}
\textit{Case 1: Simulation of Single UAV  and  UGV in simple case}

\begin{figure}[htbp]
	\centering
	\begin{subfigure}{0.45\columnwidth}
		\includegraphics[width=\textwidth]{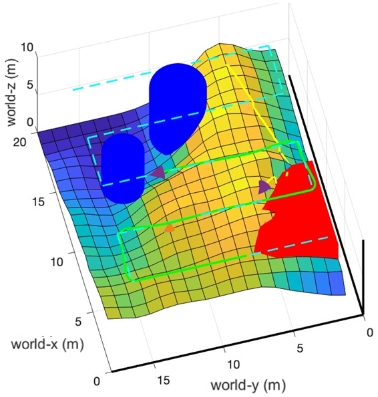}
		\caption{$ T = 11.41\, \mathrm{sec}  $}
	\end{subfigure}%
	\hfill
	\begin{subfigure}{0.4\columnwidth}
		\includegraphics[width=\textwidth]{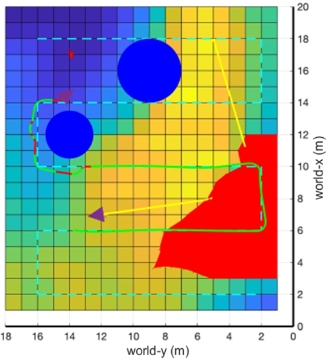}
		\caption{$ T = 15.2\, \mathrm{sec} $ }
	\end{subfigure}\\[0.5ex]
	\begin{subfigure}{0.5\columnwidth}
		\includegraphics[width=\textwidth]{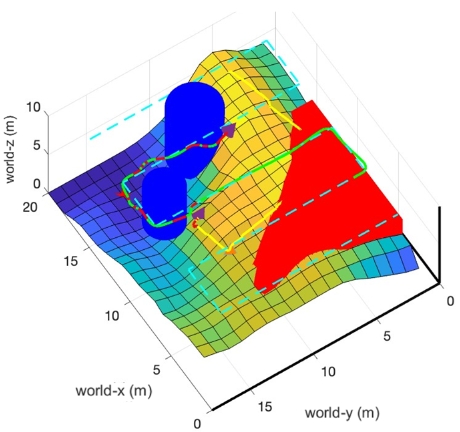}
		\caption{$ T = 17.7\, \mathrm{sec} $ }
	\end{subfigure}%
	\hfill
	\begin{subfigure}{0.5\columnwidth}
		\includegraphics[width=\textwidth]{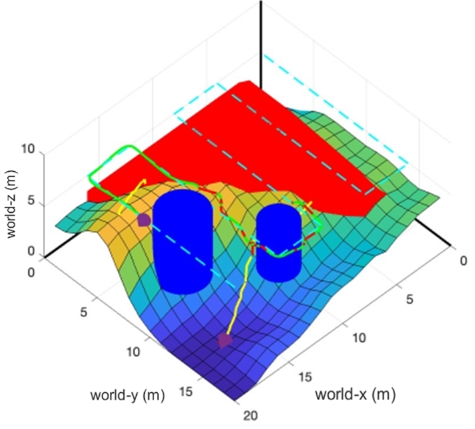}
		\caption{$ T = 21.85\, \mathrm{sec} $ }
	\end{subfigure}
	\caption{Proposed UAV/UGV cooperation algorithm simulation of simple scenario of hill fire rescue: Blue cylinders are static obstacles, red moving obstacle simulates the spread of fire.  Green line is the UAV execute path, blue dash line is the reference path from CPP. The yellow line is the trajectory from ground vehicle. $ T $ represents the total time}
	\label{my5cpp0fire}
\end{figure}

In this section, the algorithm proposed in this chapter will be validated through simulations using MATLAB software. The experiments are conducted by integrating the UAV and the unmanned ground vehicles. Initially, the UAV follows a reference path generated by the CPP algorithm. The UAV also performs obstacle avoidance during its path-following process.  During its flight, the UAV updates the map and inputs the geographical locations of targeted individuals. Moreover, the UAV undertakes environmental monitoring and activates the UGV when a fire event is detected within the mapping area. Employing a priority-oriented list of individuals in need of assistance, the UGV adopts a reactive algorithm to gerenate an effective rescue trajectory with a yellow line. Upon successfully transporting all designated individuals for rescue, the UGV returns to a pre-identified safe point on the map.  The initial point of the unmanned vehicle is [18 5 6.5] and the safe point also as the end point is [17 14 0]. The two targets to be rescued in this environment are [9.5 2.5 4.3] and [7 12 5.39]. It can be seen in Fig.~\ref{my5cpp0fire} that the car rescue is still very successful in small maps.
\begin{figure}[htbp]
	\centering
	\includegraphics[width=0.8\columnwidth]{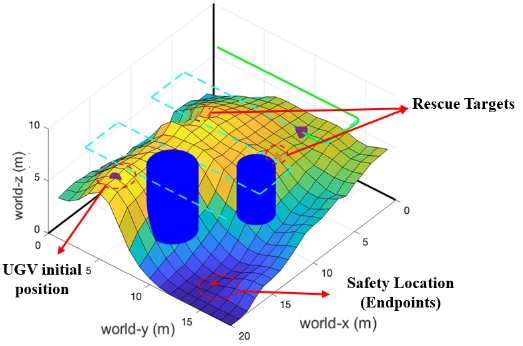}
	\caption{Single UAV  and  UGV cooperation in simple case: Green line is the UAV execute path, blue dash line is the reference path from CPP}
	\label{my5cpp0nofire}
\end{figure}

\textit{Case 2: Simulation of Single UAV and UGV in complex case}

\begin{figure}[htbp]
	\centering
	\begin{subfigure}{0.45\columnwidth}
		\includegraphics[width=\textwidth]{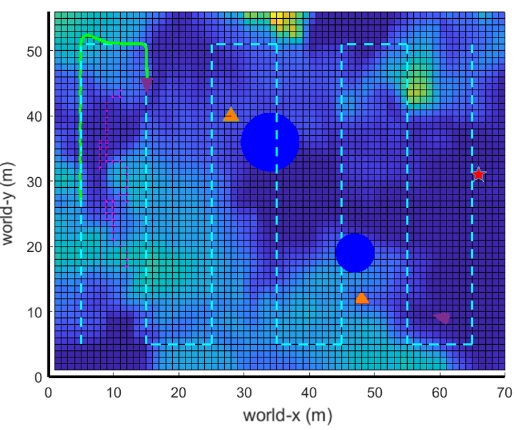}
		\caption{UAV patrol status (Top view)}
	\end{subfigure}%
	\hfill
	\begin{subfigure}{0.5\columnwidth}
		\includegraphics[width=\textwidth]{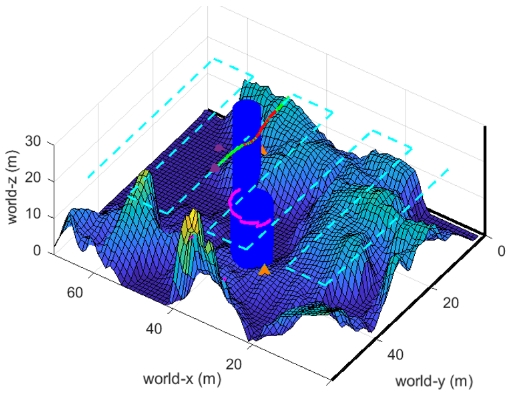}
		\caption{UAV patrol status (3D view)}
	\end{subfigure}
	\caption{Simulation of UAV patrol status without wildfire occur: Orange triangles represent needed to be rescued targets in the future.}
	\label{my5Figr1}
\end{figure}

\begin{figure}[htbp]
	\centering
	\begin{subfigure}{0.45\columnwidth}
		\includegraphics[width=\textwidth]{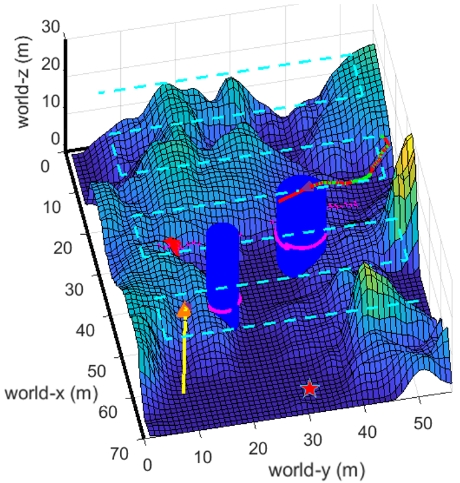}
		\caption{$ T = 52.25\, \mathrm{sec}  $}
	\end{subfigure}%
	\hfill
	\begin{subfigure}{0.5\columnwidth}
		\includegraphics[width=\textwidth]{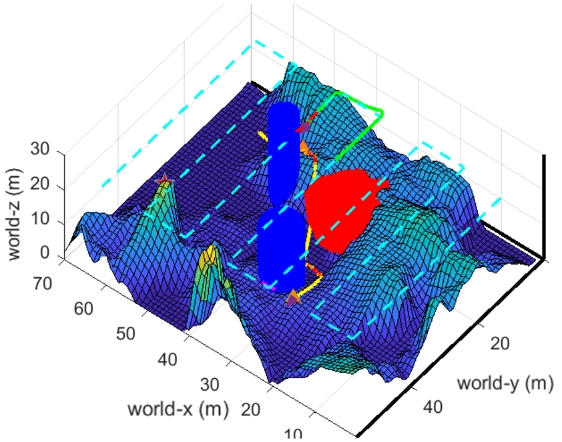}
		\caption{$ T = 68.5\, \mathrm{sec} $ }
	\end{subfigure}\\[0.5ex]
	\begin{subfigure}{0.5\columnwidth}
		\includegraphics[width=\textwidth]{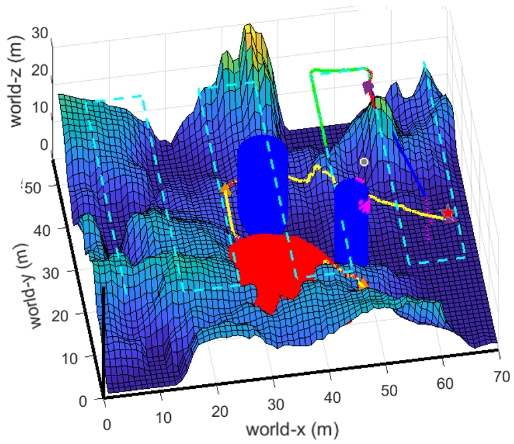}
		\caption{$ T = 85.1\, \mathrm{sec} $ }
	\end{subfigure}%
	\hfill
	\begin{subfigure}{0.45\columnwidth}
		\includegraphics[width=\textwidth]{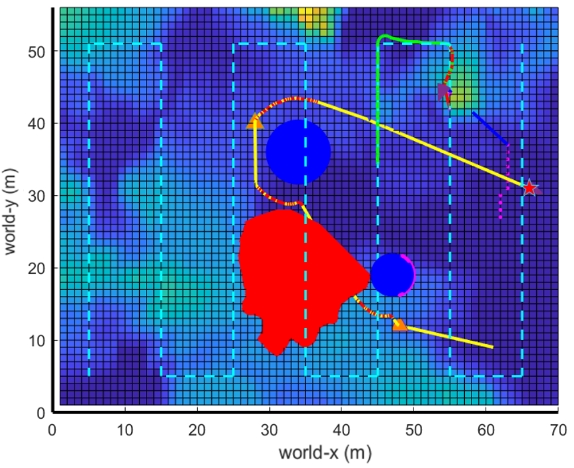}
		\caption{$ T = 85.1\, \mathrm{sec} $ }
	\end{subfigure}
	\caption{Simulation of single UAV and single UGV cooperation for forest fire rescue path planning. The fire starts in the 48 sec. The execute path of UAV represents green line and the UGV path represents as orange line. $ T $ represents the total time}
	\label{my5Figr2}
\end{figure}

Fig.~\ref{my5Figr1} illustrates the initial map where the UAV conducts environmental exploration based on the reference path. The reference path is represented as a red dashed line in the simulation, and the green-red line from the UAV denotes the path-following with no collision process. In the map, two orange triangles in the environment represent the rescue targets. The vehicle is shown as a purple cone here. The rescue path from the vehicle will be represented by a yellow line. Two blue cylinders will exist as unknown obstacles. Now, the initial setting in this experiment will be introduced. The maximum angular velocity of the UAV is \( U^{max} = 4 \, \text{rad/s} \), and the maximum linear velocity is \( V^{max} = 4 \, \text{m/s} \). The obstacle-avoidance angle is set at \(\alpha = 3 \, \text{rad} \). The cruising altitude of the UAV is set at 20 meters.
The initial position of the ground vehicle is denoted as \([61,9,0]\), and its maximum angular and linear velocities are  \(  4 \, \text{rad/s} \) and \( 3 \, \text{m/s} \), respectively. Safe coordinates within the environment are marked at \([66,31,2]\). Two rescue targets exist in this setup, located at \([48,12, 7.2]\) and \([28,40,4.2]\), respectively. These coordinates serve as inputs upon the activation of the ground vehicle. The initial point of the fire outbreak is at \([29,10,5]\), with wind speeds at \([8,8,0]\). It is known that a fire will erupt in the environment \( 47.5 \) seconds after the start. Based on these initial settings, the resultant plot will be depicted in Fig.~\ref{my5Figr2}.

In Fig.~\ref{my5Figr2}, the status of the simulation at various time instances is depicted. The ground vehicle departs at 52.25 seconds to reach the first rescue target, arrives at the second rescue target at 68.5 seconds, and finally reaches the designated safe point at 85.1 seconds.   Throughout the simulation, the UAV conducts coverage complete path planning as well as real-time local obstacle avoidance. Under the information provided by the UAV, the ground vehicle is deployed for multi-target point reactive path planning. The final goal of ensuring a successful and safe fire rescue operation is consequently achieved.Based on the previous description of homogeneity, our algorithm is scalable, and when the map size or the number of operating UAVs and UAVs increases, our algorithm is still able to meet the needs of the task and achieve optimality.

\subsubsection{Case 3: Simulation with four multiple and  UGVs}\label{my5s742}
\begin{figure}[htbp]
	\centering
	\begin{subfigure}{0.4\columnwidth}
		\includegraphics[width=\textwidth]{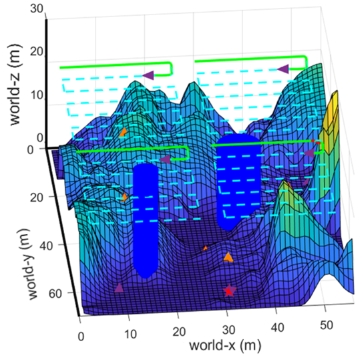}
		\caption{Routine patrol status}
	\end{subfigure}%
	\hfill
	\begin{subfigure}{0.5\columnwidth}
		\includegraphics[width=\textwidth]{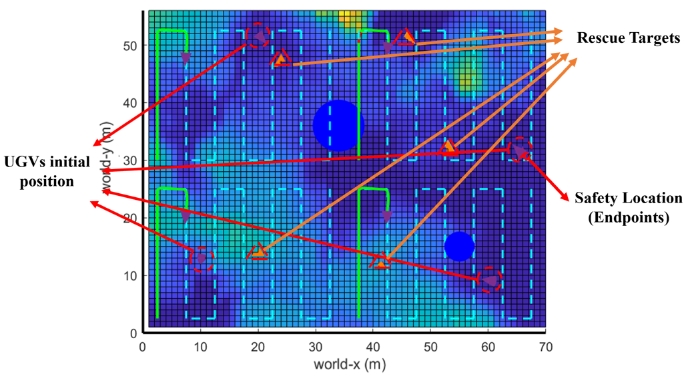}
		\caption{Initial parameters explanation}
	\end{subfigure}
	\caption{Simulation of multiple UAVs and UGVs cooperation for forest fire rescue path planning. Total map is divided into four parts with one drone and unmanned vehicle in each part to collaborate}
	\label{my5cpp1nofire}
\end{figure}

Although the task in the complex environment can be accomplished using a single drone or unmanned vehicle, this approach is often time-consuming. Consequently, we explored the strategy of combining multiple systems of drones and unmanned vehicles for forest fire rescue missions in complex map environments. This method involved segmenting the map using the K-means algorithm.  The initial application of this strategy is depicted in Fig.\ref{my5cpp1nofire}. In this case,  we divide the map into 4 parts based on the K-means algorithm. Each segment is equipped with a UAV and UGV, for a total of four UAVs and UGVs working together.

\begin{figure}[htbp]
	\centering
	\begin{subfigure}{0.5\columnwidth}
		\includegraphics[width=\textwidth]{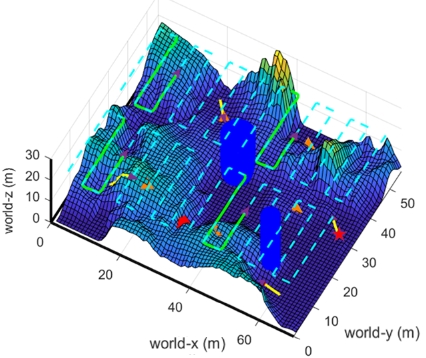}
		\caption{$ T = 16.75\, \mathrm{sec}  $}
	\end{subfigure}%
	\hfill
	\begin{subfigure}{0.4\columnwidth}
		\includegraphics[width=\textwidth]{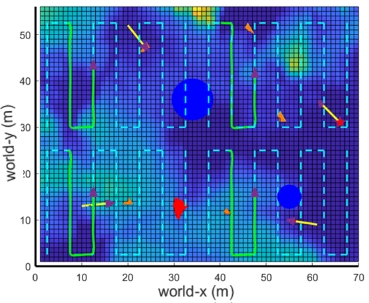}
		\caption{$ T = 16.75\, \mathrm{sec} $ }
	\end{subfigure}\\[0.5ex]
	\begin{subfigure}{0.5\columnwidth}
		\includegraphics[width=\textwidth]{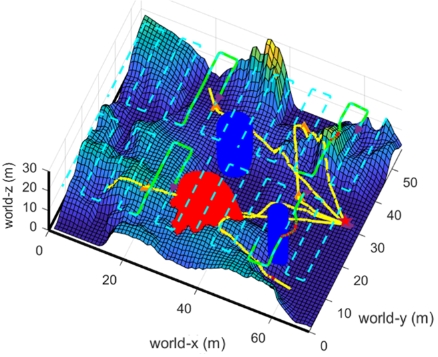}
		\caption{$ T = 38.8\, \mathrm{sec} $ }
	\end{subfigure}%
	\hfill
	\begin{subfigure}{0.4\columnwidth}
		\includegraphics[width=\textwidth]{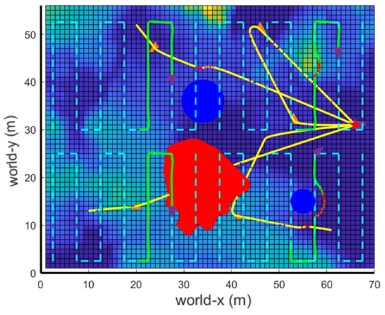}
		\caption{$ T = 38.8\, \mathrm{sec} $ }
	\end{subfigure}
	\caption{Simulation of Forest Fire Rescue with multiple UAVs and UGVs system: (a) and (c) are 3D views, (b) and (d) are top views. The fire starts in the 12 sec. The execute path of UAV represents green line and the UGV path represents as orange line. $ T $ represents the total time}
	\label{my5cpp1fire}
\end{figure}

The simulation results shown in Fig.~\ref{my5cpp1fire} demonstrate that the method significantly improves the effectiveness of the system using only a single UAV. The whole mission took a total of 38.8 seconds, while the rescue times of the unmanned vehicles in each partition were 33.25, 34.70, 35.75 and 36.80 seconds, respectively. Rapid response and route selection away from the fire source were the two key factors for the success of this rescue mission.

\subsection{Conclusion}\label{my5s5}
In this study, we propose a UAVs and UGVs collaborative planning scheme that utilizes the characteristics of UAVs to monitor bushfires and execute full-coverage path planning algorithms while simulating a bushfire model. When the fire is detected and the UAV operates cooperatively, multiple path planning algorithms are applied to move the victims to a safe area. Computer simulations show the algorithm's feasibility, and this study is instructive for bushfire surveillance and rescue.

Given that this work represents an initial attempt at integrating a multi-UAV and multi-UGV system, a simplified approach was employed where all information is centralized at a ground station, which then disseminates environmental data to both UAVs and UGVs. This approach was intended to validate the fundamental strategies before advancing to more complex, decentralized cooperation models. Further research will focus on developing and refining the cooperation models between UAVs and UGVs, and this will be explored in future work.

\section{Conclusion and Future Directions}\label{mypaperend}

This report has explored various complex navigation strategies for the  UAV and UGV. In conclusion, this section summarizes the key findings of the report and offers recommendations for future research directions.

\subsection{Contributions summary}

In this report, we investigate approaches to guarantee safe, collision-free navigation in various kinds of challenging situations. The low processing complexity of our suggested methods makes them especially appropriate for deployment under challenging surroundings, especially for rescue operations. We have created a thorough framework for avoiding barriers predicated on categorizing different challenges. Additionally, this report offers a thorough synopsis of its contributions, emphasizing the significant developments in autonomous navigation under challenging circumstances.
	
Section \ref{mypaper0} provides a comprehensive overview of the advancements in autonomous navigation algorithms and their applications in UAVs. It includes an extensive survey of the current literature on UAV-related perception, motion control, and collision avoidance challenges. Through an in-depth analysis of existing research, this work uncovers key technological advancements in UAV navigation, especially in terms of efficient path planning and precise obstacle avoidance strategies in complex terrain. Moreover, the report discusses the potential applications of UAVs in emergency rescue missions, particularly emphasizing their significance in search and rescue operations in inaccessible or high-risk areas. Through these studies, the report aims to provide a theoretical foundation and practical guidance for the broader application of UAV technology.

Section \ref{mypaper1} proposes a hybrid planning method to solve the navigation problem in a partially known closed environment filled with several stationary and moving obstacles. The RRT-connect method use prior knowledge to generate an initial required reference waypoints as reactive operation inputs of motion planner. In addition, an improved switcher control method further enhances the stability and efficiency of the algorithm.  However, the objective of this research is limited to the applications inside 2D settings that consist of regular shape obstacles, such as circles, squares and rectangles. Further developments and depth studies is presented in the subsequent chapters.

Section \ref{mypaper2} based on previous results, presents a real time obstacle avoidance algorithm for mobile robots also applying in 2D dynamic environments. This research primarily focuses on applying obstacle avoidance algorithms to deformable obstacles. By adopting a multi-layered, hierarchical control approach, we effectively overcome the limitations of previous algorithms that were only suitable for regular environments. The addition of a replanning phase in the decision layers further enhances the completeness of the algorithm. When the vehicle detects excessively large avoidance angles, the triggering of the replanning mechanism not only increases the effectiveness of the algorithm but also reduces the energy consumption of the drone. In terms of simulation result validation, we still employed two-dimensional maps and simplified the fire propagation model to an extension of deforming obstacles. Compared to the singular reactive method, the optimized algorithm proposed in this chapter performs better in terms of safety, deploying paths that are farther away from dynamic fire sources and have smaller turning angles, aligning more closely with the objectives of the research problem.

Section \ref{mypaper3} presents a novel 3D reactive navigation method for UAV addresses the problem in uneven terrain navigation without collision. This algorithm employs the coordinate conversion matrix to facilitate the transformation between ground and drone body coordinate systems. When faced with obstructions, the drone can swiftly determine an avoidance plane, facilitating rapid simulation to navigate to the desired destination in a 3D dynamic scenario without any collisions. The simulation results provide evidence of the algorithm's performance in cases involving simple uneven terrain and elaborate forest landscapes derived from real-world data.

Section \ref{mypaper4} aims to simulate the adoption of drones to execute rescue missions in forest fires through the application of path planning methodologies. We propose a base model that more accurately represents the spread of real forest fires and applies it in uneven terrains with static and dynamic obstacles, utilizing a hybrid navigation method for rescue efforts. To ensure the superiority of the drone-deployed rescue routes, the first two cost functions proposed in this chapter enhance the optimization of the path trajectory, while the adoption of a disaster coefficient in the third cost function reduces the risks associated with uneven terrain and proximity to the fire source. The effectiveness of the algorithm is validated through comparative experiments, considering factors such as environmental complexity, path deployment time, path length, and algorithm complexity.

Section \ref{mypaper5} implements a multiple-system framework designed for application in fire rescue missions. This architecture effectively addresses the previously found deficiency of pre-disaster detection. The combination of Unmanned Aerial Vehicles (UAVs) and Unmanned Ground Vehicles (UGVs)  not only enabled early warning before disasters but also facilitated the efficient rescue of victims, ensuring efficient navigation around obstacles. UAVs employed a coverage path planning (CPP) approach to continuously patrol and map unknown areas. Upon detecting a fire, unmanned vehicles prioritized rescue efforts based on the victims' weightings within the area. Additionally, considering the limitations of single drone and vehicle applications in large-scale maps, we proposed a sub-area strategy, ensuring at least one drone and vehicle are deployed in each area for coordinated operation. In the validation of simulation results, we initially tested the algorithm with a single drone and vehicle in simple maps, followed by further validation with multiple drones and vehicles in complex environments. The final results confirmed the high performance of the algorithm.

\subsection{Future work}
In this report, we conducted extensive research in motion control. However, it still needs to improve to keep pace with the rapid rate of progress. Our study accomplished two notable breakthroughs: firstly, the shift from a 2D to a 3D obstacle avoidance system; secondly, the progression from a single UAV system to a multi-UAV system, investigating the possibility of integrating UAVs with unmanned vehicles. Nevertheless, every progress in UAV technology is essential to achieve complete automation. As indicated in our Chapter 2 survey, the major issues that require significant attention for future study in motion control include establishing effective communication protocols among several UAVs, overcoming obstacles that obstruct the line of sight, optimizing energy consumption and other areas.

In complex urban environments, surveillance based on UAVs has increasingly become a mainstream approach due to issues with obstructions \cite{qadir2021addressing,huang2021path,lee2017optimal}. The presence of high-rise buildings and narrow streets can limit the field of vision of UAVs, making the execution of continuous surveillance tasks in such challenging urban landscapes a daunting task. In reference \cite{semsch2009autonomous}, the authors propose an obstruction-aware control mechanism. This mechanism initially constructs a set of advantageous observation points and then formulates a trajectory that respects the movement constraints of UAVs, facilitating UAV-based surveillance in complex urban settings. To address the limited perception range of UAVs, a covert video surveillance method is introduced in reference \cite{huang2021online}, which enables effective monitoring of ground-moving targets. 

Additionally, the issues of energy and communication in UAVs should be given priority in future research developments. Due to UAVs' outstanding manoeuvrability and autonomous capabilities were initially used in both military and commercial applications. Nevertheless, the constraints imposed by onboard batteries provide an essential challenge for UAVs in carrying out extended missions. Hence, the advancement of solar-powered unmanned aerial vehicles holds great potential, with the primary goal being resolving the energy consumption issue. Numerous studies and research endeavours have been undertaken to substantiate the substantial market potential of SUAV. Having solar panels on the UAV increases its carrying capacity and supports sustainable growth  \cite{safyanu2019review,morton2015solar,klesh2009solar}. 

Using solar-powered drones to optimise energy collection presents substantial opportunities for developing UAV navigation and obstacle avoidance technologies. By enhancing energy efficiency, the UAV extends its operational range and ensures that its sophisticated capabilities, including accurate navigation and efficient obstacle avoidance, are adequately powered. These developments will broaden the range of circumstances in which solar-powered UAVs can be utilised and increase their dependability. For instance, in a study \cite{huang2020autonomous} similar to my research, the authors employed solar-powered UAVs for safety and rescue missions in mountainous areas, examining their effectiveness. This study also utilized the RRT method but with an added focus on optimizing energy efficiency to find shorter paths while ensuring that the UAVs avoid collisions with mountains. In reference \cite{huang2016energy}, the authors explored the task of solar-powered UAVs tracking moving ground targets, validating the energy-optimal routes proposed by their method through theoretical and practical verification. In reference \cite{di2016coverage}, researchers investigated the application issues of coverage path planning and UAV battery energy, aiming for the model to reduce energy consumption without decreasing the required resolution, thereby implementing an improved coverage path planning method. 

A municipality's infrastructure and communications systems may be entirely or partially devastated in the event of a natural disaster. When this occurs, communications become critical to the search and rescue operation. Currently, the deployment efficiency of traditional communication base stations could be better. In contrast, solar-powered UAVs possess wireless transceivers that can hover at designated locations. This configuration facilitates the provision of communication services that are both more practicable and efficient. By adopting this methodology, the overall system's responsiveness and dependability are effectively enhanced, and the flexibility and adaptability of the communications coverage are substantially improved. In \cite{song2021energy,padilla2020flight}, the authors have conducted studies on optimising SUAVs communication systems serving multiple terrestrial users. These studies consider the signal path loss and shadowing effects and focus on maximising the energy efficiency and sustainability of the system. The extended flight duration has garnered primary academic interest since this strategy can enhance cost-effectiveness and minimise maintenance expenses while simultaneously ensuring extended flight duration \cite{romeo2007design,gao2013energy}. 

The primary obstacle for SUAVs in achieving long-endurance flight time is to enhance solar energy utilisation efficiency throughout energy collecting, flight and communication processes \cite{zhang2020power}. Therefore, energy harvesting models hold significant importance in examining SUAV navigation issues. These models are not only connected to the energy efficiency of the SUAV, but also directly linked to its flight performance and the effectiveness of mission execution. The relationship between the solar panel surface and the solar radiation illumination is a fundamental aspect of the modelling discussed in \cite{padilla2020flight}. On this basis, the SUAV motion and communication model is established, and the sustainability of a solar-powered UAV platform for communication in an urban environment is validated. Moreover, this methodology can also be applied as a primary research framework for mobile aerial platforms for disaster relief assistance, thereby showing its potential application in emergency relief. In \cite{sun2019optimal}, the authors employed a combined optimal 3D trajectory and power adaptive strategy to optimise the SUAV communication system. By harnessing solar energy while maintaining communication performance, this strategy seeks to optimise the energy consumption of the trajectory by utilising the propagation properties of wireless signals and sunlight. In Xin's research \cite{wei2020comprehensive}, the problem of optimal trajectories using SUAVs can be solved by modifying the obstacle reaction coefficients for the interfered fluid dynamical system (IFDS) and then adjusting the factor of adjustable Lyapunov guidance vector field (ALGVF). The objective of this approach is to maximise solar energy storage while minimising tracking error in order to complete the target tracking task.

Over the years, the investigations of Multi-robot cooperation in various fields of perception, decision-making processes, and motion have constantly faced difficulties. The Air-Ground-Cooperation (AGC) system \cite{guerin2015uav} has demonstrated notable efficiency advantages in handling complex tasks in disaster areas. Subsequent works will focus more on distributed control laws for multi-robot systems to enhance the maintenance of inter-robot communication during locomotion. Given this objective, the wireless sensor network challenge encountered by UAVs and unmanned ground vehicles working together is a significant and demanding research field \cite{sneha2020localization}. Integrating wireless networked sensors and distributed navigation algorithms can effectively establish escape routes and ensure secure navigation during emergencies \cite{tseng2006wireless}. Wireless sensor networks are precious in terms of cost-effectiveness for navigating and managing many flying robots in the Industrial Internet of Things. This technology improves emergency response skills and introduces creative and efficient ways for navigating and managing multi-robot systems \cite{li2018wireless}.

Regarding the obstacle avoidance problem we have explored, further research might be expanded to other application scenarios outside the confines of terrestrial mountainous regions. For example, for Autonomous Underwater Vehicles (AUVs) and Unmanned Surface Vessels (USVs), the problem of autonomous and safe navigation on the surface or underwater is worthy of research. Unknown risks brought on by changes in the maritime environment, such as waves and currents, are required to be included in the surface environment when simulating the kinematics of unmanned surface vehicles. The kinematic features resulting from these parameters are pertinent to investigating obstacle avoidance strategies in uneven terrain discussed in this study \cite{polvara2018obstacle,marzoughi2021autonomous}. Autonomous underwater vehicles (AUVs) must prioritise the prevention of accidental collisions with obstacles while navigating underwater. This involves avoiding stationary obstacles and requires the ability to steer away from unknown obstacles \cite{sahoo2019advancements,savkin2022optimal}. Furthermore,  the formation of autonomous navigation systems of unmanned vehicles for the rapid and safe guidance of unmanned vessels in specific sea environments is of potential research value for studying multi-robot systems \cite{sun2020formation}.

Future advancements are highly probable in the UAV low-level control design domain, particularly in light of the drag effects and other disturbances that multirotor UAVs encounter. Congested spaces and similar environments present enormous challenges to developing navigation strategies for UAVs. To tackle this issue, additional investigation may be warranted into the integration of sophisticated controllers—including sliding mode controllers \cite{vaidyanathan2017applications,bandyopadhyay2009sliding,utkin2013sliding,reguii2022mobile}, H-controllers \cite{petersen2000robust,fu2006robust,liu2021robust}, switching controllers \cite{savkin2002hybrid,liberzon1999basic,kumar2021stable}, and Kalman state estimator based controllers \cite{petersen2012robust,hu2003adaptive,GG8,jing2017attitude,liu2020interacting}, which can be combined with the navigation system as the core part of the coordinated control strategy for UAV motion. This integrated control method can effectively mitigate uncertainty in the mathematical model of UAVs, enhancing the aircraft's manoeuvring capability in complex environments and ensuring safer and more efficient operations.

In conclusion, further investigation may expand on our prior experience employing UAVs and other UGVs to address mountain fire rescue missions. This research aims to enhance the efficiency of  UAV drone cooperatives by incorporating factors such as energy consumption and communication interactions into CPP deployments. Furthermore, considering the intricate and continuously evolving nature of urban settings, future studies must investigate methods for accurately monitoring mobile subjects inside these areas. This may involve the creation of novel algorithms to enhance the precision of UAV navigation in urban landscapes, along with enhancing UAVs' communication and data processing capabilities in intricate settings. This research has the potential to enhance the effectiveness and security of rescue missions, as well as offer technological assistance for other uses in urban settings, like traffic surveillance and environmental monitoring.

	\bibliographystyle{IEEEtran}
	\bibliography{mypaper}
\end{document}